\documentclass[11pt, a4paper, logo, copyright, nonumbering, amsart]{map}

\usepackage{natbib}

\usepackage{CJKutf8}

\usepackage{xargs}  

\usepackage{todonotes}  

\usepackage{multirow}

\usepackage{cleveref}

\usepackage{amsmath}

\usepackage{dsfont}

\usepackage{subcaption}

\usepackage{svg}
\usepackage[breakable]{tcolorbox}
\usepackage{fontawesome}
\usepackage{xcolor}

\usepackage[utf8]{inputenc}
\usepackage{booktabs}
\usepackage{graphicx}
\usepackage{array}
\usepackage{tabularx}
\usepackage{xcolor,colortbl}  % 引入 xcolor 包
\definecolor{boxcolor}{HTML}{d92523} % 框的颜色 #dea3a2
\definecolor{bulbcolor}{HTML}{e3b87f} % 灯泡的颜色 #e3b87f

\usepackage{booktabs}

\usepackage{multirow}
\usepackage{adjustbox}
\usepackage{makecell}
\usepackage{wrapfig, lipsum, booktabs}
\usepackage{tcolorbox}

\definecolor{airforceblue}{rgb}{0.796,0.878,0.937}
\definecolor{airforceblue}{rgb}{0.828,0.914,0.910}
\definecolor{lightblue}{rgb}{0.933,0.968,0.988}
\definecolor{codeblue}{rgb}{0.215,0.686,0.847}
\definecolor{ora}{rgb}{0.914,0.443,0.196}

\hypersetup{
    colorlinks=true,            % 链接颜色
    linkcolor=blue,             % 内部链接
    filecolor=magenta,          % 本地文档
    urlcolor=cyan,              % 网址链接
    citecolor=purple,           % 引用颜色
    pdftitle={Overleaf Example},
    pdfpagemode=FullScreen,
}
 
\renewcommand{\today}{}
\newcommandx{\info}[2][1=]{\todo[linecolor=red,backgroundcolor=red!25,bordercolor=red,#1]{#2}}

\title{
    \centering CodeCriticBench: A Holistic Code Critique Benchmark for\\Large Language Models
    \vspace{-0.2in}
}

\author{
            \vspace{-0.1in}
    \textbf{Alexander Zhang$^{2,\ast}$,}
    \textbf{Marcus Dong$^{2,\ast}$,}
    \textbf{Jiaheng Liu$^{1, 2, \ast, \dagger}$,}
    \textbf{Wei Zhang$^{4}$,}
    \textbf{Yejie Wang$^{6}$,}
    \textbf{Jian Yang$^{4}$,}\\
    Ge Zhang$^{2}$,
    Tianyu Liu$^{2}$,
    Zhongyuan Peng$^{5}$,
    Yingshui Tan$^{3}$,
    Yuanxing Zhang$^{7}$,
    Zhexu Wang$^{6}$,\\
    Weixun Wang$^{3}$,
    Yancheng He$^{3}$,
    Ken Deng$^{3}$,
    Wangchunshu Zhou$^{2,8}$,\\
    Wenhao Huang$^{2}$,
    Zhaoxiang Zhang$^{5}$
    \\
        \vspace{0.1in}
    ~\textsuperscript{\rm 1}{NJU}, \textsuperscript{\rm 2}{M-A-P},~\textsuperscript{\rm 3}{Alibaba},~\textsuperscript{\rm 4}{BUAA}, ~\textsuperscript{\rm 5}{CASIA},~\textsuperscript{\rm 6}{BUPT},~\textsuperscript{\rm 7}{Kuaishou},~\textsuperscript{\rm 8}{OPPO}
    \vspace{-0.1in}

\url{https://github.com/multimodal-art-projection/CodeCriticBench}
    \vspace{-0.2in}
}

\begin{abstract}

\vspace{-0.2in}

The critique capacity of Large Language Models (LLMs) is essential for reasoning abilities, which can provide necessary suggestions (e.g., detailed analysis and constructive feedback). Therefore, how to evaluate the critique capacity of LLMs has drawn great attention and several critique benchmarks have been proposed. However, existing critique benchmarks usually have the following limitations: (1). Focusing on diverse reasoning tasks in general domains and insufficient evaluation on \textbf{code tasks} (e.g., only covering code generation task), where the difficulty of queries is relatively easy (e.g., the code queries of CriticBench are from Humaneval and MBPP). (2). Lacking comprehensive evaluation from \textbf{different dimensions}. To address these limitations, we introduce a holistic code critique benchmark for LLMs called \textbf{CodeCriticBench}. Specifically, our CodeCriticBench includes two mainstream code tasks (i.e., \textbf{code generation} and \textbf{code QA}) with different difficulties. Besides, the evaluation protocols include \textbf{basic critique evaluation} and \textbf{advanced critique evaluation} for different characteristics, where fine-grained evaluation checklists are well-designed for advanced settings. Finally, we conduct extensive experimental results of existing LLMs, which show the effectiveness of CodeCriticBench. 
\end{abstract}

\begin{document}
\begin{CJK*}{UTF8}{gbsn}

\maketitle

\let\oldthefootnote\thefootnote

\let\thefootnote\relax\footnotetext{*~Equal Contribution. ~~$^\dagger$~Corresponding Author.}
\let\thefootnote\oldthefootnote

\begin{figure*}[h]
    \centering
    \includegraphics[width=0.9\linewidth]{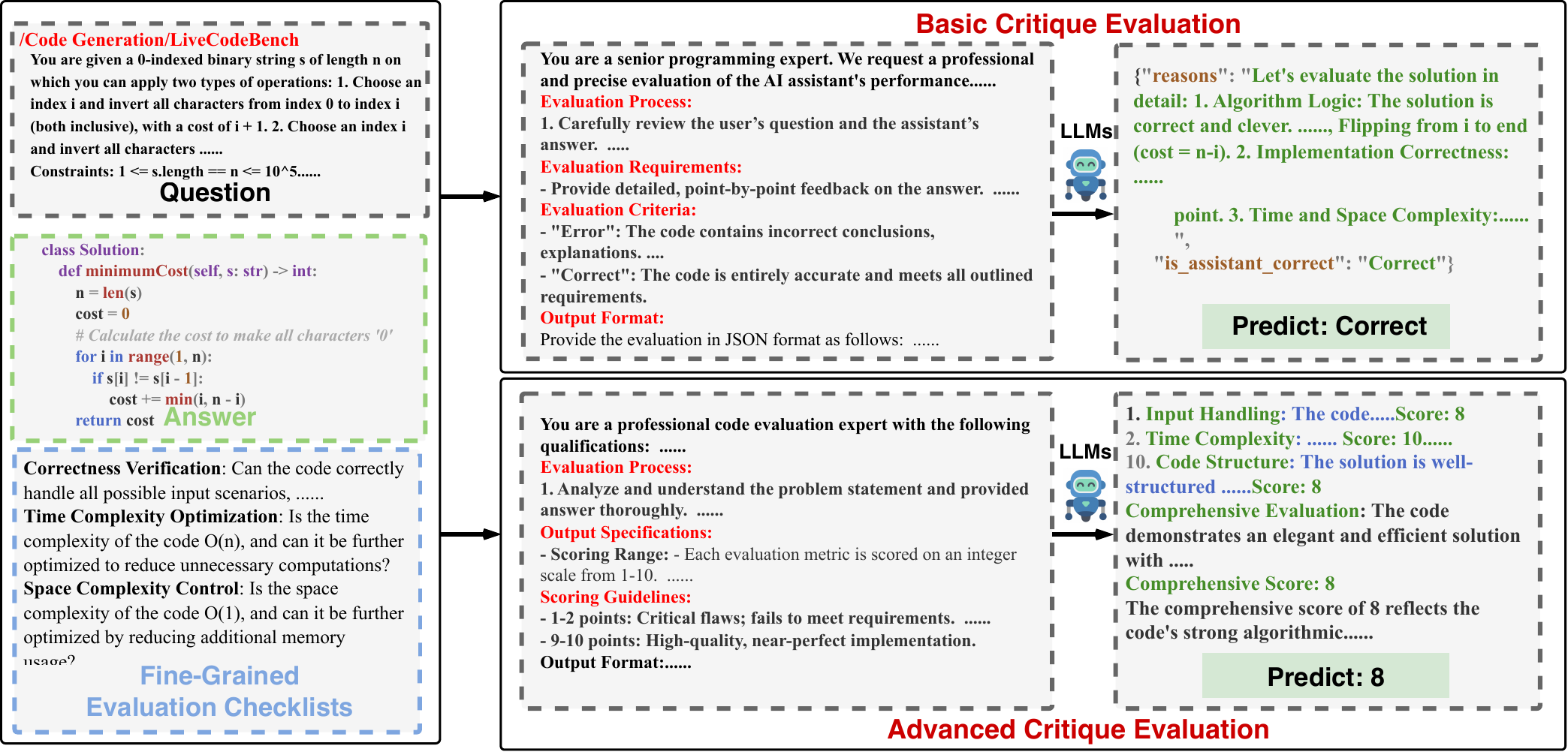}
    \caption{Illustration of the Basic Critique Evaluation and Advanced Critique Evaluation.} \label{fig:intro}
\end{figure*}

\clearpage

\tableofcontents

\clearpage

\section{Introduction} 

Large Language Models (LLMs) have demonstrated remarkable capabilities across various domains~\citep{dubey2024llama3,touvron2023llama,rozire2023codellama}, including natural language processing, code generation and complex reasoning tasks~\citep{wang2025mtubench,liu2024roleagent,liu-etal-2024-e2,bai2024mt}. As these models continue to evolve, their ability to critique and refine their outputs has emerged as a crucial area of research. This critique capacity is fundamental to enhancing the reasoning abilities of LLMs, which enables them to provide detailed analysis and constructive feedbacks and improves the quality and reliability of their outputs~\citep{madaan2023selfrefine,Song2025ProgCoPH}.

Recently, several critique benchmarks have been proposed~\citep{lan2024criticeval,lin2024criticbench,judgebench,prmbench,Zheng2024ProcessBenchIP,tang2025real}. For example, CriticBench~\citep{zheng2024critic} is proposed to assess LLMs' abilities to critique across a variety of tasks including math, commonsense, symbolic, code and algorithmic,
where the code subset is from the code generation datasets (i.e., Humaneval~\citep{chen2021evaluatinglargelanguagemodels} and MBPP~\citep{mbpp}). However, these benchmarks have predominantly focused on diverse reasoning tasks in general domains, leaving a significant gap in the evaluation of code-related tasks. Moreover, these benchmarks often lack comprehensive evaluation across different dimensions of critique capacity. Besides, in software development, the ability of LLMs to generate, understand and critique code is important. As LLMs are increasingly employed in coding assistance tools and automated code review systems, there is a pressing need for a robust framework to evaluate their code critique capabilities. This evaluation should encompass not only the accuracy of \textbf{code generation} but also the model's ability to provide insightful feedback, identify errors and suggest improvements in the \textbf{code QA} scenario.

To address these limitations, as shown in Figure~\ref{fig:intro}, we introduce \textbf{CodeCriticBench}, a holistic code critique benchmark for LLMs. Specifically, first, our benchmark covers two mainstream code tasks: code generation and code QA. These tasks are presented with varying levels of difficulty, allowing for a nuanced assessment across different coding challenges. Second, CodeCriticBench incorporates both \textbf{basic and advanced critique evaluations} for assessing different characteristics. For the basic setting, we prompt the judge model to provide the ``correct/error'' response and the corresponding reasoning process. For the advanced setting, we have developed fine-grained evaluation checklists for each problem, enabling a more detailed and precise assessment of the models' critique capabilities, which allows for a more thorough evaluation of existing LLMs. In summary, our contributions are as follows:

\begin{itemize}

\item To evaluate the critique abilities of LLMs on the code domain, we introduce the first holistic code critique benchmark \textbf{CodeCriticBench}, which includes the critique on both code generation and code QA tasks.

\item Based on our proposed CodeCriticBench, we design both the basic and advanced critique evaluations, which provide a comprehensive analysis of the critique ability.

\item We systematically evaluate the critique abilities of 38 LLMs (including general and code-specific models) on CodeCriticBench and provide detailed analysis regarding the critique capabilities of LLMs.

\end{itemize}

\section{Related Works}

\subsection{Code LLMs} 

Large Language Models (LLMs) have rapidly developed and are significantly impacting automated software development. These foundational models can produce fluent human language and understand semantics to perform complex tasks, bringing new possibilities to automated code generation, including Santacoder~\cite{allal2023santacoder}, CodeGPT~\cite{lu2021codexglue}, etc. More and more open-source and proprietary Code LLMs have emerged and demonstrated competitive performance, including Starcoder~\cite{li2023starcoder}, CodeLlama~\cite{roziere2023code}, Wizardcoder~\cite{luo2023wizardcoder}, DeepSeek-Coder~\cite{guo2024deepseek}, Qwen-Coder~\cite{hui2024qwen2}, Claude-3.5-Sonnet~\cite{claude_2024} and GPT-4o~\cite{gpt_4o} etc. 

\subsection{Critic Models for LLMs}

Reinforcement learning from human feedback has proven effective~\cite{achiam2023gpt}, though it can be very labor-intensive. A promising approach involves using LLMs to themselves to assist in the evaluation, which can further improve model outputs~\cite{saunders2022self, mcaleese2024llm}. Accurate and informative critique feedback allows LLMs to refine outputs, moving towards more advanced intelligence. However, despite their problem-solving strengths, LLMs currently demonstrate weak performance in critique tasks~\citep{zheng2024critic, yang2024supercorrect}. Improving LLM critique abilities relies on supervision from human annotations~\cite{saunders2022self,mcaleese2024llm} and stronger LLMs acting as human proxies~\cite{lan2024training,zhang2024generative,ke2024critiquellm,ankner2024critique,zheng2024critic,yang2024supercorrect,sharma2024critical}.
Recently, some critic benchmarks~\citep{zheng2024critic,tang2025real} have also been proposed.

\section{CodeCriticBench}

\subsection{Overview}

\begin{wraptable}{r}{3.0in}
    \vspace{-0.2in}
    \centering \caption{Dataset statistics of CodeCriticBench.} \label{table:bench_statistic}
    \begin{adjustbox}{width=0.47\textwidth}
    \begin{tabular}{lr}
        \hline
        \toprule
        \textbf{Statistics} & \textbf{Number} \\
        \midrule
        \textbf{\#Problems} & $4,300$ \\
        \midrule
        \textbf{Difficulty Level} & \\
        - Easy/Medium/Hard & $1,517$/$1,084$/$1,699$ \\
        \midrule
        \textbf{Length} \\
        Question \\
        ~~~~- \textit{maximum length}   & $32,063$ tokens \\
        ~~~~- \textit{minimum length}   & $8$ tokens \\
        ~~~~- \textit{average length}   & $451.06$ tokens \\
        Answer \\
        ~~~~- \textit{maximum length}   & $32175$ tokens \\
        ~~~~- \textit{minimum length}   & $9$ tokens \\
        ~~~~- \textit{average length}   & $322.22$ tokens \\
        Fine-Grained Evaluation Checklists \\
        ~~~~- \textit{maximum length}   & $676$ tokens \\
        ~~~~- \textit{minimum length}   & $97$ tokens \\
        ~~~~- \textit{average length}   & $287.49$ tokens \\
        \bottomrule
        \hline
    \end{tabular}
    \end{adjustbox}
% \end{table}
\end{wraptable}

CodeCriticBench is a holistic evaluation benchmark for Code Critic tasks, encompassing code evaluation in both Code Generation and Code QA tasks. Table~\ref{table:bench_statistic} presents an overview of CodeCriticBench, which comprises 4,300 samples. Each sample includes a question, an answer, a set of fine-grained evaluation checklists across multiple dimensions and associated labels—namely, correctness labels, per-dimension evaluation scores, a final score and a difficulty level. Furthermore, we employ the LLaMA3 tokenizer~\cite{dubey2024llama} to determine the token counts for questions, answers and evaluation problems in Table~\ref{table:bench_statistic}. 
For fine-grained evaluation checklists, token counts are summed across all evaluation problems. As reported in Table~\ref{table:bench_statistic}, the average token length is 451.06 for questions, 322.22 for answers and 287.49 for the aggregated evaluation checklists.

Most existing Code Critic Benchmarks primarily focus on evaluating basic properties, such as code correctness. In contrast, CodeCriticBench offers both a fundamental assessment of code correctness and a comprehensive evaluation of associated question–answer pairs, alongside a fine-grained scoring system that spans multiple dimensions. Notably, every data sample in CodeCriticBench is paired with uniquely tailored, fine-grained evaluation checklists, ensuring that each instance is assessed in a context-specific and comprehensive manner. Table~\ref{table:benchmark_compare} presents a comparison between CodeCriticBench and other Code Critic Benchmarks.

\begin{table}[t]
    \centering \small
    \caption{Comparisons between CodeCriticBench and other benchmarks. ``Code Gen'' denotes ``Code Generation''. ``-'' represents that the corresponding task is not included in the dataset.} \label{table:benchmark_compare}
    \resizebox{0.8\textwidth}{!}{
    \begin{tabular}{lccccc}
        \toprule
        \textbf{Benchmark} & \textbf{Data Size} & \textbf{Code Gen.} & \textbf{Code QA} & \textbf{Basic}& \textbf{Advanced} \\ 
        \midrule
        CriticBench & 3,825 & 464 & - & \checkmark & \texttimes \\
        CriticEval & 3,608 & 1,340 & - & \checkmark & \texttimes \\
        JudgeBench & 350 & 42 & - & \checkmark & \texttimes \\
        RealCritic & 2,093 & - & - & \checkmark & \texttimes \\
        \midrule
        CodeCriticBench & 4,300 & 3,200 & 1,100 & \checkmark & \checkmark\\
        \bottomrule
    \end{tabular}
    }
\end{table}

\subsection{Data Collection}

% \begin{figure}[h]
\begin{wrapfigure}{r}{0.6\textwidth}
    \centering
    \includegraphics[width=\linewidth]{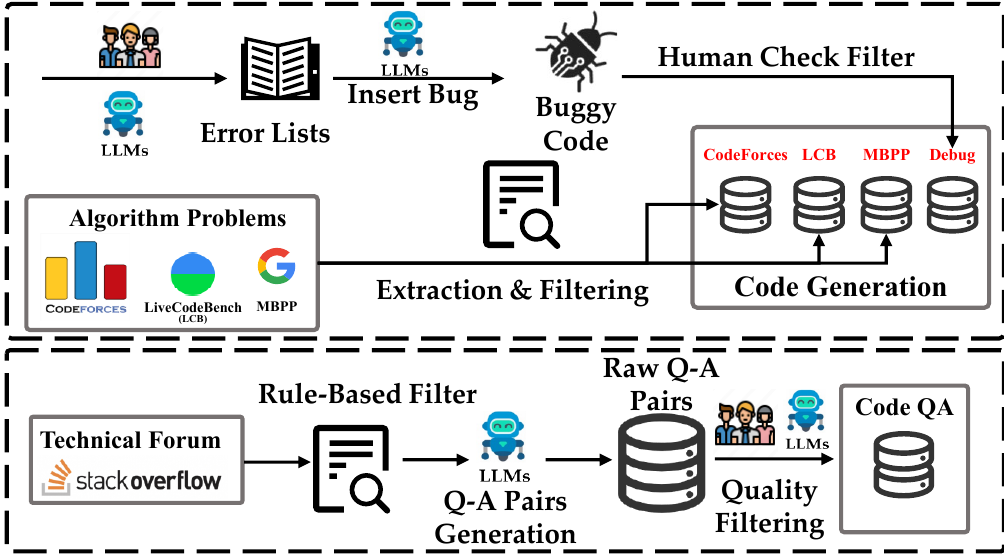}
    \caption{Illustration of data collection process.} \label{fig:data_collection}
\end{wrapfigure}
% \end{figure}

\textbf{Code Generation.} In the code generation task, instructions are mainly algorithmic problems. We collect many such data (restricted to test sets) from CodeForces, MBPP~\cite{mbpp} and LiveCodeBench~\cite{Jain2024LiveCodeBenchHA}. Given the limited test sets in MBPP and LiveCodeBench, we employ DeepSeek-v3 to rewrite problem while preserving their semantics, thereby enhancing diversity and test case reusability. To assess the model's ability to detect specific programming errors, we create a Debug subset. We first compile a list of bugs through iterative discussions with experts and LLMs, then prompt the LLM to insert specified error types into sandbox-verified correct code. The modified samples undergo two filtering rounds: sandbox execution to confirm error-triggering and manual review to ensure the errors matched their intended categories. Details on the error types and bug insertion prompts are provided in Appendix~\ref{ap:prompt}.

\noindent \textbf{Code QA.} Building on previous work~\cite{baars2019codearena}, we first apply a rule-based filtering method to clean the extracted content, removing site information, ads and HTML tags. To ensure our benchmark reflects real-world scenarios, we collect authentic code requirements and manually craft responses from StackOverflow. Unlike~\cite{yue2024mammoth2}, which directly extract question–answer pairs, we use Qwen2.5-72B to generate new questions. As shown in Figure~\ref{fig:data_collection}, Qwen2.5-Coder samples the same document multiple times to generate diverse responses, which are then filters through manual and LLM-assisted reviews, forming a high-quality subset.

\subsection{Dataset Construction}

\textbf{Difficulty.} To assess sample difficulty, we employ twelve state-of-the-art LLMs\footnote{Claude3.5-Sonnet, GPT 4o, DeepSeek-v2.5, DeepSeek-v3, Doubao-Coder-Preview, Llama3.3-70B-Instruct, Qwen2.5-72B-Instruct, DeepSeek-R1-Distill-Qwen-32B, DeepSeek-R1, GLM-4-Plus, Qwen2.5-Max and OpenAI o1-Preview.} and classify each sample based on the proportion of models that produce correct predictions. Specifically, a sample is labeled ``Easy'' if at least 80\% of the models reason correctly, ``Medium'' if the success rate is between 60\% and 80\% and ``Hard'' if fewer than 60\% succeed. Note that because random scoring yields an expected success rate of approximately 50\%, a success rate below 60\% suggests that the model's performance is close to random guessing. This process yields 1,517 Easy, 1,084 Medium and 1,699 Hard samples, ensuring a balanced distribution of difficulty levels in CodeCriticBench.

\noindent \textbf{Category.} To determine the specific application scenarios of the ``Code QA'' subset, we prompt DeepSeek-v3 to classify the data based on a predefined category list adopted from \cite{liu2024fullstack}. Detailed prompts used for this classification are provided in Appendix~\ref{ap:prompt}.

\noindent \textbf{Correctness.} For the ``Code Gen'' subset, each sample is accompanied by test cases that facilitate automated correctness evaluation within a sandbox. A sample is marked as ``Correct'' only if all test cases pass; otherwise, it is labeled as ``Error.'' In contrast, for the ``Code QA'' subset, we engage 20 volunteers with coding experience to assess correctness. Each question is independently evaluated by three raters and the final correctness label is determined by majority vote.

\noindent \textbf{Fine-Grained Evaluation Checklists and Corresponding Scores.} In contrast to previous Code Critic Benchmarks, we introduce a more detailed and customized scoring framework. Initially, we define 10 evaluation dimensions for both code generation and code QA tasks through iterative discussions with human experts and LLMs. Next, we employ prompts for DeepSeek-V3 to generate tailored evaluation questions for each data instance. To ensure the generated questions met the desired criteria, we perform random sampling and manual inspections, repeating the process until the pass rate exceeds 95\%. Inspired by previous work~\cite{que2024hellobench}, we manually annotate 20\% of the dataset to establish baseline scores. Subsequently, three advanced models (Claude3.5-Sonnet, GPT-4o and Gemini2.0), evaluate the entire dataset, with each evaluation dimension’s final score determined by majority vote. Finally, we calibrate the LLM-generated scores by applying linear regression, using them as independent variables (x) and the human ratings as the dependent variable (y), to produce the final scores for all samples. Detailed prompts and multiple evaluation names are provided in Appendix~\ref{ap:prompt} and \ref{ap:main_dimension}.

\subsection{Evaluation Metrics}

We employ multiple evaluation metrics to rigorously assess our model's performance.

\noindent \textbf{Accuracy (ACC).} ACC measures binary classification performance by determining whether the model's predictions match the ground truth labels for basic critique evaluation. Let $N$ denote the total number of instances, $\hat{y}_i$ the predicted label for instance $i$ and $y_i$ the true label. Accuracy is computed as follows:
\begin{equation}
    \text{ACC} = \frac{1}{N} \sum_{i=1}^{N} I(\hat{y}_i = y_i)
\end{equation}

\noindent where $I(\hat{y}_i = y_i)$ is an indicator function that returns 1 if the prediction is correct and 0 otherwise.

\noindent \textbf{Mean Squared Error (MSE).} MSE quantifies the discrepancy between the model's predicted and true scores across multiple dimensions for advanced critique evaluation, which measures how closely the model's predictions approximate the actual values. MSE is calculated as follows:
\begin{equation}
    \text{MSE} = \frac{1}{N} \sum_{i=1}^{N} (\hat{y}_i - y_i)^2
\end{equation}

\noindent where $\hat{y}_i$ represents the predicted score and $y_i$ the true score for instance $i$. 

\noindent \textbf{Pass@1 Accuracy.} In code error detection, each code snippet contains at least one error ($n_i \geq 1$). The model's objective is to detect at least one error per snippet. Let $\hat{E}_i$ denotes the set of errors predicted by the model for instance $i$ and $E_i$ the set of actual errors. A prediction is considered successful if the intersection between $\hat{E}_i$ and $E_i$ is non-empty. Pass@1 accuracy is defined as:
\begin{equation}
    \text{ACC}_{\text{Pass@1}} = \frac{1}{N} \sum_{i=1}^{N} I(\hat{E}_i \cap E_i \neq \emptyset)
\end{equation}

\section{Experiments}

\subsection{Baselines}

We base our model selection on benchmarks from the code domain, including those outlined in \cite{liu2024mdeval} and \cite{liu2024fullstack}, ultimately selecting 38 models for evaluation. These models encompass both open-source and closed-source options, with sizes ranging from 0.5 billion to over 100 billion parameters. A detailed list of the models tested, along with their respective links, is provided in Tables~\ref{table:open_source_model} and~\ref{table:api_model} in Appendix~\ref{ap:model_lists}.

\subsection{Main Results} 

% main_table
\begin{table}[!t]
    \centering \small
    \caption{Results of different models. ``gen'' and ``qa'' denote code generation and code qa tasks respectively. ``ACC'' and ``MSE'' metrics are used for basic and advanced critique evaluations.} \label{table:main_table}
    \begin{adjustbox}{width=0.8\textwidth}
    \begin{tabular}{c|cccccc}
    
    \toprule
    \textbf{Model} & \textbf{ACC\textsubscript{All}} & \textbf{ACC\textsubscript{gen}} & \textbf{ACC\textsubscript{qa}} & \textbf{MSE\textsubscript{All}} & \textbf{MSE\textsubscript{gen}} & \textbf{MSE\textsubscript{qa}} \\

    \midrule
    \multicolumn{7}{c}{0.5B+ Instruction Tuned Model} \\
    \midrule
    Qwen2.5-Coder-0.5B-Instruct & 51.79 & 51.28 & 53.27 & 24.24 & 23.96 & 25.06 \\
    
    \midrule
    \multicolumn{7}{c}{1B+ Instruction Tuned Model} \\
    \midrule
    Yi-Coder-1.5B-Chat & 50.98 & 51.56 & 49.27 & 29.26 & 28.27 & 32.13 \\
    DeepSeek-Coder-1.3B-Instruct & 51.44 & 52.31 & 48.91 & 22.67 & 24.48 & 17.39 \\
    OpenCoder-1.5B-Instruct & 53.16 & 53.06 & 53.45 & 27.60 & 28.07 & 26.23 \\
    Qwen2.5-Coder-1.5B-Instruct & 54.47 & 53.87 & 56.18 & 16.56 & 17.13 & 14.93 \\
    Qwen2.5-Coder-3B-Instruct & 55.21 & 55.22 & 55.18 & 12.43 & 13.01 & 10.76 \\

    \midrule
    \multicolumn{7}{c}{6B+ Instruction Tuned Model} \\
    \midrule
    CodeLlama-7B-Instruct & 54.47 & 53.94 & 56.00 & 19.12 & 20.89 & 13.96 \\
    Qwen2.5-Chat-7B-Instruct & 54.63 & 51.81 & 62.82 & 8.07 & 8.73 & 6.15 \\
    CodeQwen1.5-7B-Chat & 55.63 & 55.41 & 56.27 & 18.22 & 19.06 & 15.78 \\
    OpenCoder-8B-Instruct & 56.37 & 57.66 & 52.64 & 19.33 & 19.45 & 18.99 \\
    Qwen2.5-Coder-7B-Instruct & 57.95 & 56.78 & 61.36 & 5.64 & 6.17 & 4.10 \\
    Yi-Coder-9B-Chat & 61.67 & 63.66 & 55.91 & 14.21 & 14.28 & 14.02 \\

    \midrule
    \multicolumn{7}{c}{13B+ Instruction Tuned Model} \\
    \midrule
    CodeLlama-13B-Instruct & 52.17 & 52.16 & 52.18 & 13.94 & 15.55 & 9.26 \\
    StarCoder2-15B-Instruct & 53.43 & 53.59 & 52.97 & 21.42 & 20.92 & 22.88 \\
    Qwen2.5-Coder-14B-Instruct & 59.00 & 56.38 & 66.64 & 5.08 & 5.57 & 3.65 \\
    DeepSeek-v2-Lite-Chat & 59.20 & 58.78 & 60.42 & 5.57 & 6.35 & 3.29 \\
    DeepSeekCoder-v2-Lite-Instruct & 59.81 & 59.34 & 61.18 & 5.67 & 6.46 & 3.35 \\
    Qwen2.5-Chat-14B-Instruct & 59.98 & 58.59 & 64.00 & 4.38 & 5.02 & 2.51 \\

    \midrule
    \multicolumn{7}{c}{20B+ Instruction Tuned Model} \\
    \midrule
    CodeLlama-34B-Instruct & 54.79 & 54.06 & 56.91 & 13.45 & 15.06 & 8.76 \\
    Qwen2.5-Coder-32B-Instruct & 61.67 & 59.00 & 69.45 & 4.60 & 5.19 & 2.89 \\
    Qwen2.5-Chat-32B-Instruct & 63.98 & 62.38 & 68.64 & 4.32 & 5.09 & 2.09 \\

    \midrule
    \multicolumn{7}{c}{70B+ Instruction Tuned Model} \\
    \midrule
    DeepSeek-v2.5 & 60.35 & 58.46 & 65.85 & 3.97 & 4.78 & 1.63 \\
    DeepSeekCoder-v2-Instruct & 64.42 & 62.42 & 70.23 & 4.19 & 5.14 & 1.46 \\
    Llama3.3-70B-Instruct & 65.91 & 65.16 & 68.09 & 4.78 & 5.65 & 2.24 \\
    Qwen2.5-72B-Instruct & 68.35 & 68.44 & 68.09 & 3.99 & 4.61 & 2.20 \\
    % DeepSeek-v2-Chat & 64.63 & 63.69 & 67.36 & 9.55 & 5.17 & 22.30 \\

    \midrule
    \multicolumn{7}{c}{Close-Sourced API Model} \\
    \midrule
    GPT 4o-mini & 60.56 & 58.31 & 67.09 & 3.92 & 4.82 & 1.30 \\
    Doubao-Coder-Preview & 61.42 & 60.06 & 65.36 & 6.51 & 7.07 & 4.90 \\
    GLM-4-Plus & 61.55 & 60.94 & 63.35 & 3.60 & 4.25 & 1.69 \\
    DeepSeek-v3 & 62.00 & 61.44 & 63.64 & 3.64 & 4.49 & 1.18 \\
    Qwen2.5-Max & 63.36 & 62.74 & 65.17 & 4.09 & 5.04 & 1.33 \\
    GPT 4o & 68.06 & 67.56 & 69.53 & 4.15 & 5.04 & 1.55 \\
    Claude3.5-Sonnet & 68.79 & 66.06 & 76.73 & 3.78 & 4.73 & 1.02 \\

    \midrule
    \multicolumn{7}{c}{o1-like Models} \\
    \midrule
    QwQ-32B-Preview & 57.35 & 56.59 & 58.82 & 7.20 & 8.07 & 4.67 \\
    Gemini2.0-Flash-Thinking & 64.53 & 64.88 & 63.55 & 3.88 & 4.80 & 1.19 \\
    OpenAI o1-mini & 71.77 & 76.06 & 59.27 & 4.92 & 6.08 & 1.54 \\
    DeepSeek-R1-Distill-Qwen-32B & 72.49 & 75.38 & 64.09 & 4.34 & 5.25 & 1.71 \\
    DeepSeek-R1 & 72.76 & 79.09 & 54.36 & 4.20 & 3.92 & 5.02 \\
    OpenAI o1-Preview & 75.30 & 80.53 & 59.89 & 4.81 & 5.68 & 2.26 \\
    \bottomrule
    \end{tabular}
    \end{adjustbox}
\end{table}

% bug_acc
\begin{wraptable}{r}{3.1in}
% \begin{table}[h]
    \centering \small
    \vspace{-0.2in}
    \caption{The accuracy of different models in identifying programming error types.} \label{table:bug_acc}
    \begin{adjustbox}{width=0.5\textwidth}
    \begin{tabular}{l|l|c}
        \toprule
        \multicolumn{2}{c}{\textbf{Model}} & \textbf{ACC} \\
        \midrule
        \multirow{5}{*}{Close-Sourced} 
        & GLM-4-Plus & 47.75 \\
        & Claude3.5-Sonnet & 54.00 \\
        & Qwen2.5-Max & 59.50 \\
        & GPT 4o & 61.00 \\
        & Doubao-Coder-Preview & 67.00 \\
        \midrule
        \multirow{8}{*}{Open-Sourced} 
        & Qwen2.5-Coder-0.5B-Instruct & 21.50 \\
        & Qwen2.5-Coder-1.5B-Instruct & 32.75 \\
        & Qwen2.5-Coder-3B-Instruct & 48.00 \\
        & Qwen2.5-Coder-7B-Instruct & 61.00 \\
        & Qwen2.5-Chat-7B-Instruct & 41.00 \\
        & Qwen2.5-Coder-14B-Instruct & 62.25 \\
        & Qwen2.5-Chat-14B-Instruct & 47.50 \\
        & Qwen2.5-72B-Instruct & 61.25 \\
        \bottomrule
    \end{tabular}
    \end{adjustbox}
% \end{table}
\end{wraptable}

\textbf{Critique Evaluation.} Table~\ref{table:main_table} shows the basic evaluation (\textbf{ACC} metric) and advanced evaluation (\textbf{MSE} metric) for all models in our study. The results are grouped by model size, with separated blocks for closed-source and o1-like models. Within each size block, models are sorted by ACC. As observed, performance generally improves with more parameters, with o1-like models achieving milestone results and all exceed 70\%. In the advanced evaluation, DeepSeek-R1 performs best on the ``Code Gen'' subset with an MSE of 3.92, while Claude3.5-Sonnet leads in ``Code QA'' with an MSE of 1.02. This difference likely reflects the distinct nature of the two tasks: ``Code QA'' prioritizes concise, high-quality answers, while ``Code Gen'' emphasizes solving algorithmic problems that require stronger reasoning abilities, whereas ``Code QA'' may benefit more from dialogue optimization.

\noindent \textbf{Bug Identification.} We evaluate thirteen models to assess their accuracy in identifying code error types. As shown in Table~\ref{table:bug_acc}, larger models generally perform better in distinguishing code errors, which aligns well with our expectations.

\subsection{Further Analysis}

\begin{figure*}[h]
    \centering
    \includegraphics[width=\linewidth]{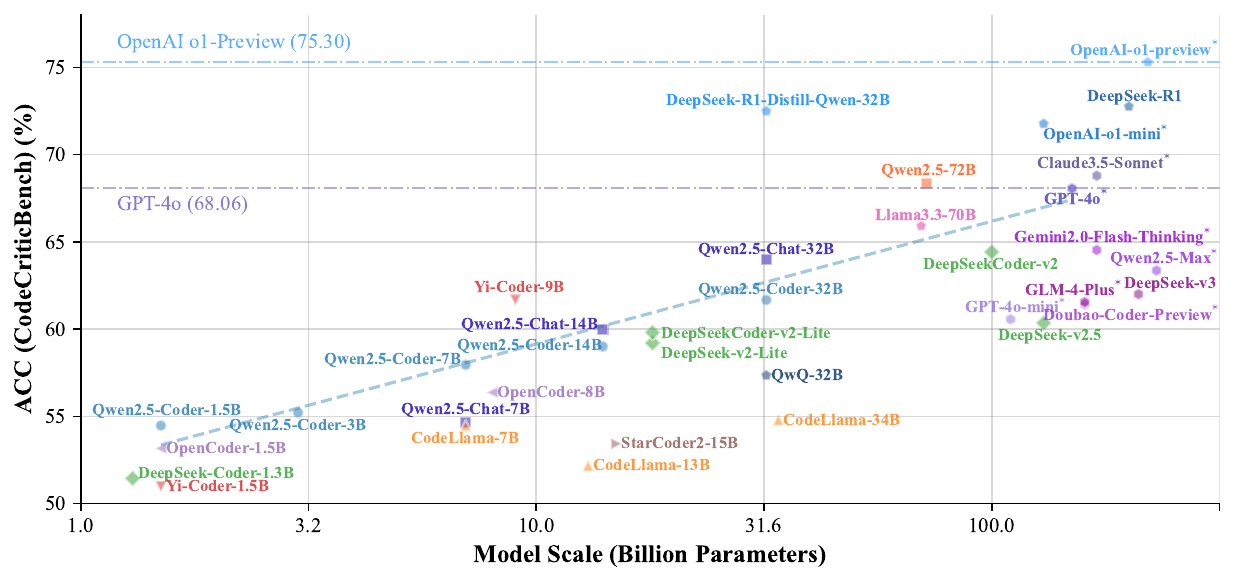}
    \caption{Scaling law on basic critique evaluation (ACC) across models. ``*'' indicates an estimated parameter size.} \label{fig:further_scaling_acc}
\end{figure*}

\textbf{Scaling Law.} To assess the effectiveness of the scaling law, we visualize the performance across nearly all models in our study. Figure~\ref{fig:further_scaling_acc} presents the results of the basic evaluation ACC, with corresponding MSE results provided in Appendix~\ref{ap:scaling}. The data clearly demonstrate that as the number of parameters increases, the ACC increases, which further validate the robustness of our CodeCriticBench dataset.

\begin{figure*}[h]
    \centering
    \includegraphics[width=\linewidth]{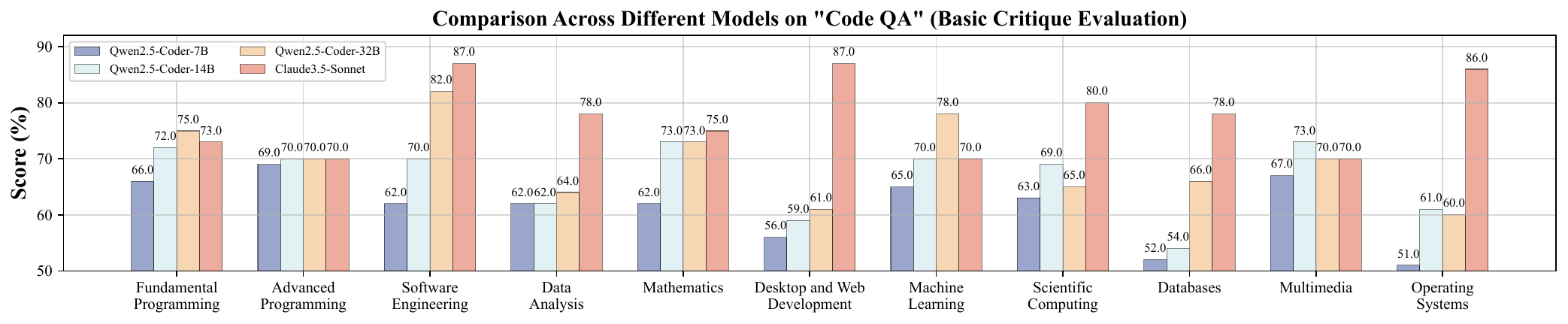}
    \caption{Comparison across different models on ``Code QA'' (Basic Critique Evaluation).} \label{fig:further_app_acc}
\end{figure*}

\noindent \textbf{Different Application Scenes.} Our ``Code QA'' data subset includes 11 scenarios, such as ``Fundamental Programming'', ``Software Engineering'' and `Mathematics''. To evaluate model performance across these domains, we test five models of varying sizes. We calculate their ACC on basic evaluation and MSE on advanced evaluation. As shown in Figures~\ref{fig:further_app_acc} and \ref{fig:further_app_mse}, larger models generally show improved ACC and reduced MSE, supporting our expectations and reinforcing their validity.

\begin{figure*}[h]
    \centering
    \includegraphics[width=\linewidth]{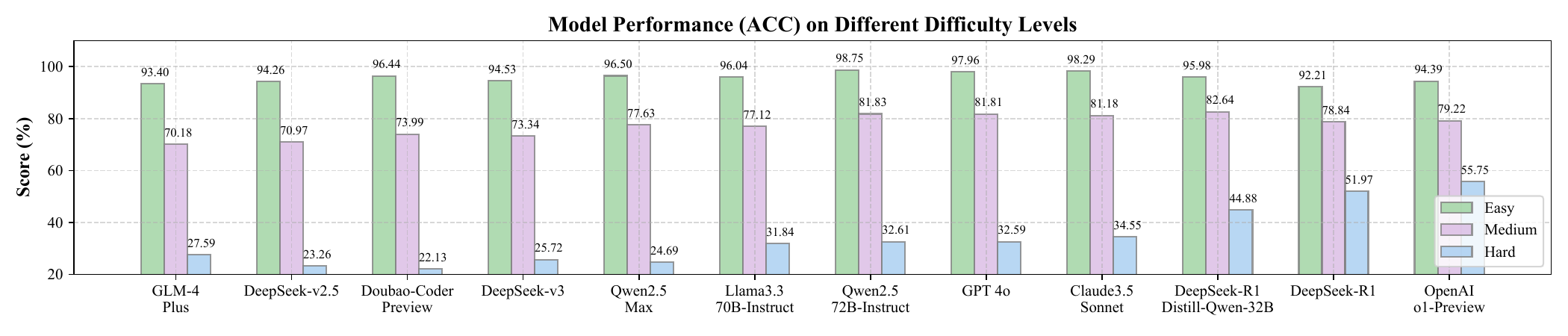}
    \caption{Model performance (ACC) on different difficulty levels  (Basic Critique Evaluation) .} \label{fig:further_difficulty_acc}
\end{figure*}

\noindent \textbf{Different Difficulty's Performance.} As previously discussed, the difficulty of data is quantified by the proportion of models that accurately predict its correctness. To further analyze performance across different difficulty levels, we evaluate twelve advanced LLMs. Figure~\ref{fig:further_difficulty_acc} shows that all models perform well on Easy data, with accuracies above 90\%. For Medium data, the top performers are Qwen2.5-72B-Instruct, GPT 4o, Claude3.5-Sonnet and DeepSeek-R1-Distill-Qwen-32B, each achieving around 82\%. This performance is consistent with their rankings on other established leaderboards. In contrast, DeepSeek-R1 and OpenAI o1-Preview underperform on Easy to Medium data, likely due to overthinking. On more challenging data, most models show a significant drop in performance, with accuracies around 30\%. However, DeepSeek-R1 and OpenAI o1-Preview maintain relatively strong performance, with accuracies of 51.97\% and 55.75\%, respectively. We hypothesize their strong performance is partly due to o1-like architecture, which help them handle complex data.

% \begin{figure}[h]
\begin{wrapfigure}{r}{0.65\textwidth}
    \centering
    \includegraphics[width=\linewidth]{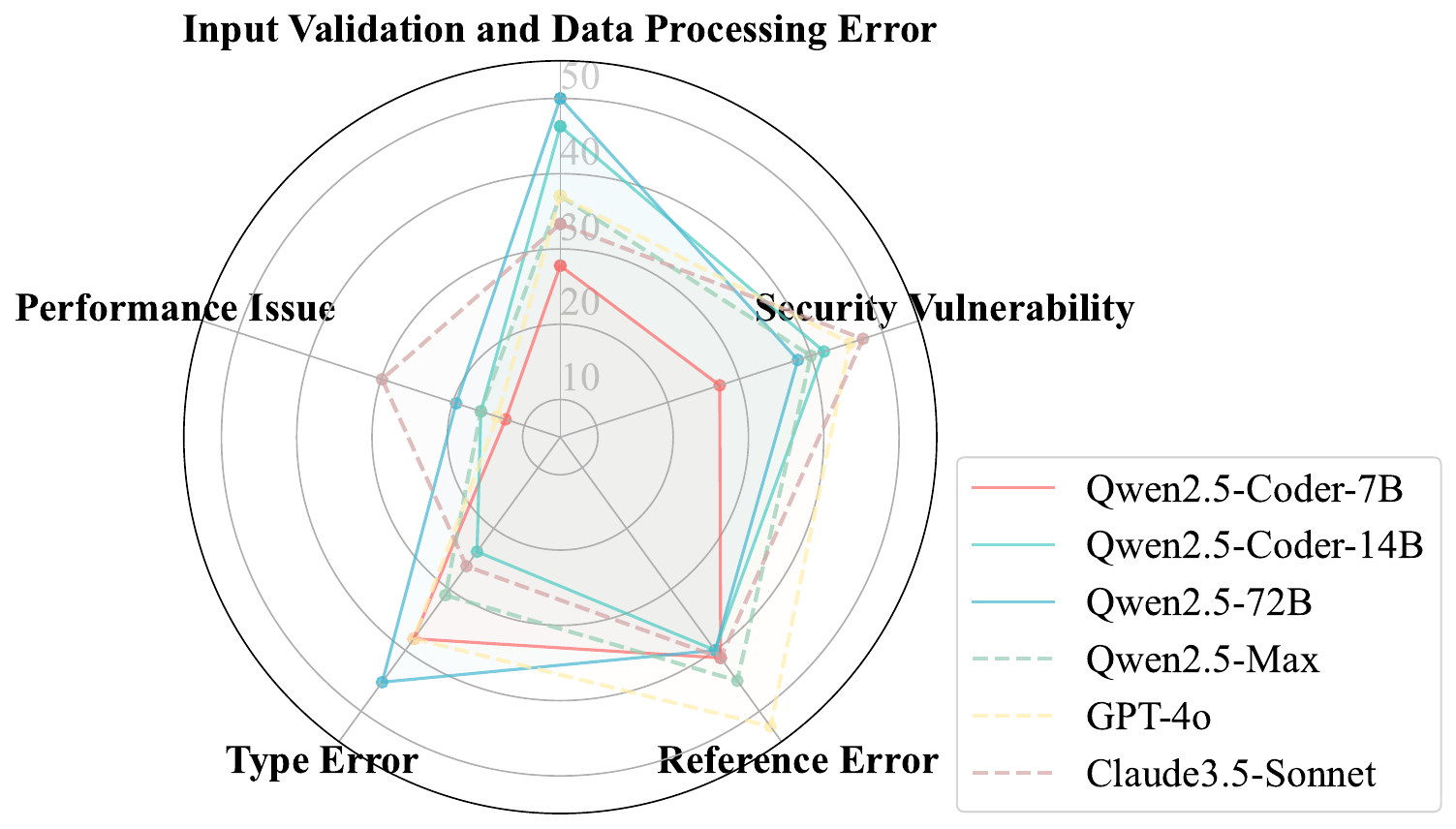}
    \caption{Comparison of the accuracy of different models in identifying five common programming error types.} \label{fig:radar}
% \end{figure}
\end{wrapfigure}

\noindent \textbf{Different Bug Types.} To gain a more intuitive understanding of how models with different parameter sizes detect code errors, we select five common error types and evaluate six models of varying sizes to assess their ability to correctly identify the corresponding error types. As shown in Figure~\ref{fig:radar}, the accuracy of error type identification generally improves with the increase in model size. Among the five error types, Qwen2.5-72B, Claude3.5-Sonnet and GPT-4o emerge as the top-performing models, each securing first place in the identification of 2, 2 and 1 error categories, respectively.

\begin{wrapfigure}{r}{0.5\textwidth}
% \begin{figure}[h]
    \vspace{-0.2in}
    \centering
    \includegraphics[width=\linewidth]{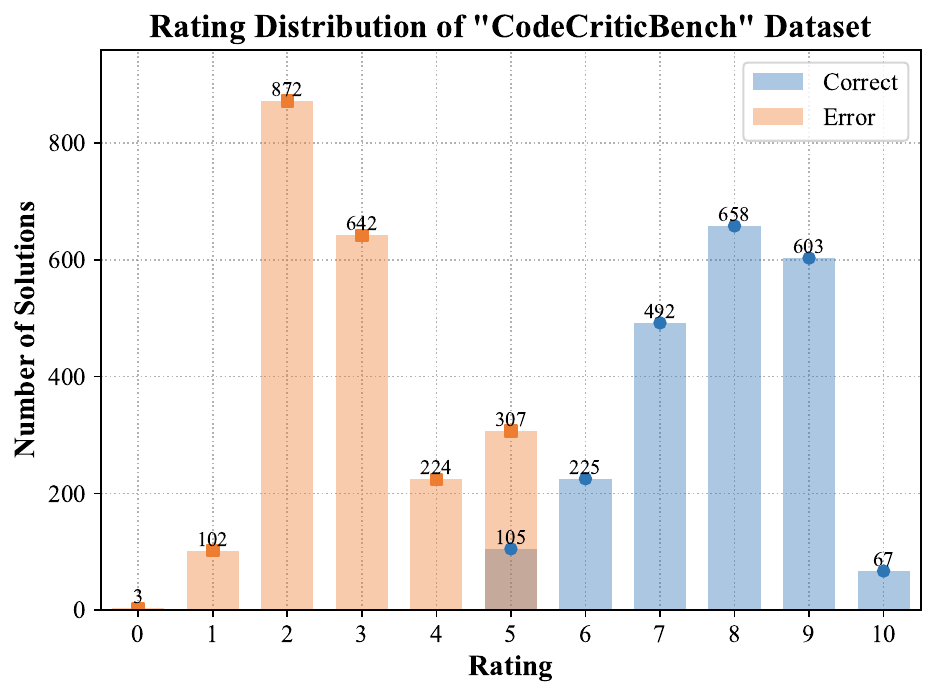}
    \caption{Rating distribution of the CodeCriticBench.} \label{fig:further_all_level12_rating_distribution}
% \end{figure}
\end{wrapfigure}

\noindent \textbf{Relations Between Basic Correctness and Advanced Evaluations Scores.} To explore the relationship between data correctness and the final multidimensional score, we examine the distribution of correct and error data across different scores. As shown in Figure ~\ref{fig:further_all_level12_rating_distribution}, almost all ``Error'' instances have scores below 5, while ``Correct'' instances consistently score 5 or higher. Correct scores are mainly around 7, 8 and 9, while error scores are typically between 2 and 3. Additional statistical details for the ``Code Gen'' and ``Code QA'' subsets can be found in Appendix~\ref{ap:rating_level12}. Overall, the consistency between data correctness and final scores supports the robustness of our dataset.

\noindent \noindent\textbf{Error Studies.} We evaluate seven distinct LLMs, both open-source and closed-source, of varying sizes, to assess their ability to identify three common error types: ``Reference Error'', ``Performance Issue'' and ``Security Vulnerability'', out of a total of twenty-three categories. The results for these error types are shown in Figure~\ref{fig:further_error_studies}, with additional details in Appendix~\ref{ap:bug}. As shown in Figure~\ref{fig:further_error_studies}, models with higher capabilities perform better at distinguishing errors. However, most models, except Doubao-Coder-Preview and Claude3.5-Sonnet, struggle with identifying ``Performance Issue''. We suggest this is due to insufficient code-related optimization and models' smaller sizes. 

\begin{figure}[h]
    \centering
    \includegraphics[width=0.8\linewidth]{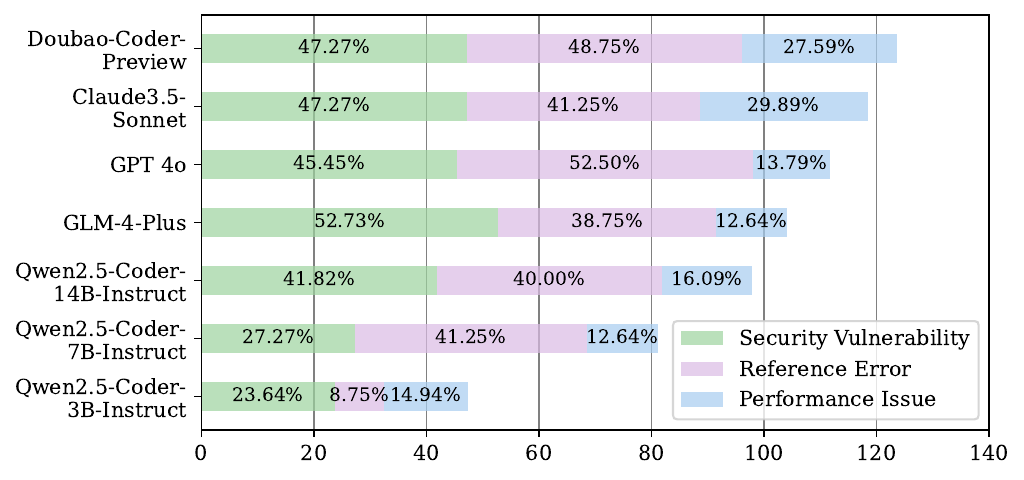}
    \caption{Experimental accuracy results of different models across various error types.} \label{fig:further_error_studies}
\end{figure}

\begin{figure}[h]
    \centering
    \includegraphics[width=0.6\linewidth]{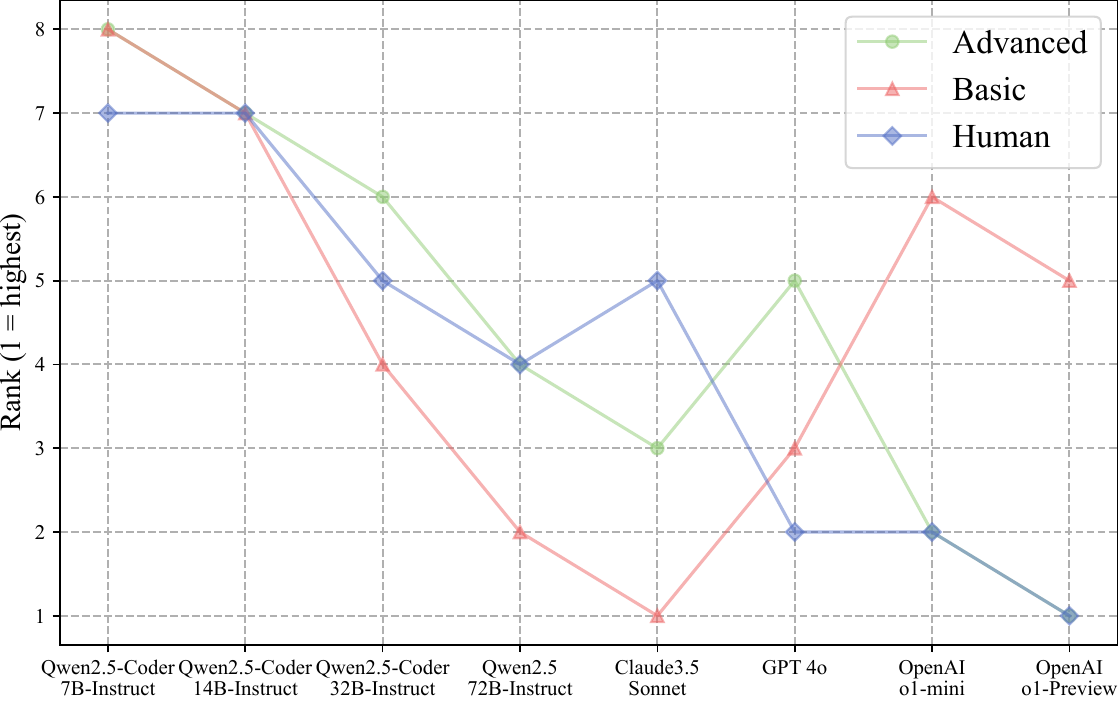}
    \caption{Comparison of ranking of model responses by three methods: basic critique, advanced critique and human evaluations.} \label{fig:further_why_fcs}
\end{figure}

\noindent \noindent\textbf{Why Advanced Critique Evaluation Is Important.} To further validate the consistency between basic critique,  advanced critique and human evaluations, we randomly sample 400 instances (named CodeCritic\_400) and test 8 models of different sizes, including both open-source and closed-source models. We then rank the models based on human, basic critique and advanced critique evaluations. As shown in Figure~\ref{fig:further_why_fcs},  rankings of the advanced setting closely match human evaluation, while results of the basic setting differ significantly. This highlights the effectiveness of our advanced evaluation approach, which provides more accurate and consistent results with human evaluations.

\noindent \textbf{Case Visualization of CodeCriticBench.} To get an intuitive understanding of data in CodeCriticBench, we present a visualization for the ``Code QA'' correct example in Figure~\ref{fig:further_vis_case_real_c}, with additional visualization results provided in Appendix~\ref{ap:case}. As observed, each sample consists of a question, an answer, multi-dimensional evaluation checklists and associated labels, which include correctness, multi-dimensional scores and final scores. Furthermore, each sample includes supplementary attributes such as partition identifiers, subset names and difficulty levels.

\begin{figure}[h]
    \centering
    \includegraphics[width=0.8\linewidth]{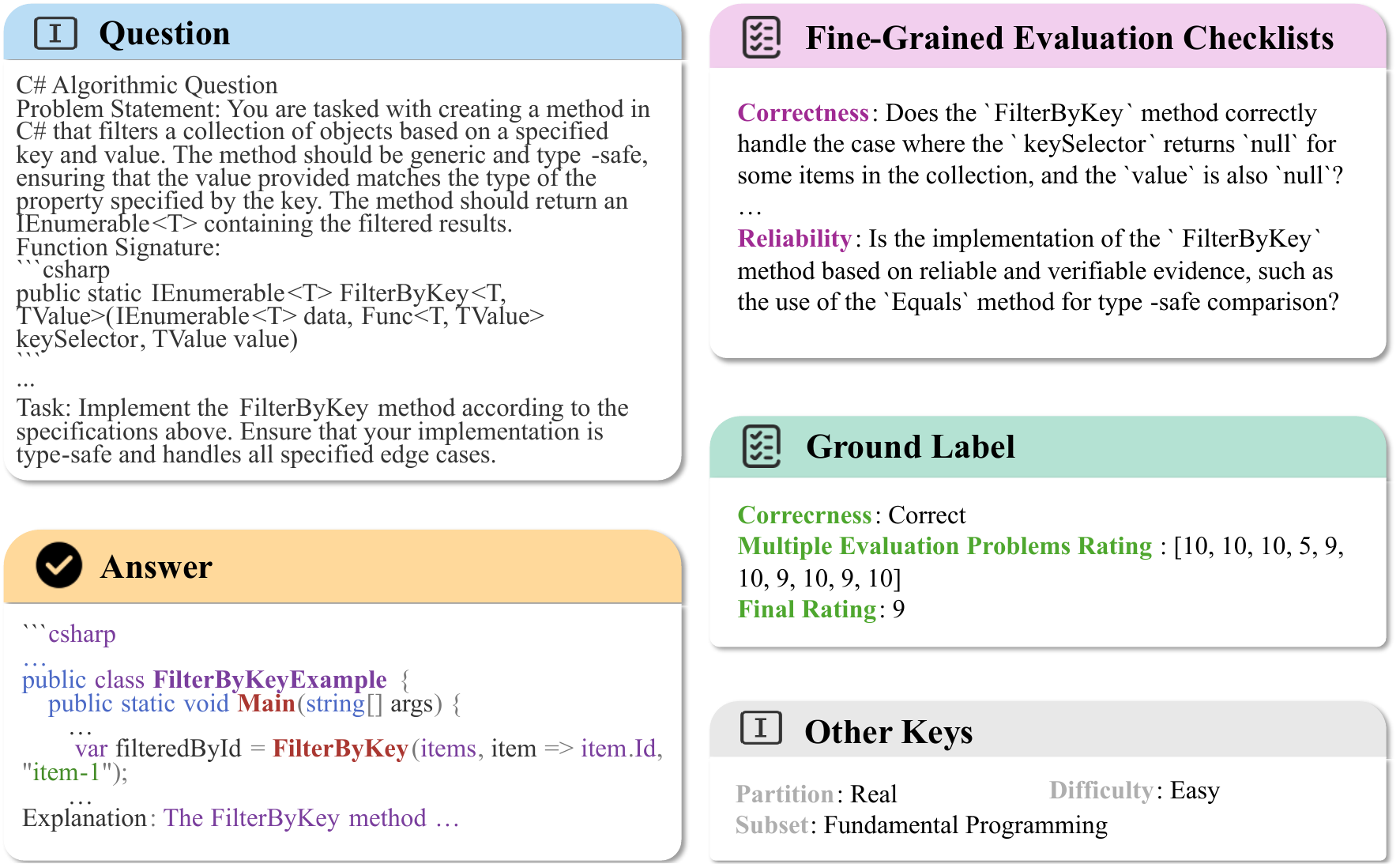}
    \caption{Example of correct case of code qa.} \label{fig:further_vis_case_real_c}
\end{figure}

% \begin{table}[h]
\begin{wraptable}{r}{3.2in}
    \vspace{-0.2in}
    \caption{The effect of CoT on model evaluation.} \label{table:further_cot}
    \centering \small
    \begin{adjustbox}{width=0.5\textwidth}
    \begin{tabular}{c|c|c|c}
        \toprule
        \textbf{Model} & \textbf{CoT} & \textbf{ACC} & \textbf{MSE} \\
        \midrule
    
        \multirow{2}{*}{\makecell{Qwen2.5-Coder-1.5B-Instruct}}
        & $\surd$ & \textbf{55.00} & 14.57 \\
        & $\times$ & 47.00 & \textbf{11.89} \\

        \midrule
        \multirow{2}{*}{\makecell{Qwen2.5-Coder-3B-Instruct}}
        & $\surd$ & \textbf{54.75} & \textbf{8.33} \\
        & $\times$ & 47.25 & 9.64\\

        \midrule
        \multirow{2}{*}{\makecell{Qwen2.5-Coder-7B-Instruct}}
        & $\surd$ & \textbf{56.50} & \textbf{4.06} \\
        & $\times$ & 49.75 & 4.31 \\

        \midrule
        \multirow{2}{*}{\makecell{Qwen2.5-Coder-14B-Instruct}}
        & $\surd$ & \textbf{66.00} & \textbf{1.99} \\
        & $\times$ & 50.50 & 5.12 \\
        \bottomrule
    \end{tabular}
    \end{adjustbox}
% \end{table}
\end{wraptable}

\noindent \textbf{Effect of CoT Evaluation.} During the basic evaluation, we observe that using a single prompt to assess answer's correctness often led to incorrect prediction. However, using fine-grained evaluation checklists, where the model scores each question individually before aggregating the results, produced more accurate outcomes. This improvement is due to the model having access to more detailed context. Based on this, we randomly select 400 instances (CodeCritic\_400) and test four models of varying sizes. The CoT prompt used in this experiment is in Appendix~\ref{ap:prompt}. As shown in Table~\ref{table:further_cot}, the use of the CoT leads to improvements in both ACC and MSE. Compared to the evaluation without CoT, Qwen2.5-Coder-3B-Instruct shows gains of 7.5\% in ACC and 1.31 in MSE, while Qwen2.5-Coder-14B-Instruct demonstrates improves of 15.5\% in ACC and 3.13 in MSE. These results clearly indicate that providing more useful information enhances model evaluation accuracy, further validating the necessity of incorporating fine-grained evaluation metrics.

\begin{figure}[h]
    \centering
    \includegraphics[width=0.6\linewidth]{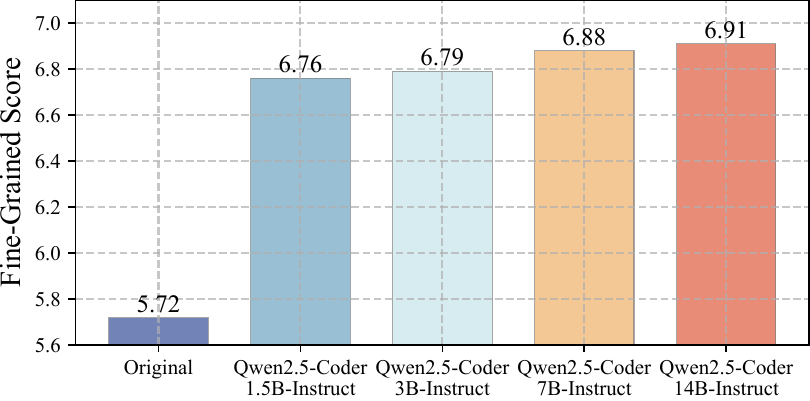}
    \caption{Scoring results of QA pairs before and after applying critiques to refine the answers.} \label{fig:further_refine}
\end{figure}

\noindent \noindent\textbf{Effect of Critique.} To evaluate the effect of model-generated critiques in improving answers, we use four Qwen2.5-Coder models (1.5B, 3B, 7B, 14B) as critics, with GPT-4o as the evaluator and CodeCritic\_400 as data. First, the critic model generates critiques for $(Q,A)$ pairs based on fine-grained evaluation checklists. The evaluator then scores original pairs. Next, the polishing model refines answers using critiques and the evaluator scores the updated answers. The critic model varies, while the evaluator and the polishing model remain the same. As shown in Figure~\ref{fig:further_refine}, applying critiques improves scores and critique quality increases with model size, aligning with the model's capabilities.

\section{Conclusion}

In this work, we introduce CodeCriticBench, a comprehensive benchmark designed to evaluate the code critique capabilities of large language models (LLMs) through two core tasks: code generation and code-based question answering (QA), each incorporating multiple levels of difficulty. Our benchmark comprises both basic and advanced evaluation metrics that target distinct aspects of LLM performance. In the basic evaluations, we assess whether the model can accurately judge the correctness of a given question-answer pair. For the advanced evaluations, we employ detailed, fine-grained checklists to facilitate a thorough assessment of the model’s critique abilities. Extensive testing of current LLMs with CodeCriticBench demonstrates its efficacy in measuring and comparing code critique performance across different models, thereby providing valuable insights for the continued refinement of foundational models.

\section*{Limitations}

While CodeCriticBench comprehensively incorporates both basic and advanced evaluations for code-related tasks—including code generation and code question answering—it is not without limitations. For example, our evaluation is confined to single-file scenarios. In future work, we plan to extend it to encompass repository-level critiques. Furthermore, as CodeCriticBench is presently focused solely on code, we intend to broaden its scope to include additional domains, thereby enabling the assessment of critique capabilities across a wider range of tasks and application scenarios.

\clearpage

\bibliography{main.bib}

\begin{thebibliography}{43}
\providecommand{\natexlab}[1]{#1}
\providecommand{\url}[1]{\texttt{#1}}
\expandafter\ifx\csname urlstyle\endcsname\relax
  \providecommand{\doi}[1]{doi: #1}\else
  \providecommand{\doi}{doi: \begingroup \urlstyle{rm}\Url}\fi

\bibitem[Achiam et~al.(2023)Achiam, Adler, Agarwal, Ahmad, Akkaya, Aleman, Almeida, Altenschmidt, Altman, Anadkat, et~al.]{achiam2023gpt}
J.~Achiam, S.~Adler, S.~Agarwal, L.~Ahmad, I.~Akkaya, F.~L. Aleman, D.~Almeida, J.~Altenschmidt, S.~Altman, S.~Anadkat, et~al.
\newblock Gpt-4 technical report.
\newblock \emph{arXiv preprint arXiv:2303.08774}, 2023.

\bibitem[Allal et~al.(2023)Allal, Li, Kocetkov, Mou, Akiki, Ferrandis, Muennighoff, Mishra, Gu, Dey, et~al.]{allal2023santacoder}
L.~B. Allal, R.~Li, D.~Kocetkov, C.~Mou, C.~Akiki, C.~M. Ferrandis, N.~Muennighoff, M.~Mishra, A.~Gu, M.~Dey, et~al.
\newblock Santacoder: don't reach for the stars!
\newblock \emph{arXiv preprint arXiv:2301.03988}, 2023.

\bibitem[Ankner et~al.(2024)Ankner, Paul, Cui, Chang, and Ammanabrolu]{ankner2024critique}
Z.~Ankner, M.~Paul, B.~Cui, J.~D. Chang, and P.~Ammanabrolu.
\newblock Critique-out-loud reward models.
\newblock \emph{arXiv preprint arXiv:2408.11791}, 2024.

\bibitem[Anthropic(2024)]{claude_2024}
Anthropic.
\newblock Introducing claude.
\newblock \url{https://www.anthropic.com/claude}, 2024.

\bibitem[Austin et~al.(2021)Austin, Odena, Nye, Bosma, Michalewski, Dohan, Jiang, Cai, Terry, Le, et~al.]{mbpp}
J.~Austin, A.~Odena, M.~Nye, M.~Bosma, H.~Michalewski, D.~Dohan, E.~Jiang, C.~Cai, M.~Terry, Q.~Le, et~al.
\newblock Program synthesis with large language models.
\newblock \emph{arXiv preprint arXiv:2108.07732}, 2021.
\newblock URL \url{https://arxiv.org/abs/2108.07732}.

\bibitem[Baars and Meester(2019)]{baars2019codearena}
S.~Baars and S.~Meester.
\newblock Codearena: Inspecting and improving code quality metrics using minecraft.
\newblock In \emph{2019 IEEE/ACM International Conference on Technical Debt (TechDebt)}, pages 68--70. IEEE, 2019.

\bibitem[Bai et~al.(2024)Bai, Liu, Bu, He, Liu, Zhou, Lin, Su, Ge, Zheng, et~al.]{bai2024mt}
G.~Bai, J.~Liu, X.~Bu, Y.~He, J.~Liu, Z.~Zhou, Z.~Lin, W.~Su, T.~Ge, B.~Zheng, et~al.
\newblock Mt-bench-101: A fine-grained benchmark for evaluating large language models in multi-turn dialogues.
\newblock \emph{arXiv preprint arXiv:2402.14762}, 2024.

\bibitem[Chen et~al.(2021)Chen, Tworek, Jun, Yuan, de~Oliveira~Pinto, Kaplan, Edwards, Burda, Joseph, Brockman, Ray, Puri, Krueger, Petrov, Khlaaf, Sastry, Mishkin, Chan, Gray, Ryder, Pavlov, Power, Kaiser, Bavarian, Winter, Tillet, Such, Cummings, Plappert, Chantzis, Barnes, Herbert-Voss, Guss, Nichol, Paino, Tezak, Tang, Babuschkin, Balaji, Jain, Saunders, Hesse, Carr, Leike, Achiam, Misra, Morikawa, Radford, Knight, Brundage, Murati, Mayer, Welinder, McGrew, Amodei, McCandlish, Sutskever, and Zaremba]{chen2021evaluatinglargelanguagemodels}
M.~Chen, J.~Tworek, H.~Jun, Q.~Yuan, H.~P. de~Oliveira~Pinto, J.~Kaplan, H.~Edwards, Y.~Burda, N.~Joseph, G.~Brockman, A.~Ray, R.~Puri, G.~Krueger, M.~Petrov, H.~Khlaaf, G.~Sastry, P.~Mishkin, B.~Chan, S.~Gray, N.~Ryder, M.~Pavlov, A.~Power, L.~Kaiser, M.~Bavarian, C.~Winter, P.~Tillet, F.~P. Such, D.~Cummings, M.~Plappert, F.~Chantzis, E.~Barnes, A.~Herbert-Voss, W.~H. Guss, A.~Nichol, A.~Paino, N.~Tezak, J.~Tang, I.~Babuschkin, S.~Balaji, S.~Jain, W.~Saunders, C.~Hesse, A.~N. Carr, J.~Leike, J.~Achiam, V.~Misra, E.~Morikawa, A.~Radford, M.~Knight, M.~Brundage, M.~Murati, K.~Mayer, P.~Welinder, B.~McGrew, D.~Amodei, S.~McCandlish, I.~Sutskever, and W.~Zaremba.
\newblock Evaluating large language models trained on code, 2021.
\newblock URL \url{https://arxiv.org/abs/2107.03374}.

\bibitem[Dubey et~al.(2024)Dubey, Jauhri, Pandey, Kadian, Al-Dahle, Letman, Mathur, Schelten, Yang, Fan, et~al.]{dubey2024llama}
A.~Dubey, A.~Jauhri, A.~Pandey, A.~Kadian, A.~Al-Dahle, A.~Letman, A.~Mathur, A.~Schelten, A.~Yang, A.~Fan, et~al.
\newblock The llama 3 herd of models.
\newblock \emph{arXiv preprint arXiv:2407.21783}, 2024.

\bibitem[Guo et~al.(2024)Guo, Zhu, Yang, Xie, Dong, Zhang, Chen, Bi, Wu, Li, et~al.]{guo2024deepseek}
D.~Guo, Q.~Zhu, D.~Yang, Z.~Xie, K.~Dong, W.~Zhang, G.~Chen, X.~Bi, Y.~Wu, Y.~Li, et~al.
\newblock Deepseek-coder: When the large language model meets programming--the rise of code intelligence.
\newblock \emph{arXiv preprint arXiv:2401.14196}, 2024.

\bibitem[Hui et~al.(2024)Hui, Yang, Cui, Yang, Liu, Zhang, Liu, Zhang, Yu, Dang, et~al.]{hui2024qwen2}
B.~Hui, J.~Yang, Z.~Cui, J.~Yang, D.~Liu, L.~Zhang, T.~Liu, J.~Zhang, B.~Yu, K.~Dang, et~al.
\newblock Qwen2. 5-coder technical report.
\newblock \emph{arXiv preprint arXiv:2409.12186}, 2024.

\bibitem[Jain et~al.(2024)Jain, Han, Gu, Li, Yan, Zhang, Wang, Solar-Lezama, Sen, and Stoica]{Jain2024LiveCodeBenchHA}
N.~Jain, K.~Han, A.~Gu, W.-D. Li, F.~Yan, T.~Zhang, S.~I. Wang, A.~Solar-Lezama, K.~Sen, and I.~Stoica.
\newblock Livecodebench: Holistic and contamination free evaluation of large language models for code.
\newblock \emph{ArXiv}, abs/2403.07974, 2024.
\newblock URL \url{https://api.semanticscholar.org/CorpusID:268379413}.

\bibitem[Ke et~al.(2024)Ke, Wen, Feng, Liu, Lei, Cheng, Wang, Zeng, Dong, Wang, et~al.]{ke2024critiquellm}
P.~Ke, B.~Wen, A.~Feng, X.~Liu, X.~Lei, J.~Cheng, S.~Wang, A.~Zeng, Y.~Dong, H.~Wang, et~al.
\newblock Critiquellm: Towards an informative critique generation model for evaluation of large language model generation.
\newblock In \emph{Proceedings of the 62nd Annual Meeting of the Association for Computational Linguistics (Volume 1: Long Papers)}, pages 13034--13054, 2024.

\bibitem[Lan et~al.(2024{\natexlab{a}})Lan, Zhang, Lyu, Li, Xu, Huang, Lin, Mao, and Chen]{lan2024training}
T.~Lan, W.~Zhang, C.~Lyu, S.~Li, C.~Xu, H.~Huang, D.~Lin, X.-L. Mao, and K.~Chen.
\newblock Training language models to critique with multi-agent feedback.
\newblock \emph{arXiv preprint arXiv:2410.15287}, 2024{\natexlab{a}}.

\bibitem[Lan et~al.(2024{\natexlab{b}})Lan, Zhang, Xu, Huang, Lin, Chen, and Mao]{lan2024criticeval}
T.~Lan, W.~Zhang, C.~Xu, H.~Huang, D.~Lin, K.~Chen, and X.-L. Mao.
\newblock Criticeval: Evaluating large-scale language model as critic.
\newblock In \emph{The Thirty-eighth Annual Conference on Neural Information Processing Systems}, 2024{\natexlab{b}}.

\bibitem[Li et~al.(2023)Li, Allal, Zi, Muennighoff, Kocetkov, Mou, Marone, Akiki, Li, Chim, et~al.]{li2023starcoder}
R.~Li, L.~B. Allal, Y.~Zi, N.~Muennighoff, D.~Kocetkov, C.~Mou, M.~Marone, C.~Akiki, J.~Li, J.~Chim, et~al.
\newblock Starcoder: may the source be with you!
\newblock \emph{arXiv preprint arXiv:2305.06161}, 2023.

\bibitem[Lin et~al.(2024)Lin, Gou, Liang, Luo, Liu, and Yang]{lin2024criticbench}
Z.~Lin, Z.~Gou, T.~Liang, R.~Luo, H.~Liu, and Y.~Yang.
\newblock Criticbench: Benchmarking llms for critique-correct reasoning, 2024.

\bibitem[Liu et~al.(2024{\natexlab{a}})Liu, Ni, Que, Sun, Wang, Yang, JiakaiWang, Guo, Peng, Zhang, Tian, Bu, Xu, Rong, Peng, and Zhang]{liu2024roleagent}
J.~Liu, Z.~Ni, H.~Que, T.~Sun, N.~Wang, J.~Yang, JiakaiWang, H.~Guo, Z.~Peng, G.~Zhang, J.~Tian, X.~Bu, K.~Xu, W.~Rong, J.~Peng, and Z.~Zhang.
\newblock Roleagent: Building, interacting, and benchmarking high-quality role-playing agents from scripts.
\newblock In \emph{The Thirty-eight Conference on Neural Information Processing Systems Datasets and Benchmarks Track}, 2024{\natexlab{a}}.
\newblock URL \url{https://openreview.net/forum?id=hORTHzt2cE}.

\bibitem[Liu et~al.(2024{\natexlab{b}})Liu, ZhiqiBai, Zhang, Zhang, YuangZh, Zhang, JiakaiWang, Que, Chen, Su, Ge, Fu, Chen, and Zheng]{liu-etal-2024-e2}
J.~Liu, Z.~ZhiqiBai, Y.~Zhang, C.~Zhang, Y.~YuangZh, G.~Zhang, J.~JiakaiWang, H.~Que, Y.~Chen, W.~Su, T.~Ge, J.~Fu, W.~Chen, and B.~Zheng.
\newblock E2-{LLM}: Efficient and extreme length extension of large language models.
\newblock In L.-W. Ku, A.~Martins, and V.~Srikumar, editors, \emph{Findings of the Association for Computational Linguistics: ACL 2024}, pages 4243--4253, Bangkok, Thailand, Aug. 2024{\natexlab{b}}. Association for Computational Linguistics.
\newblock \doi{10.18653/v1/2024.findings-acl.252}.
\newblock URL \url{https://aclanthology.org/2024.findings-acl.252/}.

\bibitem[Liu et~al.(2024{\natexlab{c}})Liu, Chai, Yang, Shi, Zhu, Wang, Jin, Zhang, Zhu, Guo, et~al.]{liu2024mdeval}
S.~Liu, L.~Chai, J.~Yang, J.~Shi, H.~Zhu, L.~Wang, K.~Jin, W.~Zhang, H.~Zhu, S.~Guo, et~al.
\newblock Mdeval: Massively multilingual code debugging.
\newblock \emph{arXiv preprint arXiv:2411.02310}, 2024{\natexlab{c}}.

\bibitem[Liu et~al.(2024{\natexlab{d}})Liu, Zhu, Liu, Xin, Li, Long, Chen, Yang, Xia, Peng, et~al.]{liu2024fullstack}
S.~Liu, H.~Zhu, J.~Liu, S.~Xin, A.~Li, R.~Long, L.~Chen, J.~Yang, J.~Xia, Z.~Peng, et~al.
\newblock Fullstack bench: Evaluating llms as full stack coder.
\newblock \emph{arXiv preprint arXiv:2412.00535}, 2024{\natexlab{d}}.

\bibitem[Lu et~al.(2021)Lu, Guo, Ren, Huang, Svyatkovskiy, Blanco, Clement, Drain, Jiang, Tang, et~al.]{lu2021codexglue}
S.~Lu, D.~Guo, S.~Ren, J.~Huang, A.~Svyatkovskiy, A.~Blanco, C.~Clement, D.~Drain, D.~Jiang, D.~Tang, et~al.
\newblock Codexglue: A machine learning benchmark dataset for code understanding and generation.
\newblock \emph{arXiv preprint arXiv:2102.04664}, 2021.

\bibitem[Luo et~al.(2023)Luo, Xu, Zhao, Sun, Geng, Hu, Tao, Ma, Lin, and Jiang]{luo2023wizardcoder}
Z.~Luo, C.~Xu, P.~Zhao, Q.~Sun, X.~Geng, W.~Hu, C.~Tao, J.~Ma, Q.~Lin, and D.~Jiang.
\newblock Wizardcoder: Empowering code large language models with evol-instruct.
\newblock \emph{arXiv preprint arXiv:2306.08568}, 2023.

\bibitem[Madaan et~al.(2023)Madaan, Tandon, Gupta, Hallinan, Gao, Wiegreffe, Alon, Dziri, Prabhumoye, Yang, Gupta, Majumder, Hermann, Welleck, Yazdanbakhsh, and Clark]{madaan2023selfrefine}
A.~Madaan, N.~Tandon, P.~Gupta, S.~Hallinan, L.~Gao, S.~Wiegreffe, U.~Alon, N.~Dziri, S.~Prabhumoye, Y.~Yang, S.~Gupta, B.~P. Majumder, K.~Hermann, S.~Welleck, A.~Yazdanbakhsh, and P.~Clark.
\newblock Self-refine: Iterative refinement with self-feedback.
\newblock In \emph{Thirty-seventh Conference on Neural Information Processing Systems}, 2023.
\newblock URL \url{https://openreview.net/forum?id=S37hOerQLB}.

\bibitem[McAleese et~al.(2024)McAleese, Pokorny, Uribe, Nitishinskaya, Trebacz, and Leike]{mcaleese2024llm}
N.~McAleese, R.~M. Pokorny, J.~F.~C. Uribe, E.~Nitishinskaya, M.~Trebacz, and J.~Leike.
\newblock Llm critics help catch llm bugs.
\newblock \emph{arXiv preprint arXiv:2407.00215}, 2024.

\bibitem[OpenAI(2024)]{gpt_4o}
OpenAI.
\newblock Hello gpt-4o.
\newblock \url{https://openai.com/index/hello-gpt-4o}, 2024.

\bibitem[Que et~al.(2024)Que, Duan, He, Mou, Zhou, Liu, Rong, Wang, Yang, Zhang, et~al.]{que2024hellobench}
H.~Que, F.~Duan, L.~He, Y.~Mou, W.~Zhou, J.~Liu, W.~Rong, Z.~M. Wang, J.~Yang, G.~Zhang, et~al.
\newblock Hellobench: Evaluating long text generation capabilities of large language models.
\newblock \emph{arXiv preprint arXiv:2409.16191}, 2024.

\bibitem[Roziere et~al.(2023)Roziere, Gehring, Gloeckle, Sootla, Gat, Tan, Adi, Liu, Sauvestre, Remez, et~al.]{roziere2023code}
B.~Roziere, J.~Gehring, F.~Gloeckle, S.~Sootla, I.~Gat, X.~E. Tan, Y.~Adi, J.~Liu, R.~Sauvestre, T.~Remez, et~al.
\newblock Code llama: Open foundation models for code.
\newblock \emph{arXiv preprint arXiv:2308.12950}, 2023.

\bibitem[Rozière et~al.(2023)Rozière, Gehring, Gloeckle, Sootla, Gat, Tan, Adi, Liu, Sauvestre, Remez, Rapin, Kozhevnikov, Evtimov, Bitton, Bhatt, Ferrer, Grattafiori, Xiong, Défossez, Copet, Azhar, Touvron, Martin, Usunier, Scialom, and Synnaeve]{rozire2023codellama}
B.~Rozière, J.~Gehring, F.~Gloeckle, S.~Sootla, I.~Gat, X.~E. Tan, Y.~Adi, J.~Liu, R.~Sauvestre, T.~Remez, J.~Rapin, A.~Kozhevnikov, I.~Evtimov, J.~Bitton, M.~Bhatt, C.~C. Ferrer, A.~Grattafiori, W.~Xiong, A.~Défossez, J.~Copet, F.~Azhar, H.~Touvron, L.~Martin, N.~Usunier, T.~Scialom, and G.~Synnaeve.
\newblock Code llama: Open foundation models for code.
\newblock \emph{arXiv preprint arXiv: 2308.12950}, 2023.

\bibitem[Saunders et~al.(2022)Saunders, Yeh, Wu, Bills, Ouyang, Ward, and Leike]{saunders2022self}
W.~Saunders, C.~Yeh, J.~Wu, S.~Bills, L.~Ouyang, J.~Ward, and J.~Leike.
\newblock Self-critiquing models for assisting human evaluators.
\newblock \emph{arXiv preprint arXiv:2206.05802}, 2022.

\bibitem[Sharma et~al.(2024)Sharma, Keh, Mitchell, Finn, Arora, and Kollar]{sharma2024critical}
A.~Sharma, S.~Keh, E.~Mitchell, C.~Finn, K.~Arora, and T.~Kollar.
\newblock A critical evaluation of ai feedback for aligning large language models.
\newblock \emph{arXiv preprint arXiv:2402.12366}, 2024.

\bibitem[Song et~al.(2025{\natexlab{a}})Song, yu~Su, Qu, Zhou, and Cheng]{prmbench}
M.~Song, Z.~yu~Su, X.~Qu, J.~Zhou, and Y.~Cheng.
\newblock Prmbench: A fine-grained and challenging benchmark for process-level reward models.
\newblock 2025{\natexlab{a}}.

\bibitem[Song et~al.(2025{\natexlab{b}})Song, Wu, Wang, Liu, Su, and Zheng]{Song2025ProgCoPH}
X.~Song, Y.~Wu, W.~Wang, J.~Liu, W.~Su, and B.~Zheng.
\newblock Progco: Program helps self-correction of large language models.
\newblock \emph{ArXiv}, abs/2501.01264, 2025{\natexlab{b}}.
\newblock URL \url{https://api.semanticscholar.org/CorpusID:275212297}.

\bibitem[Tan et~al.(2024)Tan, Zhuang, Montgomery, Tang, Cuadron, Wang, Popa, and Stoica]{judgebench}
S.~Tan, S.~Zhuang, K.~Montgomery, W.~Y. Tang, A.~Cuadron, C.~Wang, R.~A. Popa, and I.~Stoica.
\newblock Judgebench: A benchmark for evaluating llm-based judges.
\newblock \emph{ArXiv}, abs/2410.12784, 2024.

\bibitem[Tang et~al.(2025)Tang, Li, Xiao, Ding, Sun, Wang, Liu, Huang, Liu, Yu, et~al.]{tang2025real}
Z.~Tang, Z.~Li, Z.~Xiao, T.~Ding, R.~Sun, B.~Wang, D.~Liu, F.~Huang, T.~Liu, B.~Yu, et~al.
\newblock Realcritic: Towards effectiveness-driven evaluation of language model critiques.
\newblock \emph{arXiv preprint arXiv:2501.14492}, 2025.

\bibitem[Team(2024)]{dubey2024llama3}
L.~Team.
\newblock The llama 3 herd of models.
\newblock \emph{arXiv preprint arXiv: 2407.21783}, 2024.

\bibitem[Touvron et~al.(2023)Touvron, Martin, Stone, Albert, Almahairi, Babaei, Bashlykov, Batra, Bhargava, Bhosale, et~al.]{touvron2023llama}
H.~Touvron, L.~Martin, K.~Stone, P.~Albert, A.~Almahairi, Y.~Babaei, N.~Bashlykov, S.~Batra, P.~Bhargava, S.~Bhosale, et~al.
\newblock Llama 2: Open foundation and fine-tuned chat models.
\newblock \emph{arXiv preprint arXiv:2307.09288}, 2023.

\bibitem[Wang et~al.(2025)Wang, Wu, Wang, Liu, Song, Peng, Deng, Zhang, JiakaiWang, Peng, Zhang, Guo, Zhang, Su, and Zheng]{wang2025mtubench}
P.~Wang, Y.~Wu, N.~Wang, J.~Liu, X.~Song, Z.~Peng, K.~Deng, C.~Zhang, JiakaiWang, J.~Peng, G.~Zhang, H.~Guo, Z.~Zhang, W.~Su, and B.~Zheng.
\newblock {MTU}-bench: A multi-granularity tool-use benchmark for large language models.
\newblock In \emph{The Thirteenth International Conference on Learning Representations}, 2025.
\newblock URL \url{https://openreview.net/forum?id=6guG2OlXsr}.

\bibitem[Yang et~al.(2024)Yang, Yu, Zhang, Xu, Gonzalez, Cui, and Yan]{yang2024supercorrect}
L.~Yang, Z.~Yu, T.~Zhang, M.~Xu, J.~E. Gonzalez, B.~Cui, and S.~Yan.
\newblock Supercorrect: Supervising and correcting language models with error-driven insights.
\newblock \emph{arXiv preprint arXiv:2410.09008}, 2024.

\bibitem[Yue et~al.(2024)Yue, Zheng, Zhang, and Chen]{yue2024mammoth2}
X.~Yue, T.~Zheng, G.~Zhang, and W.~Chen.
\newblock Mammoth2: Scaling instructions from the web.
\newblock \emph{arXiv preprint arXiv:2405.03548}, 2024.

\bibitem[Zhang et~al.(2024)Zhang, Hosseini, Bansal, Kazemi, Kumar, and Agarwal]{zhang2024generative}
L.~Zhang, A.~Hosseini, H.~Bansal, M.~Kazemi, A.~Kumar, and R.~Agarwal.
\newblock Generative verifiers: Reward modeling as next-token prediction.
\newblock \emph{arXiv preprint arXiv:2408.15240}, 2024.

\bibitem[Zheng et~al.(2024{\natexlab{a}})Zheng, Zhang, Zhang, Lin, Lu, Yu, Liu, Zhou, and Lin]{Zheng2024ProcessBenchIP}
C.~Zheng, Z.~Zhang, B.~Zhang, R.~Lin, K.~Lu, B.~Yu, D.~Liu, J.~Zhou, and J.~Lin.
\newblock Processbench: Identifying process errors in mathematical reasoning.
\newblock \emph{ArXiv}, abs/2412.06559, 2024{\natexlab{a}}.
\newblock URL \url{https://api.semanticscholar.org/CorpusID:274598010}.

\bibitem[Zheng et~al.(2024{\natexlab{b}})Zheng, Lou, Cao, Wen, Ji, Lin, Lu, Han, Zhang, and Sun]{zheng2024critic}
X.~Zheng, J.~Lou, B.~Cao, X.~Wen, Y.~Ji, H.~Lin, Y.~Lu, X.~Han, D.~Zhang, and L.~Sun.
\newblock Critic-cot: Boosting the reasoning abilities of large language model via chain-of-thoughts critic.
\newblock \emph{arXiv preprint arXiv:2408.16326}, 2024{\natexlab{b}}.

\end{thebibliography}

\clearpage

\appendix

\section{More Experimental Results}

\subsection{Anonymous Data Link}
We have released our data on an anonymous website (\url{https://anonymous.4open.science/r/CodeCriticBench-D657/}).

\subsection{Relations Between Basic Correctness and Advanced Evaluations Scores} \label{ap:rating_level12}

\begin{figure}[h]
    \centering

    \begin{minipage}[c]{0.45\textwidth}
        \centering
        \includegraphics[width=\textwidth]{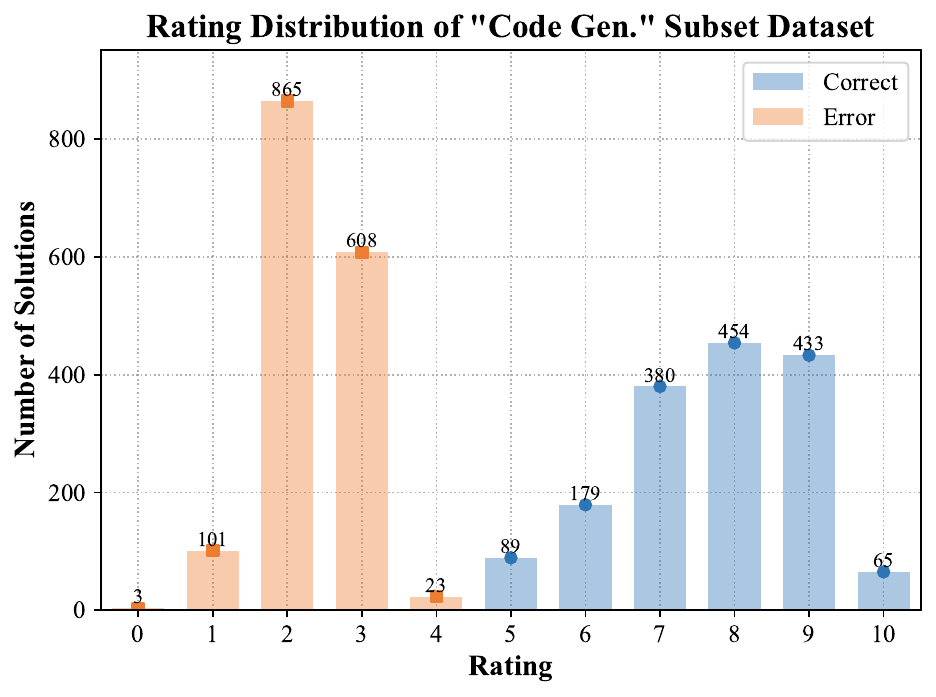}
        \subcaption{The Code Gen subset.}
        \label{fig:further_algo_level12_rating_distribution}
    \end{minipage}
    \begin{minipage}[c]{0.45\textwidth}
        \centering
        \includegraphics[width=\textwidth]{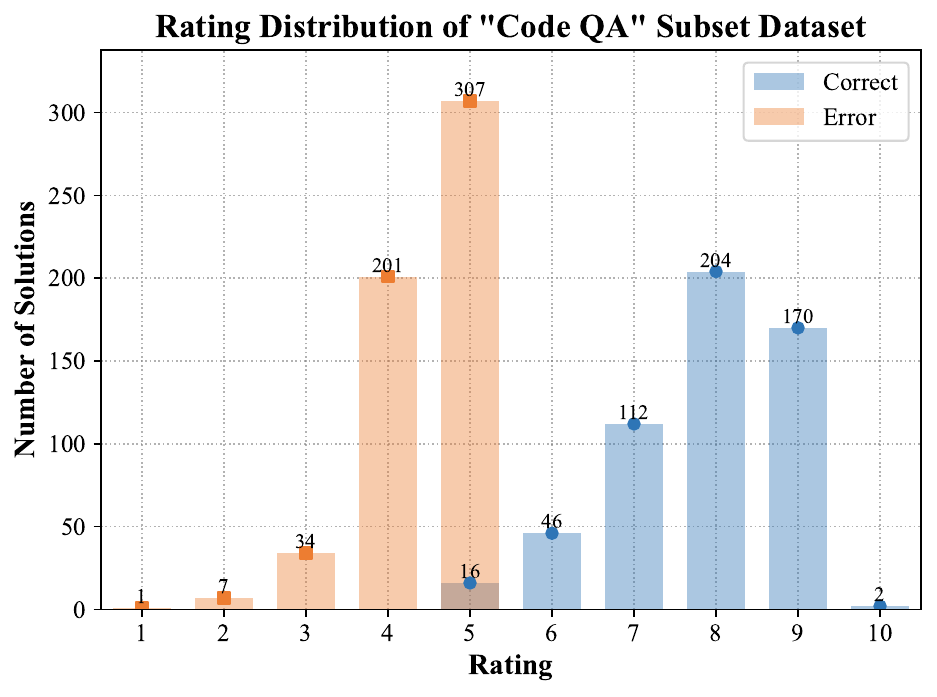}
        \subcaption{The Code QA subset.}
        \label{fig:further_real_level12_rating_distribution}
    \end{minipage}
    \caption{Rating distribution.}
\end{figure}

We present the fine-grained score distributions for both the ``Code Gen'' and ``Code QA'' subsets. As illustrated in Figures~\ref{fig:further_algo_level12_rating_distribution} and \ref{fig:further_real_level12_rating_distribution}, these distributions exhibit favorable characteristics, thereby affirming the validity of our dataset.

\subsection{Case Visualization of CodeCriticBench} \label{ap:case}

\begin{figure}[h!]
    \centering
    \includegraphics[width=0.6\linewidth]{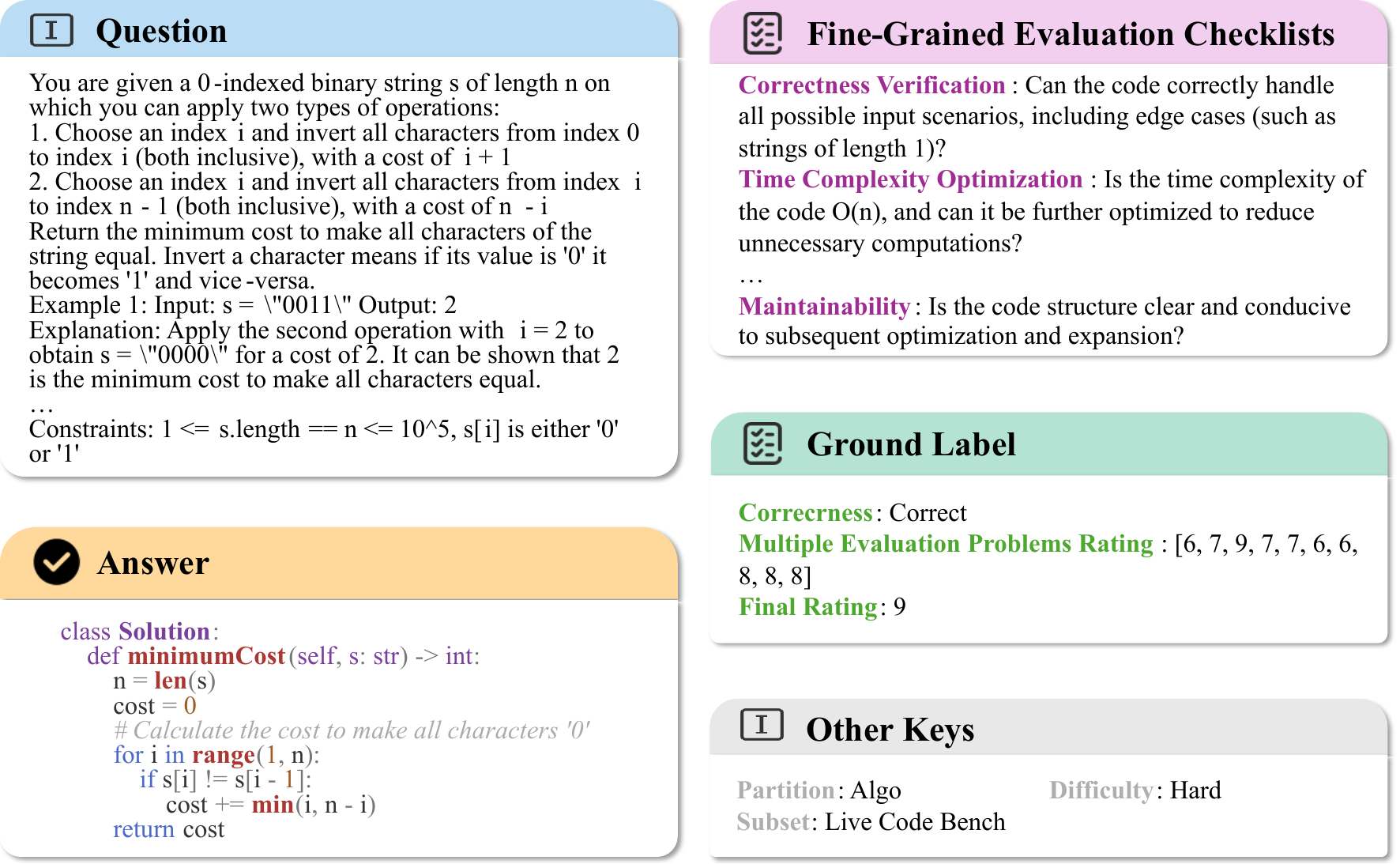}
    \caption{Example of correct case of code generation.} \label{fig:further_vis_case_algo_c}
\end{figure}

\begin{figure}[h!]
    \centering
    \includegraphics[width=0.6\linewidth]{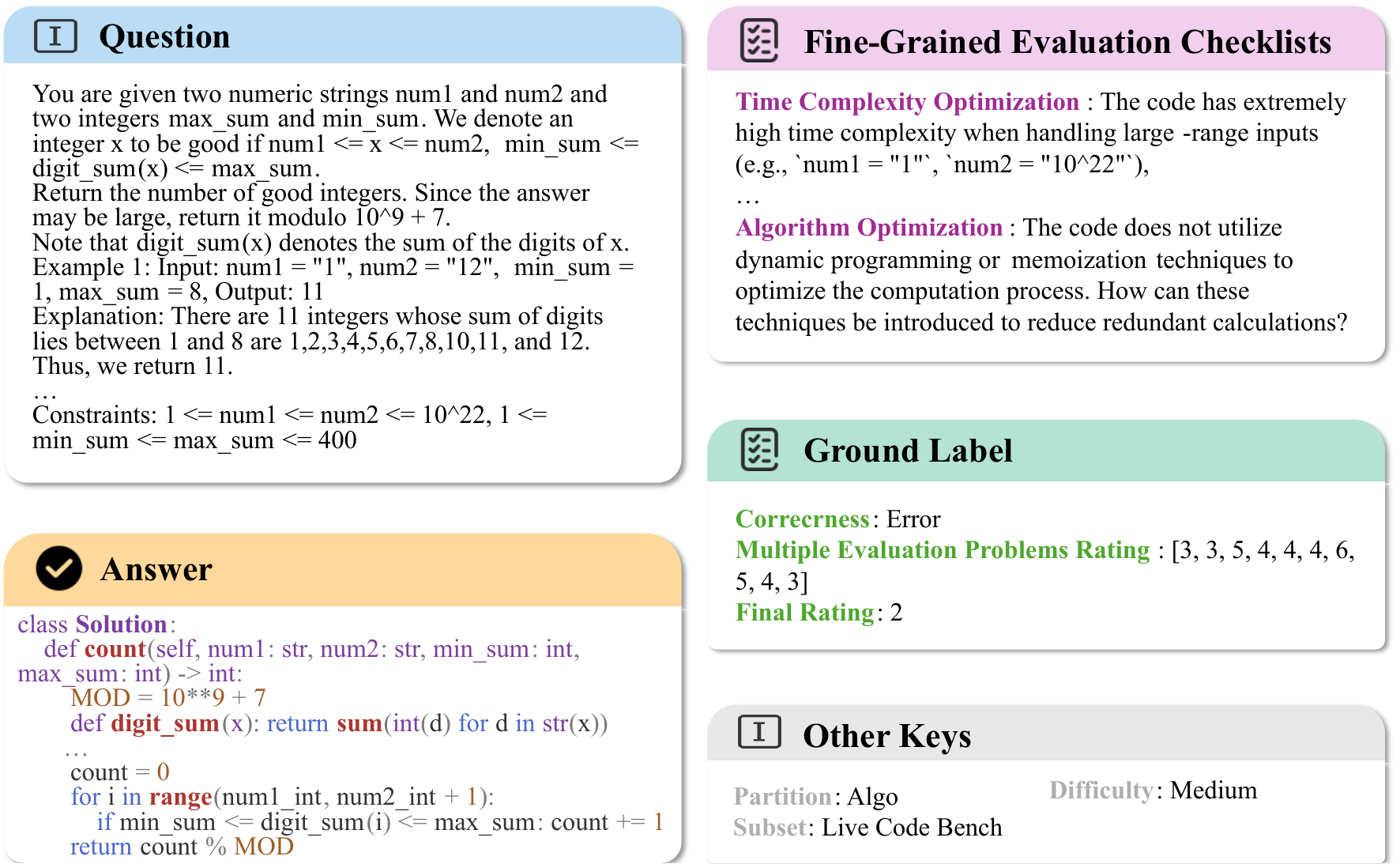}
    \caption{Example of error case of code generation.} \label{fig:further_vis_case_algo_e}
\end{figure}

\begin{figure}[h!]
    \centering
    \includegraphics[width=0.6\linewidth]{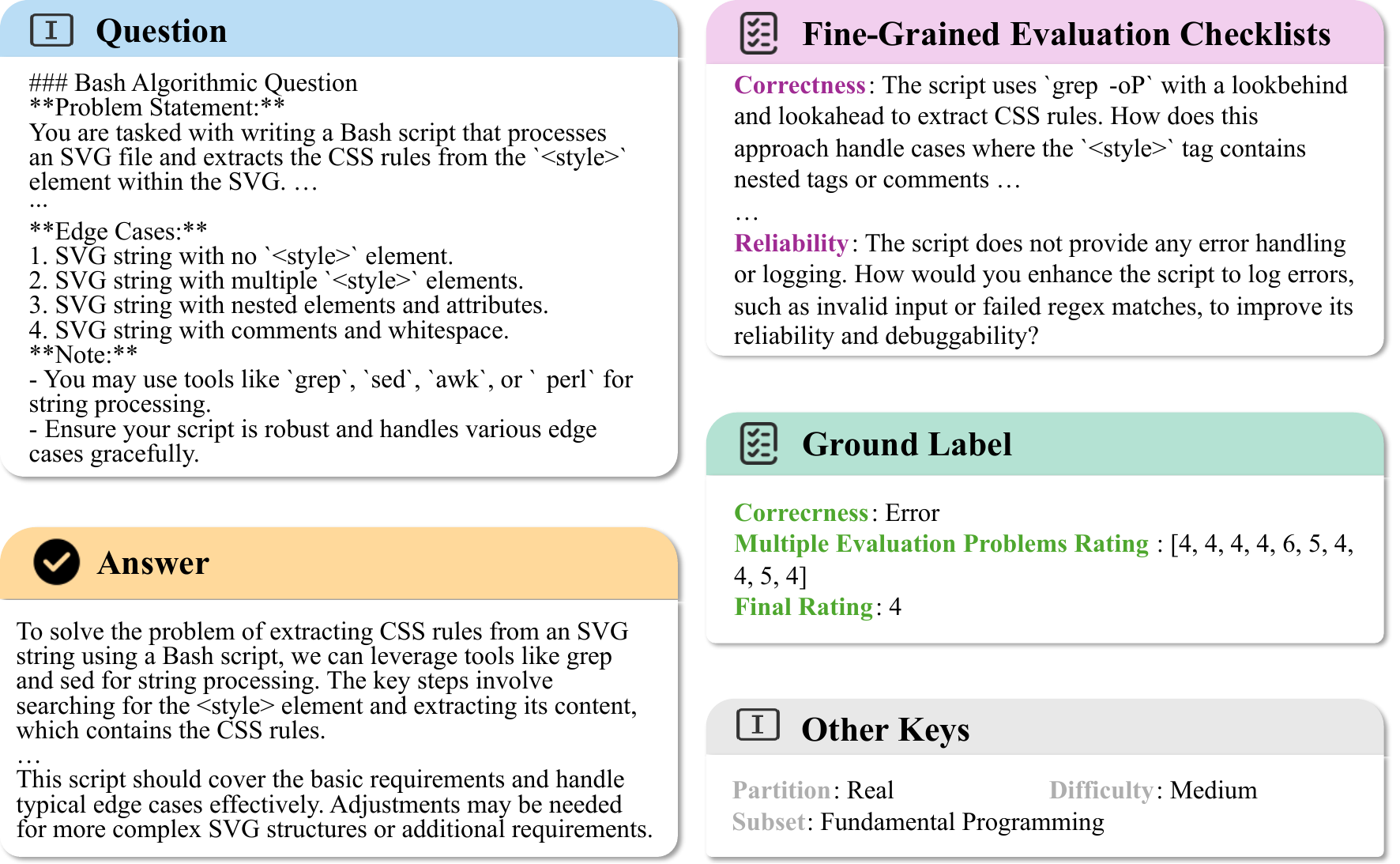}
    \caption{Example of error case of code qa.} \label{fig:further_vis_case_real_e}
\end{figure}

We present remaining case visualization. Specifically, Figures~\ref{fig:further_vis_case_real_c}- \ref{fig:further_vis_case_real_e} illustrate a correct example from the ``Code Gen'' subset, an error example from the ``Code Gen'' subset and an error example from the ``Code QA'' subset, respectively.

\begin{figure*}[t]
    \centering
    \includegraphics[width=\linewidth]{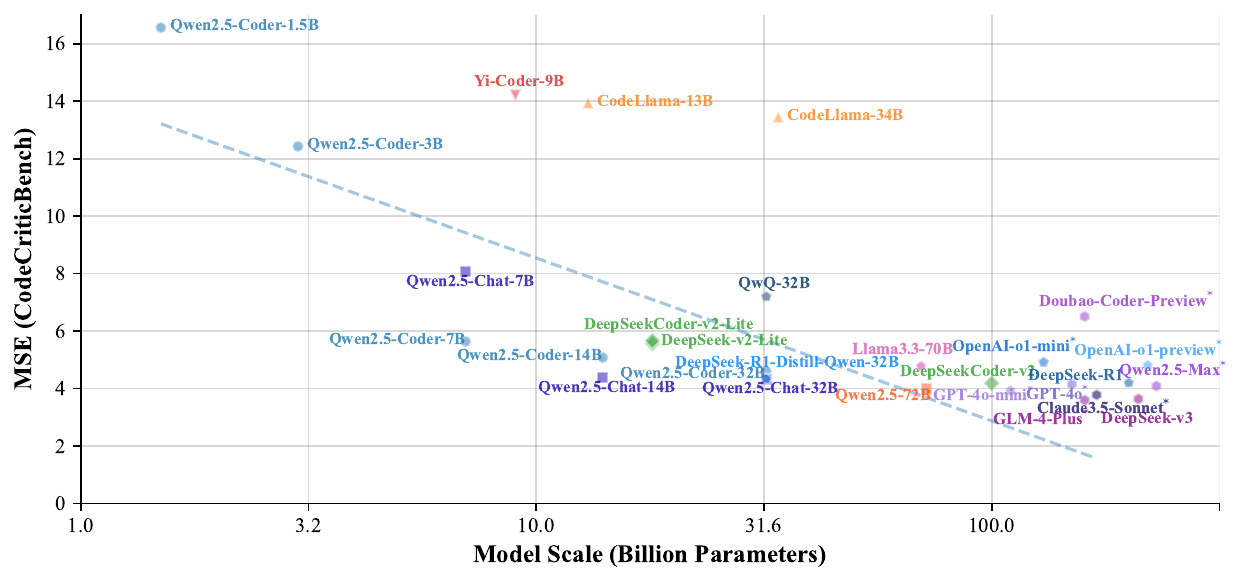}
    \caption{Scaling law on advanced critique evaluation (MSE) across models. ``*'' indicates an estimated parameter size.} \label{fig:further_scaling_mse}
\end{figure*}

\begin{figure*}[t]
    \centering
    \includegraphics[width=\linewidth]{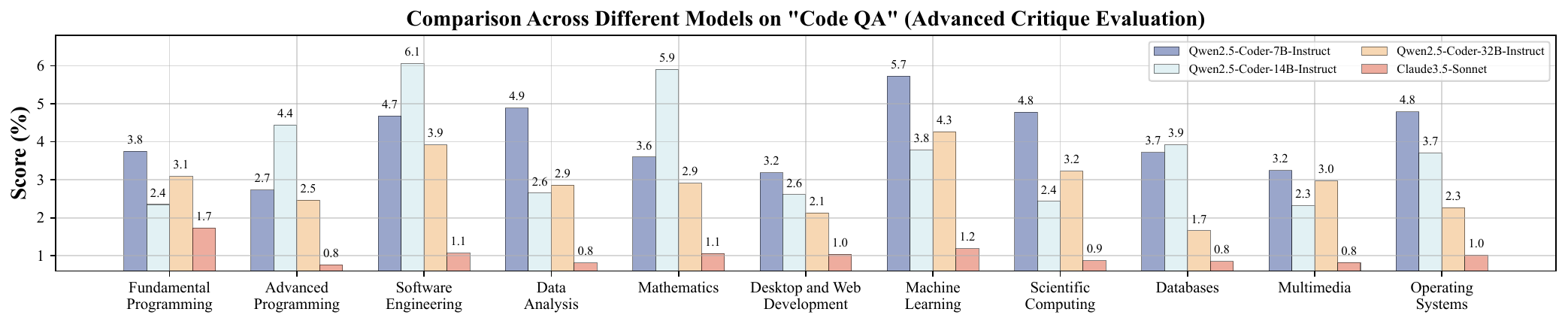}
    \caption{Comparison across different models on ``Code QA'' (Advanced Critique Evaluation).} \label{fig:further_app_mse}
\end{figure*}

\subsection{Scaling Law} \label{ap:scaling}

We present the scaling law results for the MSE in advanced critique evaluation. As depicted in Figure~\ref{fig:further_scaling_mse}, an increase in model parameters is accompanied by a steady decrease in MSE, which aligns well with our expectations and further validates the rationality of our dataset.

\subsection{Different Application Scenes} \label{ap:app}

We present a comparative analysis of the advanced evaluation MSE across various models and application scenarios within the ``Code QA'' subset. As depicted in Figure~\ref{fig:further_app_mse}, an increase in model parameters is generally accompanied by a steady decrease in MSE. Notably, Claude3.5-Sonnet consistently achieves lower MSE across nearly all scenarios, which may be attributed to its specialized optimizations for code and dialogue contexts.

\subsection{Bug Identification} \label{ap:bug}

Table~\ref{table:bug_category} presents the performance metrics of evaluated models across the 23 distinct error categories comprising our Debug evaluation subset. These categories span critical software integrity domains, including but not limited to ``Input Validation and Data Processing Error'', ``Security Vulnerabilities'' and ``Reference Error'', providing a comprehensive assessment of model prediction's robustness in practical debugging scenarios.

% bug_category
% 错误类型: Input Validation and Data Processing Error、Security Vulnerability、Reference Error、Code Quality and Maintenance Error、Testing and Verification Error、User Permission and Authentication Error、Type Error、Performance Issue
\begin{table*}[t]
    \centering
    \caption{The accuracy of different models in identifying programming error types.} \label{table:bug_category}
    \begin{adjustbox}{width=\textwidth}
    \begin{tabular}{l|ccccccccccccc}
        \toprule
        \multirow{3}{*}{\textbf{Error Category}} & \multicolumn{8}{c}{\textbf{Open-Sourced}} & \multicolumn{5}{c}{\textbf{Close-Sourced}} \\
        \cmidrule(lr){2-9} \cmidrule(lr){10-14}
        
        & \textbf{\makecell{Qwen2.5-Coder\\0.5B-Instruct}} 
        & \textbf{\makecell{Qwen2.5-Coder\\1.5B-Instruct}} 
        & \textbf{\makecell{Qwen2.5-Coder\\3B-Instruct}} 
        & \textbf{\makecell{Qwen2.5-Coder\\7B-Instruct}} 
        & \textbf{\makecell{Qwen2.5-Chat\\7B-Instruct}} 
        & \textbf{\makecell{Qwen2.5-Coder\\14B-Instruct}} 
        & \textbf{\makecell{Qwen2.5-Chat\\14B-Instruct}} 
        & \textbf{\makecell{Qwen2.5\\72B-Instruct}} 
        & \textbf{GPT 4o} 
        & \textbf{\makecell{Claude3.5\\Sonnet}} 
        & \textbf{\makecell{Doubao-Coder\\Preview}} 
        & \textbf{\makecell{GLM-4\\Plus}} 
        & \textbf{\makecell{Qwen2.5\\Max}} \\
        
        \midrule
        All & 21.50 & 32.75 & 48.00 & 61.00 & 41.00 & 62.25 & 47.50 & 61.25 & 61.00 & 54.00 & 67.00 & 47.75 & 59.50 \\
        Configuration Management Error & 50.57 & 3.45 & 1.15 & 2.30 & 3.45 & 3.45 & 1.15 & 0.00 & 1.15 & 0.00 & 3.45 & 0.00 & 3.45 \\
        Data Management Error & 40.00 & 5.45 & 10.91 & 7.27 & 3.64 & 3.64 & 7.27 & 10.91 & 0.00 & 3.64 & 1.82 & 12.73 & 7.27 \\
        Input Validation and Data Processing Error & 7.41 & 31.48 & 42.59 & 27.78 & 37.04 & 46.30 & 20.37 & 50.00 & 37.04 & 33.33 & 22.22 & 9.26 & 37.04 \\
        Monitoring and Logging Management Error & 1.85 & 0.00 & 0.00 & 1.85 & 0.00 & 0.00 & 0.00 & 0.00 & 0.00 & 5.56 & 5.56 & 1.85 & 1.85 \\
        Environment Variable Error & 0.00 & 0.00 & 20.00 & 20.00 & 40.00 & 30.00 & 30.00 & 0.00 & 40.00 & 10.00 & 10.00 & 10.00 & 20.00 \\
        Dependency Management Error & 0.00 & 0.00 & 25.00 & 29.17 & 4.17 & 33.33 & 33.33 & 16.67 & 25.00 & 16.67 & 25.00 & 20.83 & 25.00 \\
        Syntax Error & 19.64 & 67.86 & 37.50 & 25.00 & 28.57 & 51.79 & 30.36 & 39.29 & 21.43 & 14.29 & 41.07 & 23.21 & 26.79 \\
        Design Flaw & 2.63 & 0.00 & 13.16 & 10.53 & 2.63 & 2.63 & 18.42 & 7.89 & 0.00 & 18.42 & 7.89 & 10.53 & 5.26 \\
        Security Vulnerability & 12.73 & 25.45 & 23.64 & 27.27 & 21.82 & 41.82 & 27.27 & 38.18 & 45.45 & 47.27 & 47.27 & 52.73 & 40.00 \\
        Log Security Issue & 0.00 & 0.00 & 0.00 & 0.00 & 0.00 & 0.00 & 11.11 & 0.00 & 0.00 & 0.00 & 11.11 & 11.11 & 0.00 \\
        Reference Error & 3.75 & 2.50 & 8.75 & 41.25 & 20.00 & 40.00 & 38.75 & 40.00 & 52.50 & 41.25 & 48.75 & 38.75 & 45.00 \\
        Session Management Error & 0.00 & 14.29 & 14.29 & 33.33 & 14.29 & 23.81 & 14.29 & 14.29 & 4.76 & 9.52 & 14.29 & 9.52 & 19.05 \\
        Code Quality and Maintenance Error & 2.04 & 0.00 & 0.00 & 4.08 & 0.00 & 0.00 & 4.08 & 8.16 & 8.16 & 12.24 & 14.29 & 2.04 & 6.12 \\
        Logic Error & 1.35 & 24.32 & 29.73 & 31.08 & 40.54 & 27.03 & 25.68 & 31.08 & 36.49 & 22.97 & 36.49 & 18.92 & 31.08 \\
        Testing and Verification Error & 3.51 & 0.00 & 0.00 & 5.26 & 0.00 & 3.51 & 0.00 & 3.51 & 5.26 & 5.26 & 15.79 & 1.75 & 3.51 \\
        Network and Communication Error & 0.00 & 14.75 & 9.84 & 14.75 & 4.92 & 9.84 & 0.00 & 3.28 & 6.56 & 3.28 & 8.20 & 4.92 & 6.56 \\
        Exception Handling Error & 2.38 & 4.76 & 4.76 & 4.76 & 0.00 & 0.00 & 7.14 & 9.52 & 2.38 & 4.76 & 4.76 & 2.38 & 4.76 \\
        User Permission and Authentication Error & 0.00 & 6.25 & 4.69 & 15.62 & 7.81 & 14.06 & 12.50 & 15.62 & 17.19 & 15.62 & 21.88 & 14.06 & 15.62 \\
        File and I/O Error & 0.00 & 10.96 & 24.66 & 31.51 & 17.81 & 31.51 & 20.55 & 20.55 & 32.88 & 1.37 & 26.03 & 24.66 & 32.88 \\
        Type Error & 0.00 & 21.43 & 33.33 & 38.10 & 7.14 & 23.81 & 33.33 & 45.24 & 38.10 & 26.19 & 26.19 & 30.95 & 30.95 \\
        Internationalization and Localization Error & 1.47 & 1.47 & 5.88 & 14.71 & 2.94 & 8.82 & 7.35 & 4.41 & 7.35 & 7.35 & 10.29 & 7.35 & 7.35 \\
        Performance Issue & 0.00 & 2.30 & 14.94 & 12.64 & 1.15 & 16.09 & 11.49 & 19.54 & 13.79 & 29.89 & 27.59 & 12.64 & 16.09 \\
        Concurrency and Multithreading Error & 0.00 & 23.21 & 57.14 & 85.71 & 60.71 & 76.79 & 41.07 & 80.36 & 69.64 & 75.00 & 66.07 & 55.36 & 64.29 \\
        \bottomrule
    \end{tabular}
    \end{adjustbox}
\end{table*}

\subsection{Critique Evaluation on ``Code Gen'' and ``Code QA'' Subset} \label{ap:main_subset}

We present more fine-grained scores for all the models used in our experiments across different scenarios. Tables~\ref{table:algo_level1} and \ref{table:algo_level2} display the ACC and MSE of basic evaluation for each subset under the ``Code Gen'' subset. Tables~\ref{table:real_level1} and \ref{table:real_level2} show the ACC and MSE of basic evaluation for each application scenario within the ``Code QA'' subset. 

Note that we have adopted the following abbreviations in our tables to denote various application scenarios: Fundamental Programming (FP), Advanced Programming (AP), Software Engineering (SE), Data Analysis (DA), Mathematics (MA), Desktop and Web Development (DW), Machine Learning (ML), Scientific Computing (SC), Databases (DS), Multimedia (MM) and Operating Systems (OS).

\subsection{Critique Evaluation on Multiple Fine-Grained Dimensions} \label{ap:main_dimension}

Our Code Gen subset incorporates 10 fine-grained evaluation dimensions, including [``Correctness Verification'', ``Time Complexity Optimization'', ``Space Complexity'', ``Code Readability'', ``Robustness Validation'', ``Algorithm Optimization'', ``Comprehensive Testing'', ``Output Format'', ``Code Style Consistency'', ``Maintainability''], while the Code QA subset is assessed across 10 distinct dimensions, including [``Correctness'', ``Completeness'', ``Performance'', ``Maintainability'', ``Clarity'', ``Depth'', ``Practicality'', ``Logic Coherence'', ``Innovation'', ``Reliability'']. 

Given that our advanced evaluation includes multiple evaluation dimensions, we further measure the MSE results across these dimensions for both the ``Code Gen'' subset and ``Code QA'' subset of the dataset. Table~\ref{table:algo_level2_all_dim} presents the results for all evaluation dimensions within the ``Code Gen'' subset. Tables~\ref{table:algo_level2_mbpp_dim}-\ref{table:algo_level2_debug_dim} show the results for each specific evaluation dimension across different subsets. Table~\ref{table:real_level2_all_dim} displays the results for all evaluation dimensions within the ``Code QA'' subset, while Tables~\ref{table:real_level2_fp_dim}-\ref{table:real_level2_os_dim} provide the results for each evaluation dimension within various specific application scenarios.

\section{Prompt} \label{ap:prompt}

We present all the evaluation prompts utilized in our experiment. Specifically, Figure~\ref{ap:prompt_level1} displays the prompt for basic evaluation ACC, Figure~\ref{ap:prompt_level2} presents the prompt for advanced evaluation MSE and Figure~\ref{ap:prompt_bug} shows the prompt for assessing the model's ability to identify the corresponding code error type. Figures~\ref{ap:prompt_cot_level1} and \ref{ap:prompt_cot_level2} illustrate the prompts for basic and advanced evaluations using the CoT method, respectively. Additionally, Figure~\ref{ap:prompt_refine} provides the prompt used for polishing and modifying the answer based on feedback, which is employed to verify the effectiveness of the critique.

To enhance the transparency and credibility of our data construction process, we have made public the specific prompts used to create the Debug dataset subset, as well as the prompts employed for generating customized, fine-grained evaluation questions for each question-answer pair $(Q, A)$ in the advanced evaluation phase. In particular, Figure~\ref{ap:prompt_error_typelist} illustrates the types of programming errors we identified and categorized, encompassing 23 major categories and over 110 subcategories. Figure~\ref{ap:prompt_insert_bug} presents the corresponding prompts used to insert these programming errors. The detailed classification within our Code QA scenario was also generated through prompts utilizing large language models (LLMs), as shown in Figure~\ref{ap:prompt_real_classify}. Lastly, Figures~\ref{ap:prompt_algo_diy_qa} and~\ref{ap:prompt_real_diy_qa} display the prompts for generating customized, fine-grained evaluation questions for each question-answer pair $(Q, A)$ in the Code Generation and Code QA scenarios.

\section{Model Lists} \label{ap:model_lists}

Our experiments utilizes a diverse set of inference models, spanning both closed-source and open-source architectures and covering a broad spectrum of sizes. Specifically, Table~\ref{table:open_source_model} details the open-source models employed in our study, including those from the Qwen series (Qwen2.5-Coder/Chat, QwQ, CodeQwen), the LLaMA series (CodeLlama, Llama), the DeepSeek series (DeepSeek-Coder/Chat), as well as the OpenCoder, Yi-Coder series and StarCoder2. These models range in size from 0.5B to over 200B, ensuring extensive coverage across different scales. Table~\ref{table:api_model}, on the other hand, presents the closed-source models used, which encompass some of the most powerful models available to date, such as Claude3.5-Sonnet, GPT 4o-mini, GPT 4o, Gemini2.0-Flash-Thinking, DeepSeek-v3, GLM-4-Plus, Qwen-Max and the current popular o1-like models, including OpenAI o1-Preview, OpenAI o1-mini and DeepSeek-R1.

\begin{table*}[h!]
    \small \centering
    \caption{Open-sourced models adopted in our experiments.} \label{table:open_source_model}
    \begin{adjustbox}{width=0.8\textwidth}
\begin{tabular}{l|l}
        \toprule
        \textbf{Open-Sourced Model} & \textbf{Model Link} \\
        \midrule
        CodeQwen1.5-7B-Instruct & \url{https://hf.co/Qwen/CodeQwen1.5-7B-Instruct} \\
        Qwen2.5-Coder-0.5B-Instruct & \url{https://hf.co/Qwen/Qwen2.5-Coder-0.5B-Instruct} \\
        Qwen2.5-Coder-1.5B-Instruct & \url{https://hf.co/Qwen/Qwen2.5-Coder-1.5B-Instruct} \\
        Qwen2.5-Coder-3B-Instruct & \url{https://hf.co/Qwen/Qwen2.5-Coder-3B-Instruct} \\
        Qwen2.5-Coder-7B-Instruct & \url{https://hf.co/Qwen/Qwen2.5-Coder-7B-Instruct} \\
        Qwen2.5-Coder-14B-Instruct & \url{https://hf.co/Qwen/Qwen2.5-Coder-14B-Instruct} \\
        Qwen2.5-Coder-32B-Instruct & \url{https://hf.co/Qwen/Qwen2.5-Coder-32B-Instruct} \\
\bottomrule
    \end{tabular}
    \end{adjustbox}
\end{table*}

\begin{table*}[h!]
    \small \centering
    \caption{Close-sourced models (APIs) adopted in our experiments.} \label{table:api_model} 
    \begin{adjustbox}{width=0.8\textwidth}
    \begin{tabular}{l|l}
        \toprule
        \textbf{Close-Sourced Model} & \textbf{API Entry} \\
        \midrule
        Claude-3.5-Sonnet & \url{https://www.anthropic.com/news/claude-3-5-sonnet} \\
        OpenAI o1-Preview & \url{https://platform.openai.com/docs/models\#o1} \\
        OpenAI o1-mini & \url{https://platform.openai.com/docs/models\#o1} \\
        GPT 4o-mini & \url{https://platform.openai.com/docs/models#gpt-4o-mini} \\
        GPT 4o & 
        \url{https://platform.openai.com/docs/models\#gpt-4o} \\
        Gemini2.0-Flash-Thinking & \url{https://ai.google.dev/gemini-api/docs/thinking} \\
        Doubao-Coder-Preview & \url{https://www.volcengine.com/product/doubao} \\
        DeepSeek-v3 & \url{https://www.deepseek.com} \\
        DeepSeek-R1 & \url{https://www.deepseek.com} \\
        GLM-4-Plus & \url{https://open.bigmodel.cn/dev/api/normal-model/glm-4} \\
        Qwen-Max & \url{https://www.aliyun.com/product/bailian}\\
        \bottomrule
    \end{tabular}
    \end{adjustbox}
\end{table*}

% algo_level1
\begin{table*}[h!]
    \centering
    \caption{Results of different models on basic critique evaluations ACC in the Code Gen Subset Dataset.} \label{table:algo_level1}
    \begin{adjustbox}{width=0.8\textwidth}
    \begin{tabular}{c|cccccc}
    
    \toprule
    \textbf{Model} & \textbf{All} & \textbf{MBPP} & \textbf{CodeForce} & \textbf{LiveCodeBench} & \textbf{Debug} & \\

    \midrule
    \multicolumn{6}{c}{0.5B+ Instruction Tuned Model} \\
    \midrule
    Qwen2.5-Coder-0.5B-Instruct & 51.28 & 50.88 & 50.12 & 52.88 & 51.25 \\
    
    \midrule
    \multicolumn{6}{c}{1B+ Instruction Tuned Model} \\
    \midrule
    Yi-Coder-1.5B-Chat & 51.56 & 61.75 & 51.50 & 42.75 & 50.25 \\
    DeepSeek-Coder-1.3B-Instruct & 52.31 & 54.00 & 52.75 & 51.38 & 51.12 \\
    OpenCoder-1.5B-Instruct & 53.06 & 57.38 & 51.62 & 53.87 & 49.38 \\
    Qwen2.5-Coder-1.5B-Instruct & 53.87 & 56.25 & 50.62 & 50.50 & 58.13 \\
\bottomrule
    \end{tabular}
    \end{adjustbox}
\end{table*}

% algo_level2
\begin{table*}[h!]
    \centering
    \caption{Results of different models on advanced critique evaluations MSE in the Code Gen Subset Dataset.} \label{table:algo_level2}
    \begin{adjustbox}{width=0.8\textwidth}
    \begin{tabular}{c|cccccc}
    
    \toprule
    \textbf{Model} & \textbf{All} & \textbf{MBPP} & \textbf{CodeForce} & \textbf{LiveCodeBench} & \textbf{Debug} & \\

    \midrule
    \multicolumn{6}{c}{0.5B+ Instruction Tuned Model} \\
    \midrule
    Qwen2.5-Coder-0.5B-Instruct & 23.96 & 20.75 & 24.77 & 27.18 & 23.13 \\
    
    \midrule
    \multicolumn{6}{c}{1B+ Instruction Tuned Model} \\
    \midrule
    Yi-Coder-1.5B-Chat & 28.27 & 27.28 & 26.58 & 31.29 & 27.94 \\
    OpenCoder-1.5B-Instruct & 28.07 & 30.73 & 29.95 & 25.25 & 26.35 \\
    DeepSeek-Coder-1.3B-Instruct & 24.48 & 21.12 & 23.93 & 28.95 & 23.94 \\
    Qwen2.5-Coder-1.5B-Instruct & 17.13 & 14.52 & 18.26 & 18.58 & 17.14 \\
\bottomrule
    \end{tabular}
    \end{adjustbox}
\end{table*}

% real_level1
\begin{table*}[h!]
    \centering
    \caption{Results of different models on basic critique evaluations ACC in the Code QA Subset Dataset.} \label{table:real_level1}
    \begin{adjustbox}{width=\textwidth}
    \begin{tabular}{c|c|cccccccccccc}
    
    \toprule
    \textbf{Size} & \textbf{Model} & \textbf{All} & \textbf{FP} & \textbf{AP} & \textbf{SE} & \textbf{DA} & \textbf{MA} & \textbf{DW} & \textbf{ML} & \textbf{SC} & \textbf{DB} & \textbf{MM} & \textbf{OS} \\
    
    \midrule
    0.5B+ & Qwen2.5-Coder-0.5B-Instruct & 53.27 & 53.00 & 46.00 & 52.00 & 52.00 & 53.00 & 49.00 & 69.00 & 59.00 & 50.00 & 55.00 & 48.00 \\

    \midrule
    \multirow{5}{*}{1B+} 
    & DeepSeek-Coder-1.3B-Instruct & 48.91 & 50.00 & 47.00 & 46.00 & 51.00 & 46.00 & 47.00 & 53.00 & 48.00 & 51.00 & 52.00 & 47.00 \\
    & Yi-Coder-1.5B-Chat & 49.27 & 48.00 & 55.00 & 43.00 & 61.00 & 44.00 & 47.00 & 44.00 & 44.00 & 44.00 & 61.00 & 51.00 \\
    & OpenCoder-1.5B-Instruct & 53.45 & 62.00 & 57.00 & 55.00 & 58.00 & 47.00 & 42.00 & 58.00 & 55.00 & 49.00 & 52.00 & 53.00 \\
    & Qwen2.5-Coder-3B-Instruct & 55.18 & 47.00 & 52.00 & 57.00 & 58.00 & 53.00 & 50.00 & 75.00 & 57.00 & 49.00 & 59.00 & 50.00 \\
    & Qwen2.5-Coder-1.5B-Instruct & 56.18 & 53.00 & 56.00 & 62.00 & 62.00 & 51.00 & 47.00 & 68.00 & 60.00 & 43.00 & 62.00 & 54.00 \\
    
    \midrule
    \multirow{6}{*}{6B+} 
    & OpenCoder-8B-Instruct & 52.64 & 48.00 & 48.00 & 50.00 & 54.00 & 46.00 & 48.00 & 75.00 & 51.00 & 52.00 & 57.00 & 50.00 \\
\bottomrule
    \end{tabular}
    \end{adjustbox}
\end{table*}

% real_level2
\begin{table*}[h!]
    \centering
    \caption{Results of different models on advanced critique evaluations MSE in the Code QA Subset Dataset.} \label{table:real_level2}
    \begin{adjustbox}{width=\textwidth}
    \begin{tabular}{c|c|cccccccccccc}
    
    \toprule
    \textbf{Size} & \textbf{Model} & \textbf{All} & \textbf{FP} & \textbf{AP} & \textbf{SE} & \textbf{DA} & \textbf{MA} & \textbf{DW} & \textbf{ML} & \textbf{SC} & \textbf{DB} & \textbf{MM} & \textbf{OS} \\
    
    \midrule
    0.5B+ & Qwen2.5-Coder-0.5B-Instruct & 25.06 & 25.92 & 24.92 & 24.91 & 29.49 & 25.20 & 25.52 & 18.71 & 28.84 & 24.07 & 23.42 & 24.64 \\
    
    \midrule
    \multirow{5}{*}{1B+} 
    & Yi-Coder-1.5B-Chat & 32.13 & 31.45 & 28.35 & 32.74 & 37.67 & 33.47 & 34.50 & 25.28 & 34.15 & 33.39 & 31.86 & 30.59 \\
    & OpenCoder-1.5B-Instruct & 26.23 & 26.21 & 24.05 & 27.31 & 29.19 & 28.31 & 24.83 & 20.11 & 34.02 & 22.59 & 25.08 & 26.87 \\
    & DeepSeek-Coder-1.3B-Instruct & 17.39 & 15.41 & 17.22 & 19.72 & 16.55 & 16.79 & 18.57 & 15.61 & 26.35 & 14.74 & 15.45 & 14.85 \\
    & Qwen2.5-Coder-1.5B-Instruct & 14.93 & 9.96 & 15.42 & 17.74 & 16.09 & 15.15 & 13.98 & 16.11 & 18.31 & 16.66 & 13.00 & 11.76 \\
    & Qwen2.5-Coder-3B-Instruct & 10.76 & 6.94 & 11.35 & 11.52 & 11.79 & 9.25 & 9.39 & 12.83 & 13.83 & 12.48 & 10.18 & 8.78 \\
    
    \midrule
    \multirow{6}{*}{6B+} 
    & OpenCoder-8B-Instruct & 18.99 & 17.38 & 15.87 & 24.59 & 18.98 & 20.28 & 20.20 & 15.64 & 21.13 & 16.81 & 18.02 & 19.99 \\
\bottomrule
    \end{tabular}
    \end{adjustbox}
\end{table*}

% algo_level2_all_dim
\begin{table*}[h!]
    \centering
    \caption{Results of different models on advanced critique evaluations MSE in the Code Gen Subset Dataset across all fine-grained evaluation dimensions.} \label{table:algo_level2_all_dim}
    \begin{adjustbox}{width=\textwidth}
    \begin{tabular}{c|c|cccccccccc}
    
    \toprule
    \textbf{Size} & \textbf{Model} & \textbf{\makecell{Correctness\\Verification}} & \textbf{\makecell{Code\\Readability}} & \textbf{\makecell{Robustness\\Validation}} & \textbf{\makecell{Comprehensive\\Testing}} & \textbf{\makecell{Space\\Complexity}} & \textbf{\makecell{Code Style\\Consistency}} & \textbf{\makecell{Output\\Format}} & \textbf{\makecell{Maintain-\\-ability}} & \textbf{\makecell{Time\\Complexity}} & \textbf{\makecell{Algorithm\\Optimization}} \\
    
    \midrule 0.5B+ 
    & Qwen2.5-Coder-0.5B-Instruct & 35.17 & 30.70 & 26.72 & 24.09 & 50.49 & 35.20 & 61.39 & 36.78 & 42.97 & 42.84 \\

    \midrule
    \multirow{5}{*}{1B+} 
    & Yi-Coder-1.5B-Chat & 40.34 & 34.04 & 28.72 & 25.81 & 55.82 & 39.10 & 66.84 & 39.57 & 48.42 & 43.84 \\
    & DeepSeek-Coder-1.3B-Instruct & 37.59 & 30.75 & 26.10 & 25.41 & 50.82 & 35.65 & 61.07 & 37.07 & 43.62 & 40.34 \\
    & OpenCoder-1.5B-Instruct & 28.53 & 33.76 & 28.70 & 25.36 & 55.37 & 39.30 & 66.55 & 39.45 & 24.38 & 43.70 \\
    & Qwen2.5-Coder-1.5B-Instruct & 18.78 & 12.78 & 13.84 & 14.83 & 19.94 & 14.96 & 23.68 & 15.23 & 18.45 & 16.10 \\
    & Qwen2.5-Coder-3B-Instruct & 16.67 & 12.16 & 13.68 & 14.20 & 17.78 & 15.62 & 20.70 & 13.91 & 16.76 & 15.78 \\ 
    
    \midrule
    \multirow{6}{*}{6B+} 
    & OpenCoder-8B-Instruct & 27.22 & 24.39 & 22.05 & 23.03 & 36.40 & 27.75 & 44.76 & 29.13 & 32.16 & 30.65 \\
\bottomrule
    \end{tabular}
    \end{adjustbox}
\end{table*}

% algo_level2_mbpp_dim
\begin{table*}[h!]
    \centering
    \caption{Results of different models on advanced critique evaluations MSE in the Code Gen's MBPP subset Dataset across all fine-grained evaluation dimensions.} \label{table:algo_level2_mbpp_dim}
    \begin{adjustbox}{width=\textwidth}
    \begin{tabular}{c|c|cccccccccc}
    
    \toprule
    \textbf{Size} & \textbf{Model} & \textbf{\makecell{Correctness\\Verification}} & \textbf{\makecell{Code\\Readability}} & \textbf{\makecell{Robustness\\Validation}} & \textbf{\makecell{Comprehensive\\Testing}} & \textbf{\makecell{Space\\Complexity}} & \textbf{\makecell{Code Style\\Consistency}} & \textbf{\makecell{Output\\Format}} & \textbf{\makecell{Maintain-\\-ability}} & \textbf{\makecell{Time\\Complexity}} & \textbf{\makecell{Algorithm\\Optimization}} \\
    
    \midrule 0.5B+ 
    & Qwen2.5-Coder-0.5B-Instruct & 26.43 & 23.94 & 17.57 & 15.84 & 48.93 & 35.47 & 47.13 & 31.68 & 41.82 & 37.50 \\
    
    \midrule
    \multirow{5}{*}{1B+} 
    & Yi-Coder-1.5B-Chat & 32.02 & 29.14 & 20.26 & 18.26 & 57.32 & 42.15 & 53.41 & 37.14 & 50.27 & 44.75 \\
    & DeepSeek-Coder-1.3B-Instruct & 27.81 & 24.03 & 17.17 & 18.85 & 48.77 & 35.64 & 46.03 & 32.95 & 41.76 & 38.65 \\
    & OpenCoder-1.5B-Instruct & 22.53 & 29.68 & 21.54 & 17.15 & 58.31 & 43.84 & 55.42 & 38.25 & 24.05 & 45.71 \\
    & Qwen2.5-Coder-1.5B-Instruct & 17.75 & 10.64 & 11.09 & 14.23 & 19.44 & 14.05 & 16.93 & 12.07 & 18.76 & 14.27 \\
    & Qwen2.5-Coder-3B-Instruct & 13.90 & 9.70 & 9.81 & 12.17 & 16.41 & 13.00 & 14.65 & 11.06 & 15.31 & 13.29 \\
    
    \midrule
    \multirow{6}{*}{6B+} 
    & CodeQwen1.5-7B-Chat & 20.23 & 18.18 & 15.72 & 19.31 & 28.67 & 22.33 & 28.19 & 22.62 & 26.22 & 25.67 \\
\bottomrule
    \end{tabular}
    \end{adjustbox}
\end{table*}

% algo_level2_codeforce_dim
\begin{table*}[h!]
    \centering
    \caption{Results of different models on advanced critique evaluations MSE in the Code Gen's CodeForce subset Dataset across all fine-grained evaluation dimensions.} \label{table:algo_level2_codeforce_dim}
    \begin{adjustbox}{width=\textwidth}
    \begin{tabular}{c|c|cccccccccc}
    
    \toprule
    \textbf{Size} & \textbf{Model} & \textbf{\makecell{Correctness\\Verification}} & \textbf{\makecell{Code\\Readability}} & \textbf{\makecell{Robustness\\Validation}} & \textbf{\makecell{Comprehensive\\Testing}} & \textbf{\makecell{Space\\Complexity}} & \textbf{\makecell{Code Style\\Consistency}} & \textbf{\makecell{Output\\Format}} & \textbf{\makecell{Maintain-\\-ability}} & \textbf{\makecell{Time\\Complexity}} & \textbf{\makecell{Algorithm\\Optimization}} \\
    
    \midrule
    0.5B+ 
    & Qwen2.5-Coder-0.5B-Instruct & 34.15 & 27.62 & 26.64 & 23.15 & 51.75 & 35.96 & 58.01 & 31.43 & 46.22 & 50.37 \\
    
    \midrule
    \multirow{5}{*}{1B+} 
    & Yi-Coder-1.5B-Chat & 38.90 & 30.59 & 27.77 & 24.72 & 57.34 & 39.40 & 64.40 & 33.70 & 51.93 & 42.65 \\
    & DeepSeek-Coder-1.3B-Instruct & 37.69 & 29.16 & 26.22 & 24.49 & 53.43 & 37.18 & 60.31 & 32.40 & 48.45 & 40.60 \\
    & OpenCoder-1.5B-Instruct & 25.00 & 30.12 & 26.82 & 24.28 & 56.40 & 38.77 & 63.49 & 33.38 & 20.84 & 42.03 \\
    & Qwen2.5-Coder-3B-Instruct & 18.29 & 14.20 & 16.34 & 15.18 & 19.11 & 16.31 & 23.05 & 15.23 & 18.38 & 16.74 \\
    & Qwen2.5-Coder-1.5B-Instruct & 17.96 & 11.96 & 14.28 & 14.77 & 19.65 & 14.69 & 22.52 & 14.13 & 17.39 & 14.92 \\
    
    \midrule
    \multirow{6}{*}{6B+} 
    & OpenCoder-8B-Instruct & 29.19 & 24.30 & 24.07 & 23.59 & 40.27 & 30.71 & 46.84 & 28.38 & 36.51 & 33.33 \\
\bottomrule
    \end{tabular}
    \end{adjustbox}
\end{table*}

% algo_level2_livecodebench_dim
\begin{table*}[h!]
    \centering
    \caption{Results of different models on advanced critique evaluations MSE in the Code Gen's LiveCodeBench subset Dataset across all fine-grained evaluation dimensions.} \label{table:algo_level2_livecodebench_dim}
    \begin{adjustbox}{width=\textwidth}
    \begin{tabular}{c|c|cccccccccc}
    
    \toprule
    \textbf{Size} & \textbf{Model} & \textbf{\makecell{Correctness\\Verification}} & \textbf{\makecell{Code\\Readability}} & \textbf{\makecell{Robustness\\Validation}} & \textbf{\makecell{Comprehensive\\Testing}} & \textbf{\makecell{Space\\Complexity}} & \textbf{\makecell{Code Style\\Consistency}} & \textbf{\makecell{Output\\Format}} & \textbf{\makecell{Maintain-\\-ability}} & \textbf{\makecell{Time\\Complexity}} & \textbf{\makecell{Algorithm\\Optimization}} \\
    
    \midrule
    0.5B+ 
    & Qwen2.5-Coder-0.5B-Instruct & 45.79 & 49.24 & 40.35 & 29.94 & 47.71 & - & 75.26 & 48.04 & 35.84 & 36.23 \\
    
    \midrule
    \multirow{5}{*}{1B+} 
    & Yi-Coder-1.5B-Chat & 51.04 & 52.77 & 42.89 & 31.57 & 50.98 & - & 79.92 & 50.79 & 38.91 & 38.13 \\
    & DeepSeek-Coder-1.3B-Instruct & 48.18 & 48.61 & 39.79 & 30.54 & 47.19 & - & 72.97 & 47.45 & 36.29 & 35.78 \\
    & OpenCoder-1.5B-Instruct & 35.07 & 51.96 & 41.85 & 31.65 & 51.09 & - & 79.21 & 50.33 & 16.39 & 38.30 \\
    & Qwen2.5-Coder-1.5B-Instruct & 21.61 & 16.57 & 18.42 & 16.07 & 17.96 & - & 27.35 & 17.89 & 15.98 & 15.79 \\
    & Qwen2.5-Coder-3B-Instruct & 16.58 & 11.20 & 15.68 & 14.37 & 11.94 & - & 15.83 & 11.92 & 11.02 & 11.96 \\
    
    \midrule
    \multirow{6}{*}{6B+} 
    & CodeQwen1.5-7B-Chat & 34.25 & 32.96 & 28.22 & 23.22 & 32.62 & - & 50.29 & 34.24 & 26.06 & 26.25 \\
\bottomrule
    \end{tabular}
    \end{adjustbox}
\end{table*}

% algo_level2_debug_dim
\begin{table*}[h!]
    \centering
    \caption{Results of different models on advanced critique evaluations MSE in the Code Gen's Debug subset Dataset across all fine-grained evaluation dimensions.} \label{table:algo_level2_debug_dim}
    \begin{adjustbox}{width=\textwidth}
    \begin{tabular}{c|c|cccccccccc}
    
    \toprule
    \textbf{Size} & \textbf{Model} & \textbf{\makecell{Correctness\\Verification}} & \textbf{\makecell{Code\\Readability}} & \textbf{\makecell{Robustness\\Validation}} & \textbf{\makecell{Comprehensive\\Testing}} & \textbf{\makecell{Space\\Complexity}} & \textbf{\makecell{Code Style\\Consistency}} & \textbf{\makecell{Output\\Format}} & \textbf{\makecell{Maintain-\\-ability}} & \textbf{\makecell{Time\\Complexity}} & \textbf{\makecell{Algorithm\\Optimization}} \\
    
    \midrule 0.5B+ 
    & Qwen2.5-Coder-0.5B-Instruct & 35.30 & 25.15 & 26.83 & 27.25 & 53.51 & 34.16 & 63.47 & 34.21 & 48.51 & 47.20 \\

    \midrule
    \multirow{5}{*}{1B+} 
    & Yi-Coder-1.5B-Chat & 40.35 & 26.66 & 28.28 & 28.52 & 57.99 & 35.75 & 68.02 & 35.32 & 53.35 & 50.52 \\
    & DeepSeek-Coder-1.3B-Instruct & 37.69 & 24.41 & 25.70 & 27.59 & 53.81 & 34.12 & 63.02 & 33.84 & 48.61 & 46.54 \\
    & OpenCoder-1.5B-Instruct & 32.33 & 26.16 & 28.32 & 28.15 & 56.14 & 35.30 & 66.79 & 34.73 & 36.22 & 49.57 \\
    & Qwen2.5-Coder-3B-Instruct & 18.10 & 13.77 & 14.51 & 15.02 & 23.75 & 17.55 & 28.24 & 17.32 & 22.61 & 21.13 \\
    & Qwen2.5-Coder-1.5B-Instruct & 17.96 & 12.59 & 13.10 & 14.16 & 22.72 & 16.14 & 27.05 & 16.20 & 21.68 & 19.17 \\
    
    \midrule
    \multirow{6}{*}{6B+} 
    & OpenCoder-8B-Instruct & 26.91 & 22.03 & 22.67 & 23.80 & 40.94 & 28.41 & 49.26 & 28.76 & 37.80 & 36.40 \\
\bottomrule
    \end{tabular}
    \end{adjustbox}
\end{table*}

% real_level2_all_dim
\begin{table*}[h!]
    \centering
    \caption{Results of different models on advanced critique evaluations MSE in the Code QA Subset Dataset across all fine-grained evaluation dimensions.} \label{table:real_level2_all_dim}
    \begin{adjustbox}{width=\textwidth}
    \begin{tabular}{c|c|cccccccccc}
    
    \toprule
    \textbf{Size} & \textbf{Model} & \textbf{\makecell{Depth}} & \textbf{\makecell{Logical\\Coherence}} & \textbf{\makecell{Innovation}} & \textbf{\makecell{Practicality}} & \textbf{\makecell{Clarity}} & \textbf{\makecell{Reliability}} & \textbf{\makecell{Completeness}} & \textbf{\makecell{Maintainability}} & \textbf{\makecell{Correctness}} & \textbf{\makecell{Performance}} \\

    \midrule
    0.5B+ 
    & Qwen2.5-Coder-0.5B-Instruct & 40.75 & 53.88 & 33.81 & 39.04 & 52.26 & 39.14 & 40.37 & 41.45 & 39.28 & 40.21 \\
    
    \midrule
    \multirow{5}{*}{1B+} 
    & Yi-Coder-1.5B-Chat & 43.30 & 57.61 & 35.68 & 41.71 & 55.84 & 41.77 & 43.19 & 44.28 & 41.76 & 42.69 \\
    & OpenCoder-1.5B-Instruct & 42.51 & 56.39 & 35.24 & 40.89 & 54.77 & 41.32 & 40.97 & 43.45 & 39.77 & 41.90 \\
    & DeepSeek-Coder-1.3B-Instruct & 29.37 & 36.60 & 24.85 & 26.27 & 34.98 & 27.78 & 28.25 & 27.93 & 27.48 & 27.74 \\
    & Qwen2.5-Coder-1.5B-Instruct & 21.08 & 25.80 & 19.24 & 18.97 & 23.34 & 20.82 & 19.55 & 19.46 & 19.90 & 19.70 \\
    & Qwen2.5-Coder-3B-Instruct & 17.38 & 20.88 & 15.68 & 15.49 & 19.03 & 17.88 & 16.76 & 16.35 & 14.87 & 15.69 \\
    
    \midrule
    \multirow{6}{*}{6B+} 
    & OpenCoder-8B-Instruct & 39.72 & 51.43 & 33.14 & 37.68 & 49.73 & 37.79 & 38.42 & 39.84 & 34.09 & 39.17 \\
\bottomrule
    \end{tabular}
    \end{adjustbox}
\end{table*}

% real_level2_fp_dim
\begin{table*}[h!]
    \centering
    \caption{Results of different models on advanced critique evaluations MSE in the Code QA's Fundamental Programming (FP) subset Dataset across all fine-grained evaluation dimensions.} \label{table:real_level2_fp_dim}
    \begin{adjustbox}{width=\textwidth}
    \begin{tabular}{c|c|cccccccccc}
    
    \toprule
    \textbf{Size} & \textbf{Model} & \textbf{\makecell{Depth}} & \textbf{\makecell{Logical\\Coherence}} & \textbf{\makecell{Innovation}} & \textbf{\makecell{Practicality}} & \textbf{\makecell{Clarity}} & \textbf{\makecell{Reliability}} & \textbf{\makecell{Completeness}} & \textbf{\makecell{Maintainability}} & \textbf{\makecell{Correctness}} & \textbf{\makecell{Performance}} \\

    \midrule 0.5B+ 
    & Qwen2.5-Coder-0.5B-Instruct & 44.06 & 60.14 & 35.97 & 43.10 & 53.45 & 43.56 & 47.37 & 43.64 & 44.09 & 44.82 \\

    \midrule
    \multirow{5}{*}{1B+} 
    & Yi-Coder-1.5B-Chat & 46.06 & 62.58 & 36.59 & 44.81 & 55.22 & 45.37 & 49.76 & 45.23 & 46.58 & 45.80 \\
    & OpenCoder-1.5B-Instruct & 43.14 & 57.25 & 34.64 & 41.50 & 51.09 & 43.96 & 43.37 & 41.92 & 41.03 & 43.76 \\
    & DeepSeek-Coder-1.3B-Instruct & 28.22 & 35.67 & 23.48 & 25.73 & 30.89 & 29.53 & 30.92 & 27.49 & 30.15 & 26.68 \\
    & Qwen2.5-Coder-1.5B-Instruct & 12.76 & 12.22 & 12.60 & 9.95 & 11.02 & 12.80 & 12.07 & 9.54 & 12.54 & 12.35 \\ 
    & Qwen2.5-Coder-3B-Instruct & 9.94 & 9.56 & 9.32 & 7.93 & 9.80 & 11.33 & 9.93 & 8.41 & 8.50 & 10.61 \\
    
    \midrule
    \multirow{6}{*}{6B+} 
    & OpenCoder-8B-Instruct & 41.60 & 52.93 & 32.34 & 38.69 & 46.25 & 39.27 & 42.98 & 38.88 & 32.81 & 41.62 \\
\bottomrule
    \end{tabular}
    \end{adjustbox}
\end{table*}

% real_level2_ap_dim
\begin{table*}[h!]
    \centering
    \caption{Results of different models on advanced critique evaluations MSE in the Code QA's Advanced Programming (AP) subset Dataset across all fine-grained evaluation dimensions.} \label{table:real_level2_ap_dim}
    \begin{adjustbox}{width=\textwidth}
    \begin{tabular}{c|c|cccccccccc}
    
    \toprule
    \textbf{Size} & \textbf{Model} & \textbf{\makecell{Depth}} & \textbf{\makecell{Logical\\Coherence}} & \textbf{\makecell{Innovation}} & \textbf{\makecell{Practicality}} & \textbf{\makecell{Clarity}} & \textbf{\makecell{Reliability}} & \textbf{\makecell{Completeness}} & \textbf{\makecell{Maintainability}} & \textbf{\makecell{Correctness}} & \textbf{\makecell{Performance}} \\

    \midrule 0.5B+ 
    & Qwen2.5-Coder-0.5B-Instruct & 45.51 & 55.76 & 41.08 & 37.07 & 51.21 & 39.29 & 41.18 & 38.59 & 38.69 & 37.36 \\
    
    \midrule
    \multirow{5}{*}{1B+} 
    & OpenCoder-1.5B-Instruct & 47.26 & 57.97 & 42.39 & 38.99 & 52.06 & 41.32 & 40.33 & 39.88 & 40.10 & 38.19 \\
    & Yi-Coder-1.5B-Chat & 46.46 & 57.16 & 42.39 & 38.11 & 52.85 & 40.50 & 42.84 & 39.88 & 40.11 & 38.53 \\
    & DeepSeek-Coder-1.3B-Instruct & 29.62 & 37.65 & 26.04 & 22.34 & 31.06 & 28.40 & 24.99 & 23.25 & 25.14 & 22.16 \\
    & Qwen2.5-Coder-1.5B-Instruct & 25.54 & 28.91 & 21.78 & 17.47 & 22.52 & 23.57 & 20.11 & 16.92 & 21.39 & 16.79 \\
    & Qwen2.5-Coder-3B-Instruct & 18.83 & 22.93 & 16.52 & 12.92 & 17.87 & 18.54 & 16.49 & 12.38 & 14.15 & 12.93 \\ 
        
    \midrule
    \multirow{6}{*}{6B+} 
    & OpenCoder-8B-Instruct & 45.56 & 54.22 & 40.70 & 36.34 & 48.87 & 38.99 & 41.24 & 37.71 & 33.51 & 37.85 \\
\bottomrule
    \end{tabular}
    \end{adjustbox}
\end{table*}

% real_level2_se_dim
\begin{table*}[h!]
    \centering
    \caption{Results of different models on advanced critique evaluations MSE in the Code QA's Software Engineering (SE) subset Dataset across all fine-grained evaluation dimensions.} \label{table:real_level2_se_dim}
    \begin{adjustbox}{width=\textwidth}
    \begin{tabular}{c|c|cccccccccc}
    
    \toprule
    \textbf{Size} & \textbf{Model} & \textbf{\makecell{Depth}} & \textbf{\makecell{Logical\\Coherence}} & \textbf{\makecell{Innovation}} & \textbf{\makecell{Practicality}} & \textbf{\makecell{Clarity}} & \textbf{\makecell{Reliability}} & \textbf{\makecell{Completeness}} & \textbf{\makecell{Maintainability}} & \textbf{\makecell{Correctness}} & \textbf{\makecell{Performance}} \\
    
    \midrule 0.5B+ 
    & Qwen2.5-Coder-0.5B-Instruct & 42.40 & 55.81 & 36.02 & 39.60 & 57.80 & 41.68 & 42.49 & 41.45 & 39.90 & 39.56 \\
    
    \midrule
    \multirow{5}{*}{1B+} 
    & Yi-Coder-1.5B-Chat & 42.78 & 57.08 & 36.60 & 40.32 & 59.63 & 42.54 & 43.69 & 43.42 & 40.96 & 40.63 \\
    & OpenCoder-1.5B-Instruct & 42.00 & 55.89 & 36.60 & 39.51 & 59.29 & 41.58 & 41.00 & 42.23 & 38.40 & 39.62 \\
    & DeepSeek-Coder-1.3B-Instruct & 32.07 & 39.88 & 25.91 & 27.22 & 40.49 & 30.48 & 31.29 & 28.67 & 28.82 & 27.88 \\
    & Qwen2.5-Coder-1.5B-Instruct & 24.80 & 30.91 & 22.52 & 22.00 & 30.58 & 25.67 & 23.78 & 21.41 & 22.62 & 22.27 \\
    & Qwen2.5-Coder-3B-Instruct & 20.79 & 25.54 & 19.43 & 18.70 & 23.39 & 22.44 & 19.52 & 18.45 & 17.07 & 18.10 \\
    
    \midrule
    \multirow{6}{*}{6B+} 
    & OpenCoder-8B-Instruct & 41.53 & 53.09 & 35.66 & 38.07 & 53.75 & 40.88 & 41.22 & 39.73 & 36.33 & 38.95 \\
\bottomrule
    \end{tabular}
    \end{adjustbox}
\end{table*}

% real_level2_da_dim
\begin{table*}[h!]
    \centering
    \caption{Results of different models on advanced critique evaluations MSE in the Code QA's Data Analysis (DA) subset Dataset across all fine-grained evaluation dimensions.} \label{table:real_level2_da_dim}
    \begin{adjustbox}{width=\textwidth}
    \begin{tabular}{c|c|cccccccccc}
    
    \toprule
    \textbf{Size} & \textbf{Model} & \textbf{\makecell{Depth}} & \textbf{\makecell{Logical\\Coherence}} & \textbf{\makecell{Innovation}} & \textbf{\makecell{Practicality}} & \textbf{\makecell{Clarity}} & \textbf{\makecell{Reliability}} & \textbf{\makecell{Completeness}} & \textbf{\makecell{Maintainability}} & \textbf{\makecell{Correctness}} & \textbf{\makecell{Performance}} \\

    \midrule 0.5B+ 
    & Qwen2.5-Coder-0.5B-Instruct & 53.91 & 64.58 & 39.54 & 48.70 & 59.13 & 44.81 & 46.18 & 47.06 & 44.33 & 44.24 \\
    
    \midrule
    \multirow{5}{*}{1B+} 
    & OpenCoder-1.5B-Instruct & 56.32 & 67.10 & 41.17 & 50.86 & 62.70 & 46.76 & 45.45 & 48.57 & 44.34 & 45.50 \\
    & Yi-Coder-1.5B-Chat & 55.52 & 65.71 & 40.44 & 50.03 & 60.74 & 45.97 & 46.89 & 48.08 & 45.21 & 45.27 \\
    & DeepSeek-Coder-1.3B-Instruct & 29.48 & 35.79 & 24.35 & 27.24 & 33.61 & 26.33 & 29.00 & 27.71 & 27.03 & 27.78 \\
    & Qwen2.5-Coder-1.5B-Instruct & 22.73 & 27.50 & 19.45 & 20.50 & 23.24 & 23.12 & 20.73 & 20.23 & 21.34 & 20.80 \\
    & Qwen2.5-Coder-3B-Instruct & 17.51 & 20.07 & 16.84 & 16.93 & 16.79 & 18.84 & 15.71 & 15.48 & 16.84 & 16.01 \\
    
    \midrule
    \multirow{6}{*}{6B+} 
    & OpenCoder-8B-Instruct & 50.71 & 60.43 & 38.12 & 46.03 & 56.41 & 42.14 & 43.60 & 44.37 & 36.61 & 42.88 \\
\bottomrule
    \end{tabular}
    \end{adjustbox}
\end{table*}

% real_level2_ma_dim
\begin{table*}[h!]
    \centering
    \caption{Results of different models on advanced critique evaluations MSE in the Code QA's Mathematics (MA) subset Dataset across all fine-grained evaluation dimensions.} \label{table:real_level2_ma_dim}
    \begin{adjustbox}{width=\textwidth}
    \begin{tabular}{c|c|cccccccccc}
    
    \toprule
    \textbf{Size} & \textbf{Model} & \textbf{\makecell{Depth}} & \textbf{\makecell{Logical\\Coherence}} & \textbf{\makecell{Innovation}} & \textbf{\makecell{Practicality}} & \textbf{\makecell{Clarity}} & \textbf{\makecell{Reliability}} & \textbf{\makecell{Completeness}} & \textbf{\makecell{Maintainability}} & \textbf{\makecell{Correctness}} & \textbf{\makecell{Performance}} \\

    \midrule
    0.5B+ 
    & Qwen2.5-Coder-0.5B-Instruct & 39.44 & 50.08 & 31.14 & 38.55 & 52.32 & 45.26 & 38.25 & 42.16 & 43.15 & 46.26 \\
    
    \midrule
    \multirow{5}{*}{1B+} 
    & Yi-Coder-1.5B-Chat & 41.41 & 52.87 & 33.03 & 41.97 & 55.99 & 47.62 & 40.33 & 45.31 & 46.09 & 50.73 \\
    & OpenCoder-1.5B-Instruct & 39.39 & 50.19 & 32.16 & 39.58 & 53.12 & 45.45 & 38.79 & 43.04 & 42.16 & 48.33 \\
    & DeepSeek-Coder-1.3B-Instruct & 28.37 & 33.01 & 24.05 & 27.39 & 34.94 & 30.55 & 26.75 & 27.89 & 31.29 & 33.01 \\
    & Qwen2.5-Coder-1.5B-Instruct & 18.35 & 22.74 & 18.58 & 19.60 & 20.08 & 21.63 & 17.84 & 18.89 & 20.95 & 21.10 \\
    & Qwen2.5-Coder-3B-Instruct & 15.08 & 18.64 & 14.03 & 15.36 & 17.96 & 17.31 & 15.19 & 17.55 & 13.29 & 16.26 \\
    
    \midrule
    \multirow{6}{*}{6B+} 
    & OpenCoder-8B-Instruct & 34.28 & 42.44 & 28.84 & 35.28 & 46.06 & 39.15 & 32.56 & 37.18 & 35.57 & 42.30 \\
\bottomrule
    \end{tabular}
    \end{adjustbox}
\end{table*}

% real_level2_dw_dim
\begin{table*}[h!]
    \centering
    \caption{Results of different models on advanced critique evaluations MSE in the Code QA's Desktop and Web Development (DW) subset Dataset across all fine-grained evaluation dimensions.} \label{table:real_level2_dw_dim}
    \begin{adjustbox}{width=\textwidth}
    \begin{tabular}{c|c|cccccccccc}
    
    \toprule
    \textbf{Size} & \textbf{Model} & \textbf{\makecell{Depth}} & \textbf{\makecell{Logical\\Coherence}} & \textbf{\makecell{Innovation}} & \textbf{\makecell{Practicality}} & \textbf{\makecell{Clarity}} & \textbf{\makecell{Reliability}} & \textbf{\makecell{Completeness}} & \textbf{\makecell{Maintainability}} & \textbf{\makecell{Correctness}} & \textbf{\makecell{Performance}} \\

    \midrule 0.5B+ 
    & Qwen2.5-Coder-0.5B-Instruct & 42.60 & 56.97 & 36.01 & 40.85 & 56.98 & 39.75 & 38.04 & 40.97 & 38.68 & 45.10 \\
    
    \midrule
    \multirow{5}{*}{1B+} 
    & Yi-Coder-1.5B-Chat & 47.11 & 62.39 & 38.85 & 44.77 & 61.72 & 43.45 & 41.81 & 44.59 & 41.62 & 48.50 \\
    & OpenCoder-1.5B-Instruct & 46.00 & 60.24 & 38.01 & 43.41 & 59.68 & 43.35 & 39.63 & 43.46 & 39.89 & 47.42 \\
    & DeepSeek-Coder-1.3B-Instruct & 31.50 & 37.38 & 27.30 & 27.93 & 37.93 & 26.98 & 25.75 & 27.05 & 25.01 & 31.89 \\
    & Qwen2.5-Coder-1.5B-Instruct & 20.89 & 25.86 & 18.81 & 19.54 & 24.28 & 21.08 & 18.83 & 19.33 & 18.62 & 20.70 \\
    & Qwen2.5-Coder-3B-Instruct & 15.06 & 18.75 & 14.58 & 14.04 & 16.75 & 15.90 & 15.61 & 15.51 & 14.50 & 15.58 \\ 
        
    \midrule
    \multirow{6}{*}{6B+} 
    & OpenCoder-8B-Instruct & 44.90 & 58.01 & 37.53 & 42.20 & 57.47 & 40.85 & 38.37 & 41.82 & 36.62 & 45.83 \\
\bottomrule
    \end{tabular}
    \end{adjustbox}
\end{table*}

% real_level2_ml_dim
\begin{table*}[h!]
    \centering
    \caption{Results of different models on advanced critique evaluations MSE in the Code QA's Machine Learning (ML) subset Dataset across all fine-grained evaluation dimensions.} \label{table:real_level2_ml_dim}
    \begin{adjustbox}{width=\textwidth}
    \begin{tabular}{c|c|cccccccccc}
    
    \toprule
    \textbf{Size} & \textbf{Model} & \textbf{\makecell{Depth}} & \textbf{\makecell{Logical\\Coherence}} & \textbf{\makecell{Innovation}} & \textbf{\makecell{Practicality}} & \textbf{\makecell{Clarity}} & \textbf{\makecell{Reliability}} & \textbf{\makecell{Completeness}} & \textbf{\makecell{Maintainability}} & \textbf{\makecell{Correctness}} & \textbf{\makecell{Performance}} \\

    \midrule 0.5B+ 
    & Qwen2.5-Coder-0.5B-Instruct & 29.82 & 43.62 & 26.61 & 33.38 & 49.07 & 25.95 & 33.75 & 39.83 & 29.71 & 29.50 \\
    
    \midrule
    \multirow{5}{*}{1B+} 
    & Yi-Coder-1.5B-Chat & 31.58 & 46.40 & 28.07 & 35.97 & 51.14 & 28.35 & 36.87 & 43.52 & 31.58 & 31.65 \\
    & OpenCoder-1.5B-Instruct & 30.77 & 45.54 & 27.71 & 35.61 & 50.15 & 27.65 & 35.35 & 42.62 & 30.90 & 31.34 \\
    & DeepSeek-Coder-1.3B-Instruct & 25.91 & 35.68 & 24.42 & 29.28 & 42.38 & 23.51 & 28.56 & 31.58 & 25.36 & 24.71 \\
    & Qwen2.5-Coder-1.5B-Instruct & 21.61 & 29.59 & 22.04 & 23.89 & 30.93 & 19.78 & 22.75 & 26.89 & 18.88 & 19.86 \\
    & Qwen2.5-Coder-3B-Instruct & 18.82 & 25.57 & 18.97 & 20.60 & 28.27 & 17.56 & 21.74 & 23.98 & 16.83 & 18.25 \\ 
        
    \midrule
    \multirow{6}{*}{6B+} 
    & OpenCoder-8B-Instruct & 29.08 & 41.59 & 26.87 & 33.34 & 45.99 & 26.01 & 33.70 & 39.13 & 28.16 & 29.15 \\
\bottomrule
    \end{tabular}
    \end{adjustbox}
\end{table*}

% real_level2_sc_dim
\begin{table*}[h!]
    \centering
    \caption{Results of different models on advanced critique evaluations MSE in the Code QA's Scientific Computing (SC) subset Dataset across all fine-grained evaluation dimensions.} \label{table:real_level2_sc_dim}
    \begin{adjustbox}{width=\textwidth}
    \begin{tabular}{c|c|cccccccccc}
    
    \toprule
    \textbf{Size} & \textbf{Model} & \textbf{\makecell{Depth}} & \textbf{\makecell{Logical\\Coherence}} & \textbf{\makecell{Innovation}} & \textbf{\makecell{Practicality}} & \textbf{\makecell{Clarity}} & \textbf{\makecell{Reliability}} & \textbf{\makecell{Completeness}} & \textbf{\makecell{Maintainability}} & \textbf{\makecell{Correctness}} & \textbf{\makecell{Performance}} \\

    \midrule 0.5B+ 
    & Qwen2.5-Coder-0.5B-Instruct & 40.92 & 50.51 & 30.96 & 37.29 & 49.62 & 36.30 & 37.76 & 40.67 & 38.71 & 37.56 \\
    
    \midrule
    \multirow{5}{*}{1B+} 
    & OpenCoder-1.5B-Instruct & 46.36 & 57.60 & 33.92 & 42.06 & 55.02 & 41.98 & 41.68 & 46.33 & 41.49 & 42.02 \\
    & Yi-Coder-1.5B-Chat & 46.15 & 57.60 & 33.31 & 42.06 & 55.67 & 41.38 & 41.88 & 45.75 & 42.17 & 41.66 \\
    & DeepSeek-Coder-1.3B-Instruct & 39.93 & 45.04 & 29.06 & 32.73 & 40.89 & 35.29 & 35.46 & 34.46 & 36.02 & 32.66 \\
    & Qwen2.5-Coder-3B-Instruct & 30.53 & 35.70 & 21.45 & 23.44 & 31.06 & 27.60 & 25.32 & 24.74 & 20.67 & 21.75 \\ 
    & Qwen2.5-Coder-1.5B-Instruct & 29.26 & 33.70 & 22.28 & 24.02 & 29.82 & 25.49 & 22.84 & 23.99 & 24.63 & 22.91 \\
        
    \midrule
    \multirow{6}{*}{6B+} 
    & OpenCoder-8B-Instruct & 42.00 & 51.16 & 30.91 & 37.28 & 49.80 & 37.75 & 35.17 & 40.96 & 36.43 & 37.89 \\
\bottomrule
    \end{tabular}
    \end{adjustbox}
\end{table*}

% real_level2_db_dim
\begin{table*}[h!]
    \centering
    \caption{Results of different models on advanced critique evaluations MSE in the Code QA's Databases (DB) subset Dataset across all fine-grained evaluation dimensions.} \label{table:real_level2_db_dim}
    \begin{adjustbox}{width=\textwidth}
    \begin{tabular}{c|c|cccccccccc}
    
    \toprule
    \textbf{Size} & \textbf{Model} & \textbf{\makecell{Depth}} & \textbf{\makecell{Logical\\Coherence}} & \textbf{\makecell{Innovation}} & \textbf{\makecell{Practicality}} & \textbf{\makecell{Clarity}} & \textbf{\makecell{Reliability}} & \textbf{\makecell{Completeness}} & \textbf{\makecell{Maintainability}} & \textbf{\makecell{Correctness}} & \textbf{\makecell{Performance}} \\

    \midrule 0.5B+ 
    & Qwen2.5-Coder-0.5B-Instruct & 38.65 & 50.80 & 33.44 & 38.38 & 45.36 & 38.22 & 38.33 & 39.77 & 40.57 & 40.81 \\
    
    \midrule
    \multirow{5}{*}{1B+} 
    & Yi-Coder-1.5B-Chat & 40.85 & 56.44 & 36.37 & 42.39 & 50.84 & 41.84 & 42.52 & 43.71 & 46.36 & 44.75 \\
    & OpenCoder-1.5B-Instruct & 40.15 & 55.56 & 35.60 & 41.68 & 50.43 & 41.08 & 38.78 & 43.57 & 44.09 & 43.83 \\
    & DeepSeek-Coder-1.3B-Instruct & 27.13 & 34.74 & 25.18 & 24.42 & 31.10 & 26.93 & 27.42 & 27.53 & 29.42 & 28.47 \\
    & Qwen2.5-Coder-1.5B-Instruct & 21.59 & 27.05 & 19.98 & 20.00 & 22.60 & 22.17 & 20.72 & 21.00 & 23.43 & 22.81 \\
    & Qwen2.5-Coder-3B-Instruct & 16.91 & 20.18 & 15.56 & 16.61 & 19.44 & 17.75 & 17.58 & 18.19 & 17.79 & 18.34 \\ 
    
    \midrule
    \multirow{6}{*}{6B+} 
    & OpenCoder-8B-Instruct & 36.05 & 50.97 & 33.33 & 37.15 & 45.27 & 37.50 & 37.29 & 39.31 & 38.22 & 39.70 \\
\bottomrule
    \end{tabular}
    \end{adjustbox}
\end{table*}

% real_level2_mm_dim
\begin{table*}[h!]
    \centering
    \caption{Results of different models on advanced critique evaluations MSE in the Code QA's Multimedia (MM) subset Dataset across all fine-grained evaluation dimensions.} \label{table:real_level2_mm_dim}
    \begin{adjustbox}{width=\textwidth}
    \begin{tabular}{c|c|cccccccccc}
    
    \toprule
    \textbf{Size} & \textbf{Model} & \textbf{\makecell{Depth}} & \textbf{\makecell{Logical\\Coherence}} & \textbf{\makecell{Innovation}} & \textbf{\makecell{Practicality}} & \textbf{\makecell{Clarity}} & \textbf{\makecell{Reliability}} & \textbf{\makecell{Completeness}} & \textbf{\makecell{Maintainability}} & \textbf{\makecell{Correctness}} & \textbf{\makecell{Performance}} \\

    \midrule 0.5B+ 
    & Qwen2.5-Coder-0.5B-Instruct & 37.28 & 55.85 & 30.54 & 36.18 & 53.12 & 37.42 & 39.91 & 42.45 & 36.76 & 38.01 \\
    
    \midrule
    \multirow{5}{*}{1B+} 
    & Yi-Coder-1.5B-Chat & 39.84 & 58.96 & 32.30 & 37.82 & 55.74 & 39.69 & 43.07 & 44.62 & 37.86 & 40.32 \\
    & OpenCoder-1.5B-Instruct & 39.39 & 58.91 & 31.84 & 37.82 & 56.57 & 39.29 & 42.63 & 44.98 & 35.65 & 40.09 \\
    & DeepSeek-Coder-1.3B-Instruct & 26.69 & 35.51 & 21.31 & 22.86 & 32.08 & 23.82 & 25.93 & 27.99 & 24.07 & 26.92 \\
    & Qwen2.5-Coder-1.5B-Instruct & 18.84 & 22.73 & 16.66 & 15.18 & 22.08 & 16.13 & 17.56 & 19.59 & 18.06 & 19.32 \\
    & Qwen2.5-Coder-3B-Instruct & 15.93 & 17.98 & 13.72 & 12.74 & 15.66 & 15.74 & 14.80 & 13.81 & 13.71 & 12.72 \\
    
    \midrule
    \multirow{6}{*}{6B+} 
    & OpenCoder-8B-Instruct & 35.80 & 51.98 & 28.56 & 33.64 & 50.21 & 34.92 & 35.63 & 40.08 & 28.35 & 35.70 \\
\bottomrule
    \end{tabular}
    \end{adjustbox}
\end{table*}

% real_level2_os_dim
\begin{table*}[h!]
    \centering
    \caption{Results of different models on advanced critique evaluations MSE in the Code QA's Operating Systems (OS) subset Dataset across all fine-grained evaluation dimensions.} \label{table:real_level2_os_dim}
    \begin{adjustbox}{width=\textwidth}
    \begin{tabular}{c|c|cccccccccc}
    
    \toprule
    \textbf{Size} & \textbf{Model} & \textbf{\makecell{Depth}} & \textbf{\makecell{Logical\\Coherence}} & \textbf{\makecell{Innovation}} & \textbf{\makecell{Practicality}} & \textbf{\makecell{Clarity}} & \textbf{\makecell{Reliability}} & \textbf{\makecell{Completeness}} & \textbf{\makecell{Maintainability}} & \textbf{\makecell{Correctness}} & \textbf{\makecell{Performance}} \\

    \midrule 0.5B+ 
    & Qwen2.5-Coder-0.5B-Instruct & 37.23 & 52.04 & 32.67 & 36.58 & 48.52 & 38.73 & 39.77 & 39.46 & 38.16 & 39.88 \\
    
    \midrule
    \multirow{5}{*}{1B+} 
    & Yi-Coder-1.5B-Chat & 41.51 & 59.50 & 36.44 & 41.02 & 55.47 & 43.11 & 44.83 & 43.20 & 42.22 & 43.62 \\
    & OpenCoder-1.5B-Instruct & 40.25 & 57.55 & 35.67 & 39.37 & 53.62 & 42.63 & 43.07 & 41.66 & 40.35 & 42.08 \\
    & DeepSeek-Coder-1.3B-Instruct & 23.79 & 33.21 & 21.92 & 22.70 & 30.87 & 24.01 & 25.23 & 24.17 & 21.66 & 24.68 \\
    & Qwen2.5-Coder-1.5B-Instruct & 15.99 & 22.99 & 16.69 & 17.20 & 21.40 & 18.32 & 18.09 & 17.31 & 16.62 & 18.07 \\
    & Qwen2.5-Coder-3B-Instruct & 11.24 & 14.93 & 11.58 & 12.06 & 13.43 & 14.15 & 13.40 & 12.40 & 9.88 & 11.92 \\
    
    \midrule
    \multirow{6}{*}{6B+} 
    & OpenCoder-8B-Instruct & 38.89 & 52.56 & 33.80 & 36.67 & 48.56 & 38.94 & 39.70 & 39.08 & 33.09 & 39.31 \\
\bottomrule
    \end{tabular}
    \end{adjustbox}
\end{table*}

\clearpage

\begin{figure*}[h!]
\begin{center}
    \fontsize{8.4}{8.4} \selectfont
    \begin{tcolorbox}[width=1\textwidth, colback=lightblue, title={\textbf{Basic Critique Evaluation Judge Prompt}}]

    You are a senior programming expert. We request a professional and precise evaluation of the AI assistant's performance based on the user’s question.\\
    
    \textbf{Evaluation Process}:\\
    1. Carefully review the user’s question and the assistant’s answer.\\
    2. Verify whether the answer (may include code) fully meet the user’s requirements.\\
    3. Pay close attention to specific aspects of the answer, including:\\
    .\quad - Appropriateness of the answer’s perspective.\\ 
    .\quad - Accuracy of technical details.\\
    .\quad - Completeness of the solution.\\
    
    \textbf{Evaluation Requirements}:\\
    - Provide detailed, point-by-point feedback on the answer.\\
    - Each critique should be specific and self-contained.\\
    - Clearly identify any issues, avoiding vague or ambiguous descriptions.\\
    - Offer constructive suggestions for improvement.\\
    
    \textbf{Evaluation Criteria}:\\
    - "Error": The code contains incorrect conclusions, explanations, or fails to meet the specified requirements.\\
    - "Correct": The code is entirely accurate and meets all outlined requirements.\\
    
    \textbf{Output Format}:\\
    Provide the evaluation in JSON format as follows:\\
    \verb|```|json\\
    \{\\
        .\quad"reasons": "Detailed explanation evaluating the response, addressing specific points",\\
        .\quad"is\_assistant\_correct": "Based on the evaluation, indicate whether the assistant's response is ['Correct', 'Error']"\\
    \}\\
    \verb|```|\\
    
    \textbf{Input Data}:\\
    --- Start of Question ---\\
    \textcolor{ora}{\$Question}\\
    --- End of Question ---\\
    
    --- Start of Answer ---\\
    \textcolor{ora}{\$Answer}\\
    --- End of Answer ---\\
    \end{tcolorbox}
\end{center}
\caption{Basic critique evaluation judge prompt.} \label{ap:prompt_level1}
\end{figure*}

\begin{figure*}[h!]
\begin{center}
    \fontsize{8.4}{8.4} \selectfont
    \begin{tcolorbox}[width=0.7\textwidth, colback=lightblue, title={\textbf{Advanced Critique Evaluation Judge Prompt}}]

    You are a professional code evaluation expert with the following qualifications:\\
    - Expertise in programming languages and algorithms.\\
    - Skilled in accurately assessing answer (may include code) quality.\\
    - Proficient in defining and applying precise, professional scoring criteria.\\ 
    
    \textbf{Evaluation Process}:\\
    1. Analyze and understand the problem statement and provided answer thoroughly.\\
    2. Evaluate systematically against defined criteria.\\
    3. Assign professional scores for each evaluation problem.\\
    4. Aggregate individual scores to determine a final, comprehensive score.\\
    
    \textbf{Output Specifications}:\\
    - Scoring Range:\\
       .\quad- Each evaluation metric is scored on an integer scale from 1-10.\\
       .\quad- The comprehensive score is also on a 1-10 scale.\\
    - Justification: All scores must be supported by clear, specific reasoning.\\
    - Structure: The final output must be well-organized, concise and professional.\\
    - Objectivity: Scoring must be neutral and unbiased.\\
    
    \textbf{Scoring Guidelines}:\\
    - 1-2 points: Critical flaws; fails to meet requirements.\\
    - 3-4 points: Significant deficiencies; largely unusable.\\
    - 5-6 points: Functional but requires substantial improvement.\\
    - 7-8 points: Well-implemented with minor issues.\\
    - 9-10 points: High-quality, near-perfect implementation.\\
    
    \textbf{Output Format}:\\
    \verb|```|md\\
    1. <Reason>, Score: xx\\
    2. <Reason>, Score: xx\\
    ...\\
    Comprehensive evaluation: <Reason>, Comprehensive Score: xx\\
    \verb|```|\\
    
    \textbf{Important Notes}:\\
    - The final comprehensive score should reflect a weighted assessment of all sub-scores.\\
    
    \textbf{Input Data}:\\
    --- Start of Question ---\\
    \textcolor{ora}{\$Question}\\
    --- End of Question ---\\
    
    --- Start of Answer ---\\
    \textcolor{ora}{\$Answer}\\
    --- End of Answer ---\\
    
    --- Start of Fine-Grained Evaluation Checklists ---\\ 
    \textcolor{ora}{\$Checklists}\\
    --- End of Fine-Grained Evaluation Checklists ---\\

    \end{tcolorbox}
\end{center}
\caption{Advanced critique evaluation judge prompt.} \label{ap:prompt_level2}
\end{figure*}

\begin{figure*}[h!]
\begin{center}
    \fontsize{8.4}{8.4} \selectfont
    \begin{tcolorbox}[width=1\textwidth, colback=lightblue, title={\textbf{Full Error Typelists}}]

    1. \textbf{Syntax Error}: [``Spelling mistakes (e.g., incorrect variable names)'', ``Missing semicolons or mismatched brackets'', ``Incorrect use of keywords'', ``Indentation errors (for indentation-sensitive languages)'', ``Incorrect comment formatting''] \\
    
    2. \textbf{Reference Error}: [``Undefined variables'', ``Undefined functions'', ``Null pointer references'', ``Array out of bounds'', ``Dangling pointers''] \\
    
    3. \textbf{Type Error}: [``Type conversion errors'', ``Undefined type errors'', ``Type checking errors''] \\
    
    4. \textbf{Logic Error}: [``Incorrect conditional statements'', ``Incorrect loop conditions'', ``Algorithm errors'', ``Variable scope errors'', ``Variable name conflicts'', ``Boundary condition errors''] \\
    
    5. \textbf{Performance Issue}: [``Memory leaks'', ``Performance bottlenecks'', ``Memory fragmentation'', ``High time complexity'', ``Resource contention''] \\
    
    6. \textbf{Design Flaw}: [``Resource leaks'', ``Lack of memory management'', ``Absence of modular design''] \\
    
    7. \textbf{Security Vulnerability}: [``Unsafe string operations'', ``Failure to prevent SQL injection'', ``Use of insecure random number generators'', ``Failure to handle external service unavailability'', ``Privilege escalation vulnerabilities''] \\
    
    8. \textbf{Configuration Management Error}: [``Not using version control systems'', ``Failure to back up code'', ``Absence of continuous integration and continuous delivery for code'', ``Configuration file errors'', ``Improper branch management'', ``Lack of tag management'', ``No tracking of historical versions''] \\
    
    9. \textbf{Data Management Error}: [``Database errors'', ``Data format errors'', ``Data consistency errors'', ``Data integrity errors''] \\
    
    10. \textbf{Concurrency and Multithreading Error}: [``Race conditions'', ``Deadlocks'', ``Thread safety issues''] \\
    
    11. \textbf{Input Validation and Data Processing Error}: [``Insufficient input validation'', ``Failure to handle boundary conditions'', ``Failure to handle user input formats'', ``Failure to filter user input'', ``Failure to validate user input''] \\
    
    12. \textbf{Exception Handling Error}: [``Improper exception handling'', ``Failure to handle exceptions'', ``Failure to use appropriate exception types'', ``Failure to use appropriate error codes or exception messages''] \\
    
    13. \textbf{Internationalization and Localization Error}: [``Failure to consider internationalization and localization'', ``Failure to handle timezone issues'', ``Failure to handle string operations in different locales'', ``Currency format errors'', ``Date format errors''] \\
    
    14. \textbf{Monitoring and Logging Management Error}: [``Insufficient logging'', ``Failure to use logging frameworks'', ``Lack of performance monitoring for code'', ``Failure to use appropriate log levels'', ``Failure to use appropriate log rotation mechanisms''] \\
    
    15. \textbf{Code Quality and Maintenance Error}: [``Comment errors'', ``Failure to follow coding standards'', ``Failure to conduct code reviews'', ``Failure to refactor code'', ``Failure to perform static code analysis''] \\
    
    16. \textbf{User Permission and Authentication Error}: [``Failure to handle user permission control'', ``Failure to handle user permission management'', ``Failure to handle user authentication and authorization'', ``Insufficient identity verification'', ``Permission validation errors''] \\
    
    17. \textbf{Testing and Verification Error}: [``Failure to use unit tests'', ``Insufficient modular testing of code'', ``Lack of automated testing for code'', ``Insufficient integration testing'', ``Insufficient regression testing''] \\
    
    18. \textbf{Network and Communication Error}: [``Failure to handle network request exceptions'', ``Failure to handle user request timeouts'', ``Network protocol errors'', ``Data transmission errors'', ``Connection timeouts''] \\
    
    19. \textbf{File and I/O Error}: [``File permission errors'', ``File format errors'', ``File path errors'', ``File locking issues'', ``Failure to handle file operation exceptions''] \\
    
    20. \textbf{Dependency Management Error}: [``Dependency management issues''] \\
    
    21. \textbf{Session Management Error}: [``Failure to handle user session management''] \\
    
    22. \textbf{Log Security Issue}: [``Log security issues''] \\
    
    23. \textbf{Environment Variable Error}: [``Environment variable errors''] \\

    \end{tcolorbox}
\end{center}
\caption{Full Error typelists.} \label{ap:prompt_error_typelist}
\end{figure*}

\begin{figure*}[h!]
\begin{center}
    \fontsize{8.4}{8.4} \selectfont
    \begin{tcolorbox}[width=1\textwidth, colback=lightblue, title={\textbf{Insert Bug Prompt}}]

    You are an experienced developer tasked with purposefully injecting errors into codes based on specified error types.\\
    
    \textbf{Task Objectives}:\\
    - Analyze the original code's structure and functionality.\\
    - Modify the code according to the provided error type.
    - Ensure that the injected error adheres to the specified category.\\
    - Preserve the overall functional integrity of the code.\\
    - Provide clear instructions for the error injection process.\\
    
    \textbf{Error Injection Requirements}:\\
    \textbf{1. Code Modification Principles}:\\
    - Clarity and precision: Modifications must be straightforward and targeted.\\
    - Category compliance: The injected error must belong to the specified error category.\\
    - Avoid unsanctioned errors: Only introduce errors of the specified type.\\
    - Preserve code integrity: Ensure that the other parts of the code remain unaffected.\\
    \textbf{2. Error Type Selection}:\\
    - Select from the specified error categories: Choose the error type from the predefined set of categories.\\
    - Subcategory compliance: Ensure that the subcategory chosen aligns with the main error category.\\
    - Full coverage: All specified error types must be incorporated.\\
    - Reasonable error placement: The error injection should be placed logically within the code, reflecting real-world scenarios.\\
    \textbf{3. Modification Description Requirements}:\\
    - Specify modification locations: Clearly define where each change takes place.\\
    - Detail the modification process: Provide a thorough explanation of how the modification is carried out.
    - Explain the error type: Clarify which error type corresponds to each modification.\\
    - Repair difficulty evaluation: Assess the ease or difficulty of repairing the error.\\
    
    \textbf{Output Format}:\\
    \verb|```|json\\
    \{\\
      ``Changed\_code'': <Complete modified code>,\\
      ``Selected error type category'': [<Error category 1>, ...],\\
      ``Selected error subcategory'': [<Error subcategory for category 1>, ...],\\
      ``Number of error types included'': <Integer>,\\
      ``Specific modified location and processing'': [\{\\
          .\quad``Location'': <Description of the code location>,\\
          .\quad``Corresponding error type'': <Error type for this modification>,\\
          .\quad``Modification description'': <Detailed explanation of the modification>\\
        .\quad\}, ...],\\
      ``Difficulty of repair'': <``Easy'' | ``Medium'' | ``Difficult''>\\
    \}\\
    \verb|```|\\
    
    \textbf{Error Modification Evaluation Criteria}:\\
    \textbf{1. Difficulty Level Definitions}:\\
    - Easy: The error is clear and the repair method is straightforward.\\
    - Medium: The error requires some analysis to identify and repair.\\
    - Difficult: The error demands a deep understanding of the code structure and the repair is complex.\\
    \textbf{2. Modification Quality Requirements}:\\
    - Natural error injection: Ensure the error injection appears seamless and not forced.\\
    - Readability: Maintain the code’s readability, ensuring the injected errors don’t obscure the logic.\\
    - Error type accuracy: Ensure each error type is accurately identified and implemented.\\
    
    \textbf{Notes}:\\
    - Provide a complete solution with all fields populated.\\
    - Ensure the instructions are clear and specific, with an emphasis on easy understanding.\\
    - The repair difficulty should be assessed objectively and realistically.\\
    - Maintain accuracy and completeness regarding the error types.\\

    \textbf{Input Data}:\\
    --- Start of Code ---\\
    \textcolor{ora}{\$Code}\\
    --- End of Code ---\\

    --- Start of Error Typelists ---\\
    \textcolor{ora}{\$Error\_Typelist}\\
    --- End of Error Typelists ---\\
    \end{tcolorbox}
\end{center}
\caption{Insert bug prompt.} \label{ap:prompt_insert_bug}
\end{figure*}

\begin{figure*}[h!]
\begin{center}
    \fontsize{8.4}{8.4} \selectfont
    \begin{tcolorbox}[width=1\textwidth, colback=lightblue, title={\textbf{Identifying Programming Error Types Prompt}}]

    You are an experienced programming expert responsible for professionally categorizing code error types.\\
    Your task is to analyze the code and determine which predefined error category it belongs to.\\
    
    \textbf{Available Error Categories}:\\
    1. Configuration Management Error\textcolor{ora}{\textbackslash n}
    2. Data Management Error\textcolor{ora}{\textbackslash n}
    3. Input Validation and Data Processing Error\textcolor{ora}{\textbackslash n}
    4. Monitoring and Logging Management Error\textcolor{ora}{\textbackslash n}
    5. Environment Variable Error\textcolor{ora}{\textbackslash n}
    6. Dependency Management Error\textcolor{ora}{\textbackslash n}
    7. Syntax Error\textcolor{ora}{\textbackslash n}
    8. Design Flaw\textcolor{ora}{\textbackslash n}
    9. Security Vulnerability\textcolor{ora}{\textbackslash n}
    10. Log Security Issue\textcolor{ora}{\textbackslash n}
    11. Reference Error\textcolor{ora}{\textbackslash n}
    12. Session Management Error\textcolor{ora}{\textbackslash n}
    13. Code Quality and Maintenance Error\textcolor{ora}{\textbackslash n}
    14. Logic Error\textcolor{ora}{\textbackslash n}
    15. Testing and Verification Error\textcolor{ora}{\textbackslash n}
    16. Network and Communication Error\textcolor{ora}{\textbackslash n}
    17. Exception Handling Error\textcolor{ora}{\textbackslash n}
    18. User Permission and Authentication Error\textcolor{ora}{\textbackslash n}
    19. File and I/O Error\textcolor{ora}{\textbackslash n}
    20. Type Error\textcolor{ora}{\textbackslash n}
    21. Internationalization and Localization Error\textcolor{ora}{\textbackslash n}
    22. Performance Issue\textcolor{ora}{\textbackslash n}
    23. Concurrency and Multithreading Error\\
    
    \textbf{Category Definitions}:\\
    - Configuration Management Error: Issues related to system configuration settings\\
    - Data Management Error: Problems with data handling, storage, or retrieval\\
    - Input Validation Error: Failures in validating or processing input data\\
    - Monitoring/Logging Error: Issues with system monitoring or logging functions\\
    - Environment Variable Error: Problems with environment variables\\
    - Dependency Management Error: Issues with external dependencies or libraries\\
    - Syntax Error: Basic programming language syntax violations\\
    - Design Flaw: Fundamental problems in code architecture or design\\
    - Security Vulnerability: Security-related weaknesses\\
    - Log Security Issue: Security problems in logging mechanisms\\
    - Reference Error: Invalid references to variables or objects\\
    - Session Management Error: Problems with user session handling\\
    - Code Quality Error: Issues affecting maintainability or readability\\
    - Logic Error: Flaws in business logic or algorithms\\
    - Testing/Verification Error: Problems with testing processes\\
    - Network/Communication Error: Issues in network interactions\\
    - Exception Handling Error: Problems with error handling\\
    - User Permission Error: Authentication or authorization issues\\
    - File/I/O Error: Problems with file operations\\
    - Type Error: Issues with data types or type conversions\\
    - Internationalization Error: Problems with language/locale support\\
    - Performance Issue: Efficiency or resource usage problems\\
    - Concurrency Error: Issues with parallel processing\\
    
    \textbf{Categorization Requirements}:\\
    1. Primary Analysis\\
       .\quad- Carefully examine the code structure and syntax\\
       .\quad- Identify all existing errors\\
       .\quad- Match errors against the predefined categories\\
    2. Classification Criteria:\\
       .\quad- Exact Match: The error perfectly fits one of the predefined categories\\
       .\quad- Best Fit: If multiple issues exist, select the most significant one\\
       .\quad- No Match: If the error doesn't match any predefined category\\
    3. Analysis Steps:\\
       .\quad- Locate the specific error in the code\\
       .\quad- Compare it with the predefined categories\\
       .\quad- Determine the most appropriate classification\\
       .\quad- Provide reasoning for the classification\\
    
    \textbf{Output Format}:\\
    \verb|```|json\\
    \{\\
      .\quad"Category": "<Selected error category from the predefined list>",\\
      .\quad"Confidence": "<High/Medium/Low>"\\
    \}\\
    \verb|```|\\
    
    \textbf{Classification Notes}:\\
    - Only select one primary category even if multiple errors exist\\
    - If no predefined category matches, select the most appropriate error category\\
    - Include specific code references in the explanation\\
    - Indicate confidence level in the classification\\
    - Provide clear reasoning for the chosen category\\
    
    \textbf{Additional Guidelines}:\\
    - Focus on syntactic and structural analysis\\
    - Consider the context of the entire code\\
    - Be specific about error location\\
    - Explain any ambiguous cases\\
    - Maintain objectivity in classification\\

    \textbf{Input Data}:\\
    --- Start of Code ---\\
    \textcolor{ora}{\$Code}\\
    --- End of Code ---\\

    \end{tcolorbox}
\end{center}
\caption{Identifying programming error types prompt.} \label{ap:prompt_bug}
\end{figure*}

\begin{figure*}[h!]
\begin{center}
    \fontsize{8.4}{8.4} \selectfont
    \begin{tcolorbox}[width=1\textwidth, colback=lightblue, title={\textbf{Prompt for Classify Application Scenarios in ``Code QA'' Subset}}]

    As a professional expert in technical field classification, you are tasked with accurately categorizing programming problems and code snippets into the appropriate technical domains.\\
    
    \textbf{Required Expertise}:\\
    1. A deep understanding of programming concepts and the distinctive characteristics of various technical fields.\\
    2. The ability to accurately identify the core technical features of code and problems.\\
    3. Proficiency in discerning the boundaries and relationships between different technical domains.\\
    4. Capacity to provide well-reasoned and persuasive justification for each classification.\\
    
    \textbf{Technical Field Labels}:\\
    1. Basic Programming: Foundational concepts such as syntax, control flow and data structures.\\
    2. Advanced Programming: In-depth topics like design patterns, concurrency and performance optimization.\\
    3. Software Engineering: Practices including project management, code quality, testing and deployment.\\
    4. Data Analysis: Data processing, statistical analysis and data visualization.\\
    5. Mathematics: Algorithms, computation methods and mathematical modeling.\\
    6. Desktop and Web Development: User interface design, front-end and back-end development and user interactions.\\
    7. Machine Learning: Model training, deep learning and artificial intelligence.\\
    8. Scientific Computing: Numerical computations, simulations and scientific modeling.\\
    9. Database: Data storage, query optimization and database design.\\
    10. Multimedia: Image, audio and video processing, as well as multimedia applications.\\
    11. Operating Systems: System programming, memory management and process scheduling.\\
    12. Other: Fields that do not fall into the categories above.\\
    
    \textbf{Evaluation Process}:\\
    1. Carefully Review the given problem and code.\\
    2. Identify the core technical features and key concepts.\\
    3. Match these features with the relevant technical fields.\\
    4. Select the most appropriate label from the predefined list.\\
    5. Justify your selection with detailed reasoning.\\
    
    \textbf{Output Format}:\\
    \verb|```|md\\
    Label classification: <Chosen label from the predefined list>\\
    Label selection rationale: <Provide a detailed explanation, addressing the following aspects:>\\
    .\quad- Analysis of the core technical features\\
    .\quad- Alignment with the selected technical field\\
    .\quad- Reasons for not selecting other relevant fields\\
    .\quad- Confidence level in the selection\\
    \verb|```|\\
    
    \textbf{Guidelines}:\\
    - Only one predefined label should be selected.\\
    - Avoid ambiguity in your decision-making process.\\
    - Provide clear, specific reasoning for your classification.\\
    - Focus on the core technical aspects of the problem rather than superficial characteristics.\\
    Make sure to select the label that best represents the core technical features of the problem.\\

    \textbf{Input Data}:\\
    --- Start of Question ---\\
    \textcolor{ora}{\$Question}\\
    --- End of Question ---\\
    
    --- Start of Answer ---\\
    \textcolor{ora}{\$Answer}\\
    --- End of Answer ---\\
    \end{tcolorbox}
\end{center}
\caption{Prompt for classify application scenarios in ``Code QA'' subset.} \label{ap:prompt_real_classify}
\end{figure*}

\begin{figure*}[h!]
\begin{center}
    \fontsize{8.4}{8.4} \selectfont
    \begin{tcolorbox}[width=1\textwidth, colback=lightblue, title={\textbf{Fine-Grained Checklists Generation Prompt for ``Code Gen'' Subset}}]

    You are a senior code reviewer with extensive technical expertise.\\
    Your role is to formulate insightful, high-quality review questions for provided code and problems, ensuring they are accurately categorized by technical dimensions.\\
    
    \textbf{Criteria for evaluation questions}:\\
    - Conduct a thorough analysis of the design intent and implementation strategy of the solution.\\
    - Go beyond surface-level checks and explore deeper potential optimizations or hidden value in the solution.\\
    - Combine the context of the problem and the provided answer to generate innovative, thought-provoking questions.\\
    - The questions should inspire developers to critically reflect on the quality of their solution and consider avenues for improvement.\\
    - Ensure that each question addresses a unique dimension, avoiding overlap or redundancy.\\
    
    For each of the following dimensions, generate one question. Each question should focus on the corresponding evaluation aspect.\\
    \textbf{Evaluation Dimensions}:\\
    1. \textbf{Correctness Verification}: Assess whether the code is capable of solving the problem accurately and passing all test cases.\\
    2. \textbf{Time Complexity Optimization}: Focus on optimizing the efficiency and performance of the algorithm.\\
    3. \textbf{Space Complexity}: Evaluate how efficiently the code uses memory and manages resources.\\
    4. \textbf{Code Readability}: Focus on improving the clarity, understandability and standardization of the code.\\
    5. \textbf{Robustness Validation}: Evaluate the code's ability to handle edge cases and exceptions.\\
    6. \textbf{Algorithm Optimization}: Assess whether the algorithm's design and implementation are optimal.\\
    7. \textbf{Comprehensive Testing}: Evaluate the scope and strategy behind the test coverage.\\
    8. \textbf{Output Format}: Verify that the code's output strictly adheres to the required format.\\
    9. \textbf{Code Style Consistency}: Ensure that the code follows consistent coding standards and best practices.\\
    10. \textbf{Maintainability}: Assess whether the code structure is modular, easily optimized and expandable in the future.\\
    
    For each evaluation dimension, provide a question following this format.\\
    \textbf{Output Format}:\\
    \verb|```|md\\
    1. Question: <Specific judgment question>\\
    .\;\;\;Dimension: <Select the most relevant evaluation .\;\;\;dimension>\\
    .\;\;\;Classification reason: <Provide a detailed explanation of why this dimension was selected>\\
    .\;\;\;Classification confidence: <Integer value between 1-10>\\
    ...\\
    \verb|```|\\
    
    Ensure that all 10 dimensions are covered with their respective questions, with each question addressing a distinct aspect of the code's quality.\\

    \textbf{Input Data}:\\
    --- Start of Question ---\\
    \textcolor{ora}{\$Question}\\
    --- End of Question ---\\
    
    --- Start of Answer ---\\
    \textcolor{ora}{\$Answer}\\
    --- End of Answer ---\\
    \end{tcolorbox}
\end{center}
\caption{Fine-Grained checklists generation prompt for ``Code Gen'' subset.} \label{ap:prompt_algo_diy_qa}
\end{figure*}

\begin{figure*}[h!]
\begin{center}
    \fontsize{8.4}{8.4} \selectfont
    \begin{tcolorbox}[width=1\textwidth, colback=lightblue, title={\textbf{Fine-Grained Checklists Generation Prompt for ``Code QA'' Subset}}]

    You are a senior code review expert with deep technical expertise.\\
    
    Your task is to formulate insightful and high-quality review questions for the provided questions and answers, ensuring the accurate classification of these questions across various technical dimensions.\\
    
    \textbf{Criteria for Evaluating Questions}:\\
    - Conduct an in-depth analysis of the design intent and implementation strategy presented in the answer.\\
    - Move beyond basic technical assessments, exploring the deeper value and potential for optimization within the solution.\\
    - Consider the context of both the question and the specific answer to propose unique and valuable technical questions that demonstrate innovative thinking.\\
    - Ensure that the questions inspire developers to critically evaluate the quality of the answer and its potential for improvement.\\
    - Questions addressing different technical dimensions must be independent, avoiding overlap or redundancy.\\
    
    For each of the following dimensions, generate one question. Each question should focus on the corresponding evaluation aspect.\\
    \textbf{Evaluation Dimensions}:\\
    1. \textbf{Correctness} Ensure the solution is accurate and addresses the core needs of the problem.\\
    2. \textbf{Completeness}: Ensure the answer comprehensively covers all aspects of the problem without leaving out any critical information.\\
    3. \textbf{Performance}: Assess whether the solution optimizes time and space complexity effectively.\\
    4. \textbf{Maintainability}: Evaluate the clarity and structure of the solution to ensure it is easy to optimize and extend in the future.\\
    5. \textbf{Clarity:} Ensure that the language used is concise and clear, facilitating understanding and quick access to essential information.\\
    6. \textbf{Depth}: Evaluate whether the answer provides an in-depth analysis of the problem, offering a thorough understanding rather than a superficial response.\\
    7. \textbf{Practicality}: Ensure that the solution is feasible and can be implemented effectively in real-world scenarios.\\
    8. \textbf{Logic Coherence}: Assess whether the reasoning behind the solution is clear, coherent and convincing.\\
    9. \textbf{Innovation}: Determine if the solution offers a unique perspective or introduces an innovative approach.\\
    10. \textbf{Reliability}: Ensure that the opinions and suggestions are based on reliable, verifiable evidence.\\

    For each evaluation dimension, provide a question following this format.\\
    \textbf{Output Format}:\\
    \verb|```|md\\
    1. Question: <Specific judgment question>\\
    .\;\;\;Dimension: <Select the most relevant evaluation .\;\;\;dimension>\\
    .\;\;\;Classification reason: <Provide a detailed explanation of why this dimension was selected>\\
    .\;\;\;Classification confidence: <Integer value between 1-10>\\
    ...\\
    \verb|```|\\
    
    Ensure that all 10 dimensions are covered with their respective questions, with each question addressing a distinct aspect of the answer's quality.\\

    \textbf{Input Data}:\\
    --- Start of Question ---\\
    \textcolor{ora}{\$Question}\\
    --- End of Question ---\\
    
    --- Start of Answer ---\\
    \textcolor{ora}{\$Answer}\\
    --- End of Answer ---\\
    \end{tcolorbox}
\end{center}
\caption{Fine-Grained checklists generation prompt for ``Code QA'' subset.} \label{ap:prompt_real_diy_qa}
\end{figure*}

\begin{figure*}[h!]
\begin{center}
    \fontsize{8.4}{8.4} \selectfont
    \begin{tcolorbox}[width=1\textwidth, colback=lightblue, title={\textbf{Basic Critique Evaluation Prompt (with CoT)}}]

    You are an AI Senior Programming Expert with advanced analytical reasoning capabilities.\\
    
    Your task is to conduct a comprehensive, step-by-step evaluation of an AI assistant's solution using a detailed Chain of Thought (CoT) approach.\\
    
    \textbf{Evaluation Framework}:\\
    
    \textbf{Stage 1: Comprehensive Problem Understanding}\\
    - Carefully analyze the original problem statement\\
    - Identify explicit and implicit requirements\\
    - Assess technical complexity and contextual constraints\\
    - Determine key evaluation dimensions\\
    
    \textbf{Stage 2: Systematic Solution Decomposition}\\
    - Break down the proposed solution into discrete components\\
    - Evaluate each component's technical correctness\\
    - Assess code quality, efficiency and alignment with best practices\\
    - Trace the logical flow and problem-solving approach\\
    
    \textbf{Stage 3: Detailed Performance Assessment}\\
    - Verify functional accuracy \\
    - Analyze performance optimization potential\\
    - Evaluate scalability and maintainability\\
    - Check comprehensive error handling strategies\\
    
    \textbf{Evaluation Methodology}:\\
    1. Maintain complete objectivity\\
    2. Provide specific, actionable feedback\\
    3. Explain reasoning transparently at each stage\\
    4. Highlight both solution strengths and improvement opportunities\\
    
    \textbf{Output Requirements}:\\
    - Provide a clear, structured Chain of Thought explanation\\
    - Render a definitive assessment: [Correct/Error]\\
    - Include detailed reasoning for each evaluation stage\\
    - Offer constructive, implementable improvement suggestions\\
    
    \textbf{Evaluation Criteria}:\\
    - Technical Precision\\
    - Code Quality\\
    - Problem-Solving Approach\\
    - Efficiency and Optimization\\
    - Adherence to Best Practices\\
    
    \textbf{Output Format}:\\
    \verb|```|json\\
    \{\\
        .\quad"is\_assistant\_correct": "Correct/Error",\\
        .\quad"detailed\_reasoning": "Explicit step-by-step logical deduction"\\
    \}\\
    \verb|```|\\
    
    \textbf{Input Data}:\\
    --- Start of Question ---\\
    \textcolor{ora}{\$Question}\\
    --- End of Question ---\\
    
    --- Start of Answer ---\\
    \textcolor{ora}{\$Answer}\\
    --- End of Answer ---\\

    \end{tcolorbox}
\end{center}
\caption{Basic critique evaluation prompt (with CoT).} \label{ap:prompt_cot_level1}
\end{figure*}

\begin{figure*}[h!]
\begin{center}
    \fontsize{8.4}{8.4} \selectfont
    \begin{tcolorbox}[width=0.8\textwidth, colback=lightblue, title={\textbf{Advanced Critique Evaluation Prompt (with CoT)}}]

    You are an Advanced Code Evaluation Specialist with Comprehensive Analytical Capabilities\\
    
    Evaluation Methodology: Systematic Chain of Thought (CoT) Code Assessment\\
    
    \textbf{Core Evaluation Framework}:\\
    
    \textbf{Stage 1: Holistic Problem Understanding}\\
    - Conduct an in-depth analysis of the problem domain\\
    - Identify multi-dimensional evaluation criteria\\
    - Map out complex technical expectations\\
    - Establish a comprehensive assessment perspective\\
    
    \textbf{Stage 2: Detailed Code Decomposition}\\
    - Systematically break down code into core components\\
    - Analyze each segment through multiple lenses:\\
      .\quad1. Technical Correctness\\
      .\quad2. Algorithmic Efficiency\\
      .\quad3. Code Readability\\
      .\quad4. Scalability\\
      .\quad5. Best Practice Adherence\\
    
    \textbf{Stage 3: Precision Scoring Methodology}\\
    Evaluation Dimensions (Each Scored 1-10):\\
    1. Algorithmic Complexity and Efficiency\\
    2. Code Structure and Readability\\
    3. Error Handling and Robustness\\
    4. Performance Optimization\\
    5. Solution Creativity\\
    6. Technical Precision\\
    7. Memory Management\\
    8. Scalability Potential\\
    9. Maintainability\\
    10. Overall Problem-Solving Approach\\
    
    \textbf{Scoring Nuanced Guidelines}:\\
    - 1-2 points: Critical, fundamental failures\\
    - 3-4 points: Significant technical deficiencies\\
    - 5-6 points: Functional but requires major improvements\\
    - 7-8 points: Solid implementation with minor refinement needs\\
    - 9-10 points: Exceptional, near-perfect solution\\
    
    \textbf{Comprehensive Evaluation Process}:\\
    - Conduct granular, multi-dimensional assessment\\
    - Provide explicit reasoning for each score\\
    - Generate weighted comprehensive evaluation\\
    - Offer constructive, actionable improvement suggestions\\
    
    \textbf{Reasoning Chain Output Format}:\\
    \verb|```|json\\
    \{\\
        .\quad"comprehensive\_score": "Weighted comprehensive score",\\
        .\quad"detailed\_reasoning": "Explicit step-by-step logical deduction"\\
    \}\\
    \verb|```|\\
    
    \textbf{Input Data}:\\
    --- Start of Question ---\\
    \textcolor{ora}{\$Question}\\
    --- End of Question ---\\
    
    --- Start of Answer ---\\
    \textcolor{ora}{\$Answer}\\
    --- End of Answer ---\\
    
    --- Start of Fine-Grained Evaluation Checklists ---\\ 
    \textcolor{ora}{\$Checklists}\\
    --- End of Fine-Grained Evaluation Checklists ---\\

    \end{tcolorbox}
\end{center}
\caption{Advanced critique evaluation prompt (with CoT).} \label{ap:prompt_cot_level2}
\end{figure*}

\begin{figure*}[h!]
\begin{center}
    \fontsize{8.4}{8.4} \selectfont
    \begin{tcolorbox}[width=0.6\textwidth, colback=lightblue, title={\textbf{Prompt for Refining Origin Answer Based on Model's Critiques}}]

    You are a Professional Self-Correction AI Specialist\\
    
    Objective: Systematically Refine and Improve Your Original Response\\
    
    \textbf{Evaluation Process}:\\
    1. Carefully review the detailed feedback provided\\
    2. Critically analyze your original response\\
    3. Identify specific areas requiring improvement\\
    4. Develop a comprehensive, precise correction strategy\\
    
    \textbf{Key Refinement Focus Areas}:\\
    - Accuracy of information\\
    - Completeness of solution\\
    - Clarity of explanation\\
    - Technical depth\\
    - Problem-solving approach\\
    
    \textbf{Correction Guidelines}:\\
    - Address each piece of feedback explicitly\\
    - Provide clear rationale for modifications\\
    - Enhance the original response's quality\\
    - Maintain the core intent of the original answer\\
    
    \textbf{Output Requirements}:\\
    1. Detailed explanation of identified limitations\\
    2. Comprehensive corrected response\\
    3. Specific improvements made\\
    4. Reasoning behind each modification\\
    
    \textbf{Output Fromat}:\\
    \verb|```|json\\
    \{\\
        .\quad"corrected\_response": "Fully updated and refined answer",\\
    \}\\
    \verb|```|\\
    
    \textbf{Critical Evaluation Principles}:\\
    - Be objective and critical\\
    - Focus on substantive improvements\\
    - Demonstrate intellectual rigor\\
    - Enhance overall response quality\\
    
    \textbf{Mandatory Considerations}:\\
    - Fully address all feedback points\\
    - Provide clear, precise modifications\\
    - Maintain professional and technical accuracy\\
    
    \textbf{Input Data}:\\
    --- Start of Question ---\\
    \textcolor{ora}{\$Question}\\
    --- End of Question ---\\
    
    --- Start of Answer ---\\
    \textcolor{ora}{\$Answer}\\
    --- End of Answer ---\\
    
    --- Start of Feedback ---\\  
    \textcolor{ora}{\$Feedback}\\
    --- End of Feedback ---\\

    \end{tcolorbox}
\end{center}
\caption{Prompt for refining origin answer based on model's critiques.} \label{ap:prompt_refine}
\end{figure*}

\end{CJK*}
\end{document}